\documentclass[a4paper,fleqn]{cas-sc}

\usepackage{booktabs}
\usepackage{multirow}
\usepackage{amssymb}  
\usepackage{amsmath}
\usepackage{enumitem}
\usepackage{makecell}

\usepackage{graphicx}
\usepackage{subfig}
\usepackage{rotating}
\usepackage{subcaption}
\usepackage{longtable}
\usepackage{hyperref}

\usepackage{array}
\usepackage[numbers]{natbib}
\usepackage{adjustbox}
\usepackage{ltablex} 
\usepackage{threeparttablex}

\usepackage{tikz}
\usepackage{xcolor}
\usetikzlibrary{calc, shapes.misc, positioning,arrows.meta}

\usepackage{xcolor}
\newcommand{\showrevisions}{1}

\newcommand{\rev}[1]{%
  \ifnum\showrevisions=1
    {\color{blue}#1}%
  \else
    #1%
  \fi
}

\usepackage{array}
\newcolumntype{F}{%
  >{\raggedright\arraybackslash\fontsize{7.5pt}{8pt}\selectfont}%
  p{2.7cm}%
}
\usepackage{makecell}

\newcommand{\legendrow}[1]{%
\multicolumn{19}{p{\linewidth}}{%
\footnotesize\setlength{\baselineskip}{8.5pt}%
\textcolor{black}{#1}}\\}

\newcommand{\legendrownew}[1]{%
\multicolumn{6}{p{\linewidth}}{%
\footnotesize\setlength{\baselineskip}{8.5pt}%
\textcolor{black}{#1}}\\}

\usepackage{xcolor}
\usepackage{booktabs}
\usepackage{tabularx}
\usepackage{multicol}

\usepackage{pgfplots}
\pgfplotsset{compat=1.18}

\def\tsc#1{\csdef{#1}{\textsc{\lowercase{#1}}\xspace}}
\tsc{WGM}
\tsc{QE}

\begin{document}
\let\WriteBookmarks\relax
\def\floatpagepagefraction{1}
\def\textpagefraction{.001}

\shorttitle{}    

\shortauthors{}  

\title [mode = title]{What Demands Attention in Urban Street Scenes? From Scene Understanding towards Road Safety: A Survey of Vision-driven Datasets and Studies}  

\author[1]{Yaoqi Huang}\corref{cor1}
\ead{y.huang@acfr.usyd.edu.au}

\author[1]{Julie Stephany Berrio}
\ead{j.berrio@acfr.usyd.edu.au}

\author[1]{Mao Shan}
\ead{m.shan@acfr.usyd.edu.au}

\author[1]{Stewart Worrall}
\ead{s.worrall@acfr.usyd.edu.au}

\cortext[cor1]{Corresponding author}

\affiliation[1]{organization={Australian Centre For Robotics (ACFR), The University of Sydney},
            city={Sydney},
            postcode={2008},
            state={NSW},
            country={Australia}}

\begin{abstract}
Advances in vision-based sensors and computer vision algorithms have significantly improved the analysis and understanding of traffic scenarios. To facilitate the use of these improvements for road safety, this survey systematically categorizes the critical elements that demand attention in traffic scenarios and comprehensively analyzes available vision-driven tasks and datasets. Compared to existing surveys that focus on isolated domains, our taxonomy categorizes attention-worthy traffic entities into two main groups, namely \textit{anomalies} (abnormal entities) and \textit{pertinent} entities (normal but critical elements), integrating eleven categories and twenty-three subclasses. It establishes connections between inherently related fields and provides a unified analytical framework. Based on the proposed taxonomy, our survey highlights the analysis of 40 vision-driven tasks and the comprehensive examinations and visualizations of 78 available datasets, including their basic characteristics, sensor settings, label design, visualization practices, annotation schemas, and the resulting implications. The cross-domain investigation reveals substantial variations in benchmark quality across tasks, with recurring limitations including uneven task coverage, imbalanced distributions, inconsistent or insufficient annotations, and limited multimodal and cross-task support. Our article further outlines promising solutions from the perspectives of task formulation, benchmark evaluations, dataset adoption and future dataset construction. The integrated taxonomy, comprehensive analysis, and recapitulatory tables provide researchers with a holistic overview of this rapidly evolving field, guiding strategic resource selection, and highlighting critical yet underexplored areas. 
\end{abstract}

\begin{keywords}
Computer Vision \sep Image Processing \sep Scene Understanding \sep Road Safety \sep Traffic Accidents
\end{keywords}

\maketitle

\section{Introduction}

Automobile industry has brought convenience to daily life. However, increasing vehicle ownership and usage has also put pressure on road safety. 
According to the World Health Organization \cite{WHO}, road traffic crashes cause approximately 1.19 million deaths and 20–50 million non-fatal injuries (including disability) each year, and are the leading cause of death in children and young adults. More than half of road fatalities are among vulnerable road users and 92\% occur in low- and middle-income countries. The economic losses caused by traffic crashes are estimated at 3\% of gross domestic product in most countries.
In response to growing public concerns about security and safety, cameras have been installed in vehicles and distributed throughout cities. Vision-based sensors are capable of perceiving rich surrounding information \cite{7464298}. Powered by advances in communication networks and computer vision, numerous tasks have been proposed to analyze and understand traffic scenarios from different perspectives using visual data. For example, panoptic segmentation \cite{PanopticSegmentation} aims to create a holistic view of traffic scenes by distinguishing each individual road user in the foreground and classifying each background pixel. The traffic scene graphs \cite{9900075} are designed to represent the traffic entities and their pairwise relationships in a graphical structure. Although these tasks are beneficial in ensuring the completeness of interpretation, they are not specialized in depicting the traffic scenario from a road safety perspective. 
Achieving this objective requires answering a fundamental question:
\textbf{Which elements are most critical in the current traffic scenario and require attention to road safety?}

Explicitly or implicitly, numerous studies have offered definitions of `\textit{critical elements}' \cite{mine, 9703244, 10225448} and interpretations of `\textit{attention}' \cite{mine, Xia2017PredictingDA, DADA} while exploring different regions or objects in traffic scenarios. For example, significant attention has been paid to anomalies in traffic scenes, employing various identification standards \cite{AnomalyDetectionSurvey, DoTA, RoadDamageSurvey}. Some of them focus on individuals or groups with abnormal behavior \cite{DoTA, DAD, CrashToNotCrash, JAAD}, such as jaywalking or illegal parking, while others examine individuals in unfamiliar or unseen semantic categories \cite{lostAndFound, chan2021segmentmeifyoucan, RoadAnomalyDataset, fishyscapes}. Meanwhile, given the leading position of road accidents for death \cite{WHO}, considerable effort has been put into detecting or anticipating potential traffic accidents \cite{DAD, CADPDataset, CarCrashDataset, A3D}. Although some studies classify these tasks as traffic accident detection or anticipation \cite{trafficAccidentsSurvey}, others also categorize them as traffic anomaly detection by treating traffic accidents as anomalous events \cite{DoTA}. Furthermore, the term `anomaly' takes on different meanings in various application contexts \cite{AnomalyDetectionSurvey}. Studies proposed for different tasks might address the same type of problem. In addition, even when studies do not explicitly label their mission with standard terminology, they still demonstrate clear connections to specific domains. For example, while LostAndFound \cite{lostAndFound} is described as a small obstacle detection dataset, its pixel-level annotation confirms its affiliation with obstacle segmentation task, and the nine previously unseen objects in its test subset demonstrate its suitability for anomaly segmentation applications.

\begin{figure*}[!t]
\centering
\includegraphics[width=1\textwidth]{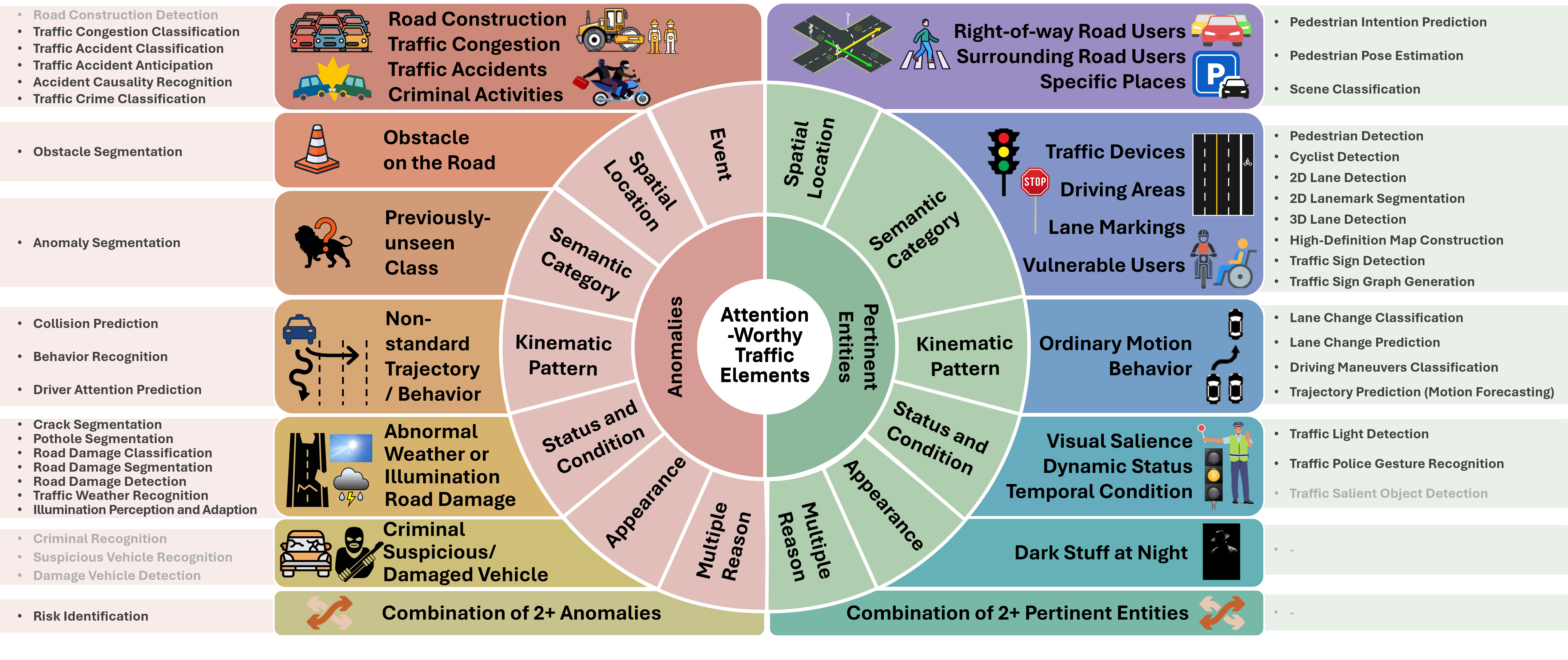}
\caption{Graphical taxonomy diagram of \textit{attention-worthy elements} for road safety, organized by reason (\textit{anomaly} vs.\ \textit{pertinent entity}) and category. Leaf nodes list the corresponding \textit{tasks} and representative datasets. Categories with empty task node indicate the lack of well-discussed tasks. Tasks shown in gray indicate a lack of publicly available datasets.}
\label{fig:graphic_taxonomy}
\end{figure*}

Hence, to prevent overlap and ensure comprehensive coverage, in this paper, we propose a taxonomy that classifies `critical traffic elements that demand attention' based on their functional role in `road safety'. As illustrated by the graphical taxonomy diagram in Fig.~\ref{fig:graphic_taxonomy}, the taxonomy was initially derived from thirteen distinct perspectives, integrating insights from traffic scene understanding literature and practical road-safety considerations. These perspectives were then grouped according to whether the element is safety-critical because it is abnormal in the scene or because it is normal but particularly relevant to the current driving task. Two perspectives were not retained in the final taxonomy due to insufficient supporting studies. As a result, as organized in Table \ref{table:Existing Tasks Classification} and summarized below, the final taxonomy consists of \textbf{two main groups}, \textbf{eleven categories}, covering \textbf{twenty-three common research topics}.
\begin{enumerate}
    \item \textbf{\textit{Anomaly}} that is abnormal because of its (1) spatial location (e.g. obstacle on the road), (2) semantic category (e.g. a trash bin), (3) event (including construction, congestion, accident, and crime), (4) kinematic pattern (e.g. trajectory), (5) status or condition (including road surface, weather, and illumination condition), (6) appearance (e.g. criminal and suspicious license plate), or (7) hybrid of aforementioned anomalies; and
    \item \textbf{\textit{Pertinent Entity}} that is normal but critical to current driving maneuver or driving goal because of its (1) spatial location (e.g. surrounding road users), (2) semantic category (including pedestrian, rider, driving lane, and traffic sign), (3) status or condition (e.g. traffic light states, traffic police gesture, and visual salience), or (4) kinematic pattern (e.g. motion behavior, and trajectory).
\end{enumerate}

These elements warrant the attention of the \textit{ego vehicle} to minimize risks. In this survey, \textit{ego vehicle} refers to the host vehicle carrying the onboard sensing system, from whose perspective the traffic scene is perceived and analyzed. The `risk' can be \textit{objective risk} that refers to the objective probability of being involved in an accident \cite{9423525,8917457} or \textit{subjective risk} that refers to the probability of an accident estimated by drivers through the cognition process \cite{9423525,8917457}.

\paragraph{\textbf{Scope.}}
Our target audiences are researchers who explore traffic scene understanding for road safety. We aim to provide them with a comprehensive overview of vision-based perception tasks and available vision-driven datasets that describe critical attention-worthy elements in traffic scenarios.
Based on the proposed taxonomy, candidate tasks and datasets were identified through major academic publication platforms, together with tracing references in related papers, project webpages, and code repositories. Candidate benchmarks were then screened according to whether they address perception-oriented scene understanding of the external driving environment. Accordingly, approaches dedicated to driver-facing or in-cabin analysis, such as inappropriate driver behavior of the ego vehicle (e.g. fatigue driving or drowsiness) \cite{kopuklu2021driver}, or cockpit cues reflected on the instrument panel or mirrors, are excluded from this survey. Studies solely based on nonvisual modalities (e.g. Global Navigation Satellite System data), as well as those primarily focused on planning, are also outside the scope of this work. As a result, this survey ultimately includes the analysis of \textbf{40 tasks} and detailed examination and visualization of \textbf{78 datasets}. 
Notably, each surveyed dataset is classified according to the type of attention-worthy traffic element it addresses and the form in which that element is represented. It suggests if the claimed task of some dataset is not aligned with what its data and annotation actually support, we classify it according to its actual functional format. For those versatile datasets that have been widely used across multiple surveyed tasks, we discuss them in all relevant task sections but only elaborate them at their first appearance. In contrast, datasets with the potential to support multiple tasks are categorized according to their broadest or most comprehensive potential applications. To our knowledge, it is the first survey that categorizes scene-understanding-driven road safety studies based on the type of attention-demanding scene elements within the driving environment.
\definecolor{vTopCV}{RGB}{33,102,172}
\definecolor{vCVc}{RGB}{116,169,207}
\definecolor{vML}{RGB}{106,61,154}
\definecolor{vRob}{RGB}{51,160,44}
\definecolor{vTrf}{RGB}{255,127,0}
\definecolor{vTopCVj}{RGB}{227,26,28}
\definecolor{vCVj}{RGB}{251,154,153}
\definecolor{vOth}{RGB}{177,89,40}
\definecolor{vOnl}{RGB}{153,153,153}

\definecolor{bc0}{RGB}{184,210,240}
\definecolor{bc1}{RGB}{170,198,233}
\definecolor{bc2}{RGB}{143,177,218}
\definecolor{bc3}{RGB}{116,156,204}
\definecolor{bc4}{RGB}{89,135,189}
\definecolor{bc5}{RGB}{62,114,175}
\definecolor{bc6}{RGB}{45,95,160}
\definecolor{bc7}{RGB}{38,78,143}
\definecolor{bc8}{RGB}{30,62,126}
\definecolor{bc9}{RGB}{22,47,109}
\definecolor{bc10}{RGB}{15,33,93}
\definecolor{bc11}{RGB}{8,20,77}

\newcommand{\inlinebarv}[2]{\textcolor{#1}{\rule{#2}{5pt}}}

\begin{figure}[t!]
\centering
\captionsetup{justification=centering,singlelinecheck=false}
\setlength{\tabcolsep}{2pt}
\renewcommand{\arraystretch}{1.1}

\begin{minipage}[t]{0.55\textwidth}
\vspace{0pt}
\scriptsize
\begin{tabular}{@{}lrl@{}}
  \toprule
  \textbf{Venue} & $n$ & \\
  \midrule
  CVPR/ICCV/ECCV & 17 & \inlinebarv{vTopCV}{1.70cm} \\[-1pt]
  ACM MM/ICCVW/CVPRW/ACCV/AVSS/DICTA & 12 & \inlinebarv{vCVc}{1.20cm} \\[-1pt]
  NeurIPS/AAAI/IJCNN &  8 & \inlinebarv{vML}{0.8cm}  \\[-1pt]
  ICRA/IROS/RA-L/ACRA &  10 & \inlinebarv{vRob}{1.0cm} \\[-1pt]
  ITSC/IV/T-ITS & 14 & \inlinebarv{vTrf}{1.40cm} \\[-1pt]
  T-PAMI/T-IP &  7 & \inlinebarv{vTopCVj}{0.7cm} \\[-1pt]
  CVIU/Computers \& Graphics/Pattern Recognition Letter & 3 & \inlinebarv{vCVj}{0.30cm} \\[-1pt]
  Neurocomputing/Constr.~Build./Geosci.~Data/IoT/EC3 &  5 & \inlinebarv{vOth}{0.50cm} \\[-1pt]
  arXiv/IEEE~Dataport &  2 & \inlinebarv{vOnl}{0.20cm} \\
  \bottomrule
\end{tabular}
\end{minipage}
\hspace{0.03\textwidth}
\begin{minipage}[t]{0.33\textwidth}
\vspace{0pt}
\begin{tikzpicture}[font=\footnotesize]
  \def\W{4.0}   
  \def\H{2.2}   
  \def\mL{0.6}  
  \def\maxV{17}
  \def\bw{0.22}

  \foreach \yr/\cnt/\col/\xi in {
    2009/1/bc0/0,   2012/1/bc1/1,
    2016/7/bc2/2,   2017/5/bc3/3,
    2018/6/bc4/4,   2019/14/bc5/5,
    2020/16/bc6/6,  2021/9/bc7/7,
    2022/8/bc8/8,   2023/6/bc9/9,
    2024/4/bc10/10, 2025/1/bc11/11%
  }{
    \pgfmathsetmacro{\xp}{\mL + (\xi/11)*\W}
    \pgfmathsetmacro{\bh}{\cnt/\maxV*\H}
    \fill[\col] (\xp-\bw, 0) rectangle (\xp+\bw, \bh);
    \node[above, font=\tiny] at (\xp, \bh) {\cnt};
    \node[below, rotate=45, anchor=north east, font=\tiny]
      at (\xp, -0.05) {\yr};
  }

  \foreach \v in {5,10,15}{
    \pgfmathsetmacro{\yg}{\v/\maxV*\H}
    \draw[gray!25, line width=0.3pt]
      (\mL-\bw-0.05, \yg) -- (\mL+\W+\bw, \yg);
    \node[left, font=\tiny] at (\mL-\bw-0.1, \yg) {\v};
  }
  \node[left, font=\tiny] at (\mL-\bw-0.1, 0) {0};

  \draw[line width=1.0pt]
    (\mL-\bw-0.05, \H) -- (\mL+\W+\bw, \H);
  \draw[line width=1.0pt]
    (\mL-\bw-0.05, 0)  -- (\mL+\W+\bw, 0);

  \node[rotate=90, font=\footnotesize, anchor=south]
    at (-0.1, \H/2) {Number of datasets};
\end{tikzpicture}
\end{minipage}

\caption{
\footnotesize Publication statistics of 78 surveyed datasets by publication venue (left) and year (right).
{\scriptsize
\linespread{0.7}\selectfont
         CVPR (IEEE/CVF Conf.\ on Computer Vision and Pattern Recognition),
         ICCV (IEEE/CVF Int.\ Conf.\ on Computer Vision),
         ECCV (European Conf.\ on Computer Vision),
         ACM MM (ACM Int.\ Conf.\ on Multimedia),
         ICCVW/CVPRW (ICCV/CVPR Workshops),
         ACCV (Asian Conf.\ on Computer Vision),
         AVSS (IEEE Int.\ Conf.\ on Advanced Video and Signal-based Surveillance),
         DICTA (Int.\ Conf.\ on Digital Image Computing),
         NeurIPS (Conf.\ on Neural Information Processing Systems),
         AAAI (AAAI Conf.\ on Artificial Intelligence),
         IJCNN (Int.\ Joint Conf.\ on Neural Networks),
         ICRA (IEEE Int.\ Conf.\ on Robotics and Automation),
         IROS (IEEE/RSJ Int.\ Conf.\ on Intelligent Robots and Systems),
         RA-L (IEEE Robotics and Automation Letters),
         ACRA (Australasian Conf.\ on Robotics and Automation),
         ITSC (IEEE Int.\ Conf.\ on Intelligent Transportation Systems),
         IV (IEEE Intelligent Vehicles Symposium),
         T-ITS (IEEE Trans.\ on Intelligent Transportation Systems),
         T-PAMI (IEEE Trans.\ on Pattern Analysis and Machine Intelligence),
         T-IP (IEEE Trans.\ on Image Processing),
         CVIU (Computer Vision and Image Understanding),
         Constr.\ Build.\ Mater.\ (Construction and Building Materials),
         Geosci.\ Data~J.\ (Geoscience Data Journal),
         IoT (IEEE Internet of Things Journal),
         EC3 (European Conf.\ on Computing in Construction).
}}
\label{fig:venue-year}
\end{figure}

\paragraph{\textbf{Research Gap 1: Unified Domain-specific Taxonomy.}}
`\textit{Anomaly}' is widely used across surveys, but its meaning varies substantially with the data modality and task definition. In image-based settings, it is often treated as category novelty introduced by unseen semantic classes (e.g. a lion in the driving lane) \cite{comment1paper4}. In contrast, video-based surveys frequently use the term to denote abnormal trajectories or behaviors revealed by temporal dynamics, such as traffic violations and irregular pedestrian behaviors reviewed in \cite{AnomalyDetectionSurvey}. More commonly, surveys may not even define `\textit{anomaly}' within a single field and instead use the term as a broad label for unusual activities. For instance, \cite{comment1paper3} examines the challenges introduced by camera motion across six general video-analysis tasks spanning maritime surveillance, general security, and urban transportation, while \cite{MSAD} surveys a wide range of anomalies captured by surveillance cameras in diverse scenarios (e.g., shop robberies and highway accidents).

More systematically, \cite{anomalysurvey} proposes a theoretical categorization to summarize anomalies in various data domains, but still places limited emphasis on traffic-specific phenomena and image-based perception. While this classic taxonomy is partially applicable to our study, such as a fallen tree blocking the road (single anomaly), a vehicle traveling at highway speed within the shared zone (contextual anomaly), and an abrupt four-lane cut (collective anomalies), its universal terminology can be too coarse for complex traffic scenarios. For example, a fallen tree is more naturally described as an \textit{obstacle} rather than an \textit{anomaly}, since its anomalousness arises from its atypical spatial configuration (lying across the road) rather than from category novelty (known semantic class, e.g. `vegetation').

Furthermore, existing anomaly-centric formulations provide inadequate coverage when the scope expands from `irregularities' to all `attention-worthy elements', because salient yet non-anomalous entities (e.g., surrounding road users), events (e.g., an ordinary cut-in in the adjacent lane), and conditions (e.g., a traffic light turning yellow) can be equally critical for maintaining road safety.

\paragraph{\textbf{Research Gap 2: Integrated Element-centric Review.}}
Unlike taxonomy-focused surveys, numerous reviews adopt an element-centric perspective but focus on a particular factor in traffic scenarios.
One line of work surveys perception sub-tasks tied to specific semantic categories. Road damage surveys review methods for detecting static pavement distress. For example, \cite{reviewer4suggestedpaper4} emphasizes repurposing sensors mounted on autonomous vehicles for pavement data collection, whereas \cite{RoadDamageSurvey} organizes the literature by data modalities and methodological choices. Vulnerable road users are similarly addressed in pedestrian detection surveys. For example, \cite{comment1paper2} focuses exclusively on low-light settings and highlights multi-spectral fusion of visible and thermal/infrared imagery, while \cite{9420291} provides a broader overview of the field evolution.
Another set of surveys treats traffic accidents as a distinct problem from traffic anomalies, focusing specifically on methodologies for accident detection and anticipation using dashcam and surveillance data \cite{trafficAccidentsSurvey}. 

However, these element-centric reviews remain largely siloed by contributing factor, limiting cross-domain visibility and, consequently, hindering the recognition and transfer of relevant tasks, datasets, and methodologies across areas. Establishing meaningful and systematic connections among these inherently related fields to support holistic safety reasoning remains therefore underexplored.

\paragraph{\textbf{Research Gap 3: Thorough Dataset Coverage.}}
Existing surveys predominantly emphasize algorithmic solutions and technical implementations \cite{comment1paper5}, without adequately addressing the fundamental question of what elements deserve attention in the driving environment by exploring existing data resources. Surveys that involve a dedicated analysis of datasets usually narrow the scope by constraining the observational viewpoint (e.g. static surveillance cameras in \cite{MSAD} versus moving perspectives in \cite{comment1paper3}) or by fixing the visual modality (e.g. video in \cite{AnomalyDetectionSurvey} versus single frame in \cite{comment1paper4}).

Although comprehensive surveys of autonomous-driving datasets exist, such as \cite{10509812}, they typically categorize datasets based on sensor types or high-level functional categories (e.g. perception, prediction, or planning) rather than analyzing their applicability across the range of perception tasks associated with attention-worthy elements in traffic scenes. Consequently, insufficient cross-analysis of datasets across different but related tasks may prevent researchers from fully leveraging existing resources, hindering both the novel applications and the development of a unified analytical framework that could bring together various attention-worthy elements in traffic scenes. 

\paragraph{\textbf{Contribution.}}
In response to these identified gaps, we further summarize our contributions below. 

\begin{enumerate}
    \item \textbf{\textit{A Novel Taxonomy}} that systematically categorizes tasks and datasets based on the attention-worthy elements they explore, including \textit{anomalies}, and \textit{pertinent entities}. This taxonomy bridges inherently related but historically isolated fields and establishes a common hierarchy for researchers within or across domains.
    \item \textbf{\textit{Comprehensive Survey}} covering 78 datasets for 40 vision-driven tasks, which comprehensively examines and illustrates the characteristics, advantages, and potential limitations of raw data and annotations. This comprehensive information could help researchers gain an overview of this domain, find appropriate resources to support their studies, and identify gaps that merit further investigation.
    \item \textbf{\textit{Cross-domain Analysis}} that examines how identical scene elements are interpreted and processed in various research contexts. It provides insights into the unification of domain standards and dataset cross-utilization, encouraging the maximization of existing resource utilities and the innovative applications of established datasets.
\end{enumerate}

\paragraph{\textbf{Navigation Guide.}} 
The remainder of this paper is organized as follows. Section \ref{sec:background} introduces fundamental computer vision tasks and terminologies. Each subsection in Section \ref{sec:tasks_datasets} elaborates one of eleven critical element categories, beginning with a conceptual overview, followed by an analysis of relevant tasks, and concluding with comprehensive examinations and visualizations of the associated datasets. Section \ref{sec:future} summarizes existing gaps and highlights future directions. Section \ref{sec:conclusion} concludes this article. Handy navigation is provided as follows. 
\begin{itemize}
  \item We recommend first reviewing Table \ref{table:Existing Tasks Classification} and Fig.~\ref{fig:graphic_taxonomy} for our taxonomy system used throughout the paper. 
  \item Readers primarily interested in a particular category of attention-worthy elements, task, or dataset may directly access the relevant subsection in Section~\ref{sec:tasks_datasets} via the navigation links embedded in the corresponding entries under the `Section', `Tasks', and `Typical Datasets' columns of Table~\ref{table:Existing Tasks Classification}.
  \item Readers dedicated to specific general topics (e.g., accident) can locate all relevant datasets by consulting the navigation given in Section \ref{subsec:generaltopics}. 
  \item Readers seeking summaries of the basic features of all available datasets can refer directly to Table \ref{table:visual_datasets_summary}.
  \item Readers seeking datasets with specific annotation levels (e.g., semantic masks) can consult Table \ref{table:visual_datasets_summary} for candidates by examining ticks on the corresponding `Level' column.
\end{itemize}

\section{Terminology and Fundamental Tasks}
\label{sec:background}
This section provides an overview of fundamental perception tasks and terminologies that underpin visual scene understanding in autonomous driving (AV). Its scope ranges from conventional two-dimensional (2D) image-based formulations, which interpret traffic scenes by extracting visual features from the image plane, to more recent three-dimensional (3D) and scene-level learning paradigms that further support geometric awareness, temporal consistency, and multimodal understanding.

\subsection{Image Classification}
\textbf{\textit{Task.}} Image classification represents one of the foundational tasks in computer vision, which focuses on answering the question `What is this image?'. It takes an image as input and assigns one of the predefined categories or labels to the entire image based on its content \cite{PANERU2021103940}. It can be divided into binary classification and multi-class classification. For instance, given an image of a car, a binary image classifier would output a label such as `Car' or `Not Car', while a multi-class classifier might categorize it among numerous possibilities like `Car', `Bus', or `Other Vehicles'. Originally, image classification was applied primarily to single-object images. Although it has since evolved to handle images containing multiple objects, it still focuses on a single element of interest, such as traffic congestion or weather, by characterizing scenes or events. 

\textbf{\textit{Methods.}} Modern deep-learning-based approaches typically employ convolutional neural networks (CNNs) with architectures such as visual geometry group (VGG) \cite{vgg} and residual network (ResNet) \cite{resnet}. 
These networks learn hierarchical feature representations from simple edges and textures in the early layers to complex object parts and eventually whole objects in deeper layers. These networks typically end with a softmax-based classification head that outputs class probabilities, and give a discrete class assignment as the final prediction. Recent models such as Vision Transformers (ViT) \cite{dosovitskiy2021an} further improved performance by capturing global context and long-range dependencies \cite{trilite}.
 
\textbf{\textit{Datasets.}} ImageNet \cite{ImageNet} remains the standard benchmark for large-scale nature image classification and backbone pretraining in most vision domains. In autonomous driving, datasets such as Berkeley DeepDrive 100K (BDD100K) \cite{BDD100K} can also support image-level classification tasks related to scene attributes. These datasets are discussed in detail in Sec.~\ref{sec:tasks_datasets}, where they are organized according to the critical scene elements underlying the classification task.

\textbf{\textit{Metrics.}} Accuracy measures the percentage of correctly classified images. Precision quantifies the proportion of true positive predictions to all positive predictions \cite{VOC}, which is especially informative when false positives are more costly than false negatives. Recall, conversely, examines the proportion of actual positives correctly identified \cite{VOC}, emphasizing the importance of minimizing false negatives. The F1-Score is the harmonic mean of precision and recall. In multi-class settings, mean average precision (mAP) and mean average recall (mAR) are essential strategies to adapt precision and recall \cite{Henderson2016EndtoEndTO}. The area under the receiver operating characteristic curve (AUC-ROC) \cite{ImageNet} and the area under the precision-recall curve (AUC-PR) offer threshold-independent assessments of classification performance.

\subsection{Object Localization}
\textbf{\textit{Tasks.}} Object localization extends the image classification task by not only recognizing which object or event is present in an image, but also determining where it is. It exclusively localizes a single instance of a particular category in an image \cite{7965889, cvl-umass2024}, which involves predicting a bounding box (typically defined by the coordinates for the top-left corner, width, and height) to enclose that instance \cite{cvl-umass2024}. This task is commonly approached as a regression problem, where the network outputs continuous values representing the coordinates.

\newpage
{\fontsize{7pt}{8pt}\selectfont
\setlength{\tabcolsep}{3.5pt}
\begin{longtable}{>{\centering\arraybackslash}m{1.1cm}
                  >{\centering\arraybackslash}m{1.5cm}
                  >{\centering\arraybackslash}m{5.7cm}
                  >{\centering\arraybackslash}m{3.7cm}
                  >{\centering\arraybackslash}m{0.55cm}
                  >{\arraybackslash}m{2.5cm}}

\caption[Classification of `Critical Elements That Demand Attention' for Road Safety]{%
Classification of `Critical Elements That Demand Attention' for Road Safety}
\label{table:Existing Tasks Classification} \\[-2pt]
\legendrownew{\textbf{App.}: Primary application layer served by each perception task based on their potential downstream role. P=\textit{perception}, Pr=\textit{prediction}, DS=\textit{decision support} for warning or supervision, and PS=\textit{planning support} for maneuvering and motion constraints.}\\[-5pt]
\toprule
Reason & Category & Justification & Tasks & App. & Typical Datasets \\
\midrule
\endfirsthead

\multicolumn{6}{l}{\textbf{Table \thetable{} (continued)}} \\
\toprule
Reason & Category & Justification & Tasks & App. & Typical Datasets \\
\midrule
\endhead

\multirow{58}{*}{Anomaly}
& \hyperref[subsec:anomaly_spatial]{\shortstack{Spatial \\ Location}} \par
&
Static obstacles on the road that impede normal traffic (e.g. force sudden break) or create collision risks
\par &
\begin{itemize}[leftmargin=*,noitemsep,topsep=0pt]
    \item \hyperref[task:obstacle_segmentation]{Obstacle Segmentation}
\end{itemize} \par
&
P \par
&
\shortstack[l]{
\hyperref[dataset:lostandfound]{LostAndFound} \cite{lostAndFound} \\[-1pt]
\hyperref[dataset:RoadObstacle21]{RoadObstacle21} \cite{chan2021segmentmeifyoucan}
} \par
\\[-\normalbaselineskip]
\cmidrule(lr){2-6}

& \hyperref[subsec:anomaly_semantic]{\shortstack{Semantic \\ Category}} \par
&
Previously unseen objects or regions that fall outside expected road-scene priors and can delay hazard appraisal and appropriate responses
\par &
\begin{itemize}[leftmargin=*,noitemsep,topsep=0pt]
   \item \hyperref[task:anomaly_segmentation]{Anomaly Segmentation}
\end{itemize}
&
P \par
&
\shortstack[l]{
\hyperref[dataset:LostAndFound]{LostAndFound} \cite{lostAndFound} \\[-1pt]
\hyperref[dataset:RoadAnomaly]{RoadAnomaly} \cite{RoadAnomalyDataset} \\[-1pt]
\hyperref[dataset:RoadAnomaly21]{RoadAnomaly21} \cite{chan2021segmentmeifyoucan} \\[-1pt]
\hyperref[dataset:fishy]{Fishyscapes} \cite{fishyscapes} \\[-1pt]
\hyperref[dataset:stu]{STU} \cite{nekrasov2025stu}
}\par
\\[-\normalbaselineskip]
\cmidrule(lr){2-6}

& \multirow{15}{*}{\hyperref[subsec:anomaly_event]{Event}} \par
&
Road construction that often introduces temporary conflicts, narrowed lanes, and unexpected merges \par
&
\begin{itemize}[leftmargin=*,noitemsep,topsep=0pt]
   \item \hyperref[task:road_construction]{Road Construction Detection}
\end{itemize}
&
P \par
&
- \par \\[-\normalbaselineskip]
\cmidrule(lr){3-6}

& &
Traffic congestion with reduced headways, frequent speed fluctuations, and dense interactions \par
&
\begin{itemize}[leftmargin=*,noitemsep,topsep=0pt]
   \item \hyperref[task:traffic_congestion]{Traffic Congestion Classification}
\end{itemize}
&
P \par
&

\shortstack[l]{
\hyperref[dataset:CCTRIB]{CCTRIB} \cite{CCTRIB} \\[-1pt]
\hyperref[dataset:UA-DETRAC]{UA-DETRAC} \cite{UA-DETRAC}
}\par
\\[-\normalbaselineskip]
\cmidrule(lr){3-6}

& &
\multirow{8}{=}{Traffic crashes that can obstruct traffic flow, create secondary hazards, and require rapid recognition for warning, avoidance, and incident management, including: \par
\begin{itemize}[leftmargin=2em,noitemsep,topsep=0pt]
   \item Types by participants, e.g., single- vehicle and vehicle-person crash
   \item Types by impact, e.g., head-on and rear-end collisions
\end{itemize}}
&
\begin{itemize}[leftmargin=*,noitemsep,topsep=0pt]
   \item \hyperref[task:accident_classification]{Traffic Accident Classification}
\end{itemize}
&
P \par
&

\shortstack[l]{
\hyperref[dataset:tadd]{TADD} \cite{TADD} \\[-1pt]
\hyperref[dataset:cadp]{CADP} \cite{CADPDataset}
}\par
\\[-\normalbaselineskip]
\cmidrule(lr){4-6}

& & &
\begin{itemize}[leftmargin=*,noitemsep,topsep=0pt]
   \item \hyperref[task:accident_anticipation]{Traffic Accident Anticipation}
\end{itemize} \par
&
Pr \par
&
 
\shortstack[l]{
\hyperref[dataset:CCD]{CCD} \cite{CarCrashDataset} \\[-1pt]
\hyperref[dataset:DAD]{DAD} \cite{DAD} \\[-1pt]
\hyperref[dataset:A3D]{A3D} \cite{A3D}
}\par
\\[-\normalbaselineskip]
\cmidrule(lr){4-6}

& & &
\begin{itemize}[leftmargin=*,noitemsep,topsep=0pt]
   \item \hyperref[task:accident_causality]{Accident Causality Recognition}
\end{itemize} \par
&
DS \par
&

\shortstack[l]{
\hyperref[dataset:DoTA]{DoTA} \cite{DoTA} \\[-1pt]
\hyperref[dataset:CTA]{CTA} \cite{you2020CTA} \\[-1pt]
\hyperref[dataset:MM-AU]{MM-AU} \cite{MM-AU}
}\par
\\[-\normalbaselineskip]
\cmidrule(lr){3-6}

& &
Criminal activities in traffic scenarios that can trigger abrupt maneuvers, distract nearby road users, or require emergency intervention altering traffic flow \par
&
\begin{itemize}[leftmargin=*,noitemsep,topsep=0pt]
   \item \hyperref[task:traffic_crime]{Traffic Crime Classification}
\end{itemize}
&
P \par
&
 
\shortstack[l]{
\hyperref[dataset:UCF-Crime]{UCF-Crime} \cite{Sultani_2018_CVPR} \\[-1pt]
\hyperref[dataset:MSAD]{MSAD} \cite{MSAD} \\[-1pt]
\hyperref[dataset:DAD]{DAD} \cite{10.1007/978-3-319-54190-7_9}
}\par
\\[-\normalbaselineskip]
\cmidrule(lr){2-6}

& \multirow{6}{*}{\hyperref[subsec:anomaly_kinematic]{\shortstack{Kinematic \\ Pattern}}} \par
&
\multirow{6}{=}{Non-standard spatial trajectories, suspicious behavior, and unexpected velocity transitions of road users that may signal imminent hazards and precede conflicts, reducing the time available for anticipation and evasive action} \par
&
\begin{itemize}[leftmargin=*,noitemsep,topsep=0pt]
   \item \hyperref[task:collision_prediction]{Collision Prediction}
\end{itemize} \par
&
Pr \par
&
 
\shortstack[l]{
\hyperref[dataset:DoTA]{DoTA} \cite{DoTA} \\[-1pt]
\hyperref[dataset:tadd]{TADD} \cite{TADD} \\[-1pt]
\hyperref[dataset:cadp]{CADP} \cite{CADPDataset} \\[-1pt]
\hyperref[dataset:CCD]{CCD} \cite{CarCrashDataset} \\[-1pt]
\hyperref[dataset:DAD]{DAD} \cite{DAD} \\[-1pt]
\hyperref[dataset:CST-S3D]{CST-S3D} \cite{CST-S3D}
}\par
\\[-\normalbaselineskip]
\cmidrule(lr){4-6}

& & &
\begin{itemize}[leftmargin=*,noitemsep,topsep=0pt]
   \item \hyperref[task:behavior_recognition]{Behavior Recognition}
\end{itemize} \par
&
Pr \par
&

\shortstack[l]{
\hyperref[dataset:YouTubeCrash]{YouTubeCrash} \cite{CrashToNotCrash} \\[-1pt]
\hyperref[dataset:GTACrash]{GTACrash} \cite{CrashToNotCrash}
}\par
\\[-\normalbaselineskip]
\cmidrule(lr){4-6}

& & &
\begin{itemize}[leftmargin=*,noitemsep,topsep=0pt]
   \item \hyperref[task:driver_attention]{Driver Attention Prediction}
\end{itemize} \par
&
Pr \par
&
 
\shortstack[l]{
\hyperref[dataset:bdda]{BDD-A} \cite{Xia2017PredictingDA} \\[-1pt]
\hyperref[dataset:dada2000]{DADA} \cite{DADA}
}\par
\\[-\normalbaselineskip]
\cmidrule(lr){2-6}

& \multirow{22}{*}{\hyperref[subsec:anomaly_status]{\shortstack{Status \\ or \\ Condition}}} \par
&
\multirow{10}{=}{Abnormal road conditions, such as potholes, cracks, protrusions, manholes, or unsurfaced roads that reduce tire-road stability, impair ride controllability, and can trigger loss of control, especially at speed or under braking}
&
\begin{itemize}[leftmargin=*,noitemsep,topsep=0pt]
   \item \hyperref[task:crack_segmentation]{Crack Segmentation}
\end{itemize}
&
P \par
&

\shortstack[l]{
\hyperref[dataset:CrackTree260]{CrackTree260} \cite{CrackTree} \\[-1pt]
\hyperref[dataset:CFD]{CFD} \cite{CrackForest} \\[-1pt]
\hyperref[dataset:CRACK500]{CRACK500} \cite{CRACK500} \\[-1pt]
\hyperref[dataset:GAPs384]{GAPs384} \cite{CRACK500} \\[-1pt]
\hyperref[dataset:EdmCrack600]{EdmCrack600} \cite{EdmCrack600} \\[-1pt]
\hyperref[dataset:NHA12D]{NHA12D} \cite{NHA12D}
}\par
\\[-\normalbaselineskip]
\cmidrule(lr){4-6}

& & &
\begin{itemize}[leftmargin=*,noitemsep,topsep=0pt]
   \item \hyperref[task:Pothole_Segmentation]{Pothole Segmentation}
\end{itemize}
&
P \par
&
 
\shortstack[l]{
\hyperref[dataset:Pothole-600]{Pothole-600} \cite{Pothole600}
}\par
\\[-\normalbaselineskip]
\cmidrule(lr){4-6}

& & &
\begin{itemize}[leftmargin=*,noitemsep,topsep=0pt]
   \item \hyperref[task:road_damage_classification]{Road Damage Classification}
\end{itemize}
&
P \par
&
 
\shortstack[l]{
\hyperref[dataset:CQU-BPDD]{CQU-BPDD} \cite{CQU-BPDD}
}\par
\\[-\normalbaselineskip]
\cmidrule(lr){4-6}

& & &
\begin{itemize}[leftmargin=*,noitemsep,topsep=0pt]
   \item \hyperref[task:road_damage_segmentation]{Road Damage Segmentation}
\end{itemize}
&
P \par
&
 
\shortstack[l]{
\hyperref[dataset:PotholeMix]{PotholeMix} \cite{SHREC} \\[-1pt]
\hyperref[dataset:M2S-RoAD]{M2S-RoAD} \cite{M2S-RoAD}
}\par
\\[-\normalbaselineskip]
\cmidrule(lr){4-6}

& & &
\begin{itemize}[leftmargin=*,noitemsep,topsep=0pt]
   \item \hyperref[task:road_damage_detection]{Road Damage Detection}
\end{itemize}
&
P \par
&
 
\shortstack[l]{
\hyperref[dataset:GAPs]{GAPs} \cite{ExtendGAPs} \\[-1pt]
\hyperref[dataset:rdd2022]{RDD2022} \cite{RDD2022}
}\par
\\[-\normalbaselineskip]
\cmidrule(lr){3-6}

& &
Abnormal weather conditions (e.g., rain, fog, snow) that degrade visibility, alter scene appearance, obscure cues, and worsen handling and stopping performance, increasing both perception and control risk \par
&
\begin{itemize}[leftmargin=*,noitemsep,topsep=0pt]
   \item \hyperref[task:Traffic_Weather_Recognition]{Traffic Weather Recognition}
\end{itemize}
&
P \par
&
 
\shortstack[l]{
\hyperref[dataset:ACDC]{ACDC} \cite{ACDC} \\[-1pt]
\hyperref[task:Traffic_Weather_Recognition]{RainCityscapes} \cite{Rainy1} \\[-1pt]
\hyperref[task:Traffic_Weather_Recognition]{Foggy Zurich} \cite{Foggy1} \\[-1pt]
\hyperref[task:Traffic_Weather_Recognition]{Foggy Cityscapes} \cite{Foggy2}
}\par
\\[-\normalbaselineskip]
\cmidrule(lr){3-6}

& &
Abnormal illumination condition (e.g., nighttime, sun glare, high beam from oncoming traffic) that degrade visibility, reduce contrast, impair object recognition, and delay detection of hazards, particularly vulnerable road users and traffic control devices \par
&
\begin{itemize}[leftmargin=*,noitemsep,topsep=0pt]
   \item \hyperref[task:illumination]{Illumination Recognition and Adaption}
\end{itemize}
&
P \par
&

\shortstack[l]{
\hyperref[dataset:ACDC]{ACDC} \cite{ACDC} \\[-1pt]
\hyperref[dataset:nightdriving]{NightDriving} \cite{NightDriving} \\[-1pt]
\hyperref[dataset:NightCity]{NightCity} \cite{nightcity} \\[-1pt]
\hyperref[dataset:darkzirich]{DarkZurich} \cite{darkzurich} \\[-1pt]
\hyperref[dataset:nightowls]{NightOwls} \cite{NightOwls} \\[-1pt]
\hyperref[dataset:GLARE]{GLARE} \cite{Glare}
}\par
\\[-\normalbaselineskip]
\cmidrule(lr){2-6}

& \hyperref[subsec:anomaly_appearance]{Appearance} \par
&
\begin{itemize}[leftmargin=*,noitemsep,topsep=0pt]
   \item Abnormal visual salience that may cause distraction while can also improve early hazard detection
   \item Suspicious or threat-associated appearance may alter driving behavior and public-security response
\end{itemize}
&
\begin{itemize}[leftmargin=*,noitemsep,topsep=0pt]
   \item \hyperref[task:criminal_recognition]{Criminal Recognition}
   \item \hyperref[task:suspicious_veh_recognition]{Suspicious Vehicle Recognition}
   \item \hyperref[task:damaged_veh_detection]{Damaged Vehicle Detection}
\end{itemize}
&
DS \par
&
- \par \\[-\normalbaselineskip]
\cmidrule(lr){2-6}

& \hyperref[subsec:anomaly_multiple]{\shortstack{Multiple \\ Reasons}} \par
&
Multi-factor anomalies can amplify uncertainty and conflict risk beyond single anomaly would imply alone \par
&
\begin{itemize}[leftmargin=*,noitemsep,topsep=0pt]
   \item \hyperref[task:risk_identification]{Risk Identification}
\end{itemize}
&
DS \par
&
\shortstack[l]{
\hyperref[dataset:riskbench]{RiskBench} \cite{RiskBench}
}
\par 
\\[-\normalbaselineskip]
\midrule

\multirow{34}{*}{\parbox[c]{1cm}{\centering Pertinent \\ Entity}}
& \multirow{7}{*}{\hyperref[subsec:Pertinency_Location]{\shortstack{Spatial \\ Location}}} \par
&
\multirow{7}{=}{Spatially relevant entities whose location relative to the ego vehicle may constrain immediate actions such as yielding, braking, turning, or lane keeping:
\begin{itemize}[leftmargin=*,noitemsep,topsep=0pt]
   \item Surrounding road users
   \item Road users possessing right-of-way
   \item Specific places (e.g., parking)
\end{itemize}} \par
&
\begin{itemize}[leftmargin=*,noitemsep,topsep=0pt]
   \item \hyperref[task:pedestrian_intention]{Pedestrian Intention Prediction}
\end{itemize} \par
&
Pr \par
&

\shortstack[l]{
\hyperref[dataset:jaad]{JAAD} \cite{JAAD} \\[-1pt]
\hyperref[dataset:PIE]{PIE} \cite{PIE} \\[-1pt]
\hyperref[dataset:STIP]{STIP} \cite{STIP}
}\par
\\[-\normalbaselineskip]
\cmidrule(lr){4-6}

& & &
\begin{itemize}[leftmargin=*,noitemsep,topsep=0pt]
   \item \hyperref[task:pedestrian_pose]{Pedestrian Pose Estimation}
\end{itemize} \par
&
P \par
&
 
\shortstack[l]{
\hyperref[dataset:PedX]{PedX} \cite{pedx} \\[-1pt]
\hyperref[dataset:waymo_pose]{Waymo} \cite{Waymo} \\[-1pt]
\hyperref[dataset:ECPDP]{ECPDP} \cite{ECPDP}
}\par
\\[-\normalbaselineskip]
\cmidrule(lr){4-6}

& & &
\begin{itemize}[leftmargin=*,noitemsep,topsep=0pt]
   \item \hyperref[task:scene_classification]{Scene Classification}
\end{itemize} \par
&
P \par
&
 
\shortstack[l]{
\hyperref[dataset:carla-syn]{Carla-Syn} \cite{CarlaSyn} \\[-1pt]
\hyperref[dataset:sub-NUDrive]{sub-NUDrive} \cite{sub-NUDrive}
}\par
\\[-\normalbaselineskip]
\cmidrule(lr){2-6}

& \multirow{24}{*}{\hyperref[subsec:pertinency_semantic]{\shortstack{Semantic \\ Category}}} \par
&
\multirow{7}{=}{Vulnerable road users with low physical protection and less predictable motion demanding earlier detection, larger safety margins, and more conservative interaction planning, e.g.,
\begin{itemize}[leftmargin=2em,noitemsep,topsep=0pt]
   \item Pedestrian
   \item Riders
\end{itemize}} \par
&
\begin{itemize}[leftmargin=*,noitemsep,topsep=0pt]
   \item \hyperref[task:pedestrian_detection]{Pedestrian Detection}
\end{itemize}
&
P \par
&
 
\shortstack[l]{
\hyperref[dataset:Caltech]{Caltech} \cite{CALTECH} \\[-1pt]
\hyperref[dataset:nightowls]{NightOwls} \cite{NightOwls} \\[-1pt]
\hyperref[dataset:CityPersons]{CityPersons} \cite{CityPersons} \\[-1pt]
\hyperref[dataset:ECP]{EuroCity} \cite{EuroCity} \\[-1pt]
\hyperref[dataset:llvip]{LLVIP} \cite{LLVIP}
}\par
\\[-\normalbaselineskip]
\cmidrule(lr){4-6}

& & &
\begin{itemize}[leftmargin=*,noitemsep,topsep=0pt]
   \item \hyperref[task:Cyclist_detection]{Cyclist Detection}
\end{itemize} \par
&
P \par
&

\shortstack[l]{
\hyperref[dataset:CityPersons]{CityPersons} \cite{CityPersons} \\[-1pt]
\hyperref[dataset:ECP]{EuroCity} \cite{EuroCity} \\[-1pt]
\hyperref[dataset:tdc]{TDC} \cite{cyclistdetection}
}\par
\\[-\normalbaselineskip]
\cmidrule(lr){3-6}

& &
\multirow{8}{=}{Road-layout entities that define the legal and physically safe space for motion, thereby governing lane keeping, path planning, and conflict avoidance.
\begin{itemize}[leftmargin=*,noitemsep,topsep=0pt]
   \item Drivable areas
   \item Lanes
   \item Lanemarks
\end{itemize}} \par
&
\begin{itemize}[leftmargin=*,noitemsep,topsep=0pt]
   \item \hyperref[task:2d_lane_detection]{2D Lane Detection}
\end{itemize}
&
PS \par
&
 
\shortstack[l]{
\hyperref[dataset:tusimple]{TuSimple} \cite{TuSimpleGit} \\[-1pt]
\hyperref[dataset:CULane]{CULane} \cite{CULane} \\[-1pt]
\hyperref[dataset:BDD100K]{BDD100K} \cite{BDD100K} \\[-1pt]
\hyperref[dataset:VIL-100]{VIL-100} \cite{VIL100} \\[-1pt]
\hyperref[dataset:OpenLane-V]{OpenLane} \cite{openlane}
}\par
\\[-\normalbaselineskip]
\cmidrule(lr){4-6}

& & &
\begin{itemize}[leftmargin=*,noitemsep,topsep=0pt]
   \item \hyperref[task:lanemark]{2D Lanemark Segmentation}
\end{itemize} \par
&
PS \par
&
 
\shortstack[l]{
\hyperref[dataset:ApolloScape]{ApolloScape} \cite{ApolloScape}
}\par
\\[-\normalbaselineskip]
\cmidrule(lr){4-6}

& & &
\begin{itemize}[leftmargin=*,noitemsep,topsep=0pt]
   \item \hyperref[task:3d_lane]{3D Lane Detection}
\end{itemize} \par
&
PS \par
&
 
\shortstack[l]{
\hyperref[dataset:openlane]{OpenLane-V} \cite{openlane-v}
}\par
\\[-\normalbaselineskip]
\cmidrule(lr){4-6}

& & &
\begin{itemize}[leftmargin=*,noitemsep,topsep=0pt]
   \item \hyperref[task:hd_map]{HD Map Construction}
\end{itemize} \par
&
PS \par
&
\shortstack[l]{
\hyperref[dataset:nuScenes]{nuScenes} \cite{nuScenes} \\[-1pt]
\hyperref[dataset:Argoverse2]{Argoverse2} \cite{Argoverse2}
}\par
\\[-\normalbaselineskip]
\cmidrule(lr){3-6}

& &
\multirow{4}{=}{Traffic control devices that regulate priority, speed, and permitted movements, failure to detect which can directly produce rule violations and collision risk e.g., traffic sign, speed bumps}
\par
&
\begin{itemize}[leftmargin=*,noitemsep,topsep=0pt]
   \item \hyperref[task:sign_detection]{Traffic Sign Detection}
\end{itemize}
&
PS \par
&

\shortstack[l]{
\hyperref[dataset:DFG]{DFG} \cite{DFG} \\[-1pt]
\hyperref[dataset:TT100K]{TT100K} \cite{Tencent2016} \\[-1pt]
\hyperref[dataset:GTSDB]{GTSDB} \cite{GTSDB2020} \\[-1pt]
\hyperref[dataset:GLARE]{GLARE} \cite{Glare}
}\par
\\[-\normalbaselineskip]
\cmidrule(lr){4-6}

& & &
\begin{itemize}[leftmargin=*,noitemsep,topsep=0pt]
   \item \hyperref[task:sign_graph]{Traffic Sign Graph Generation}
\end{itemize} \par
&
PS \par
&
 
\shortstack[l]{
\hyperref[dataset:CTSU]{CTSU} \cite{CTSU} \\[-1pt]
\hyperref[dataset:RS10K]{RS10K} \cite{RS10K}
}\par
\\[-\normalbaselineskip]
\cmidrule(lr){2-6}

& \multirow{5}{*}{\hyperref[subsec:pertinency_status]{\shortstack{Status or \\ Condition}}} \par
&
\multirow{5}{=}{Temporary states, such as signal phase, gesture instruction, or momentary salience, determine the current operational meaning of an entity and the urgency of response.}
\par
&
\begin{itemize}[leftmargin=*,noitemsep,topsep=0pt]
   \item \hyperref[task:light_detection]{Traffic Light Detection}
\end{itemize} \par
&
PS \par
&

\shortstack[l]{
\hyperref[dataset:LISA]{LISA} \cite{LISA} \\[-1pt]
\hyperref[dataset:BSTLD]{BSTLD} \cite{BOSCH} \\[-1pt]
\hyperref[dataset:DriveU]{DriveU} \cite{DriveU} \\[-1pt]
\hyperref[dataset:S2TLD]{$\text{S}^2\text{TLD}$} \cite{s2tld}
}\par
\\[-\normalbaselineskip]
\cmidrule(lr){4-6}

& & &
\begin{itemize}[leftmargin=*,noitemsep,topsep=0pt]
   \item \hyperref[task:police_gesture]{Traffic Police Gesture Recognition}
\end{itemize} \par
&
PS \par
&

\shortstack[l]{
\hyperref[dataset:tpgr]{TPGR} \cite{HE2020248}
}\par
\\[-\normalbaselineskip]
\cmidrule(lr){4-6}

& & &
\begin{itemize}[leftmargin=*,noitemsep,topsep=0pt]
   \item \hyperref[task:traffic_salient]{Traffic Salient Object Detection}
\end{itemize}
&
P \par
&
- \par \\[-\normalbaselineskip]
\cmidrule(lr){2-6}

& \multirow{9}{*}{\hyperref[subsec:pertinency_kinematic]{\shortstack{Kinematic \\ Pattern}}} \par
&
\multirow{9}{=}{Ongoing spatial trajectory or motion behavior of other agents determines near-future conflicts, time-to-collision, and the need for braking, yielding, or evasive maneuvers, e.g.,
\begin{itemize}[leftmargin=*,noitemsep,topsep=0pt]
   \item A cut-in of adjacent vehicle
   \item A braking of leading vehicle
\end{itemize}} \par
&
\begin{itemize}[leftmargin=*,noitemsep,topsep=0pt]
   \item \hyperref[task:lane_change_classification]{Lane Change Classification}
\end{itemize} \par
&
PS \par
&
\shortstack[l]{
\hyperref[dataset:PREVENTION]{PREVENTION} \cite{prevention}
}\par
\\[-\normalbaselineskip]
\cmidrule(lr){4-6}

& & &
\begin{itemize}[leftmargin=*,noitemsep,topsep=0pt]
   \item \hyperref[task:lane_change_prediction]{Lane Change Prediction}
\end{itemize} \par
&
Pr \par
&
\shortstack[l]{
\hyperref[dataset:PREVENTION]{PREVENTION} \cite{prevention}
}\par
\\[-\normalbaselineskip]
\cmidrule(lr){4-6}

& & &
\begin{itemize}[leftmargin=*,noitemsep,topsep=0pt]
   \item \hyperref[task:driving_maneuvers_classification]{Driving Maneuver Classification}
\end{itemize}\par
&
PS \par
&
\shortstack[l]{
\hyperref[dataset:BLVD]{BLVD} \cite{BLVD}
}\par
\\[-\normalbaselineskip]
\cmidrule(lr){4-6}

& & &
\begin{itemize}[leftmargin=*,noitemsep,topsep=0pt]
   \item \hyperref[task:trajectory_prediction]{\shortstack{Trajectory Prediction \\ (Motion Forecasting)}}
\end{itemize}\par
&
Pr \par
&
\shortstack[l]{
\hyperref[dataset:nuScenes]{nuScenes} \cite{nuScenes} \\[-1pt]
\hyperref[dataset:Argoverse2]{Argoverse2}-motion \cite{Argoverse2} \\[-1pt]
\hyperref[dataset:waymo_pose]{Waymo}-motion \cite{waymo_motion}
}\par
\\[-\normalbaselineskip]
\bottomrule

\end{longtable}
}

\textbf{\textit{Methods.}} Mainstream methods augment image classification backbones with regression or activation-based localization heads to predict the position of the target object. Early approaches mainly relied on CNN-based architectures and class activation mapping mechanisms to localize the dominant object from image-level supervision. More recent studies often formulate the problem as weakly supervised object localization (WSOL) \cite{wsol}, where transformer-based and vision-language-based models have become increasingly prominent, such as ViTOL \cite{vitol} and recent TriLite \cite{trilite}.

\textbf{\textit{Datasets.}} The ImageNet Large Scale Visual Recognition Challenge (ILSVRC) \cite{ILSVRC} remains the most classic benchmark for object localization. In traffic scenario, despite the scarcity of dedicated benchmarks, several datasets still contain images annotated with a single bounding box for the object of interest, which will be detailed in Sec.~\ref{sec:tasks_datasets}.

\textbf{\textit{Metrics.}} The primary metric is Intersection over Union (IoU), which measures the overlap between the predicted bounding box and the ground truth \cite{VOC}. A prediction is typically considered correct if the IoU exceeds a threshold (commonly 0.5) \cite{VOC} and the class label matches the ground truth. The localization average precision (AP) interpolates the classic AP to evaluate both classification and localization performance by summarizing AUC-PR \cite{VOC}.

\subsection{Object Detection}
\label{subsec:objectdetection}
\textbf{\textit{Task.}} Object detection generalizes object localization to handle multiple objects within a single image \cite{7965889}. This task requires systems to localize each individual instance of interest present in the image using a bounding box and classify each instance into one of the predefined categories \cite{MaskRCNN}. 

\textbf{\textit{Methods.}} Modern frameworks fall into two primary types: two-stage detectors such as Faster Region-based CNN (Faster R-CNN) \cite{fasterrcnn}, which first generates region proposals and then classifies them; single-stage detectors such as You Only Look Once (YOLO) \cite{yolo} and Single Shot Multibox Detector (SSD) \cite{ssd}, which predict object classes and bounding boxes in one pass. Recent innovations include anchor-free approaches such as Fully Convolutional One-Stage Object Detection (FCOS) \cite{fcos} and transformer-based methods such as DEtection TRansformer (DETR) \cite{DETR} and its successor \cite{rf-detr}, which reformulate detection as a set prediction problem, without relying on hand-crafted components such as non-maximum suppression \cite{DETR}. 

\textbf{\textit{Datasets.}} PASCAL Visual Object Classes (PASCAL VOC) \cite{PASCAL} and Microsoft Common Objects in Context (MS COCO) \cite{COCO} are the most influential benchmarks, with the latter now serving as the dominant modern standard. In traffic scenarios, Cityscapes \cite{Cityscapes} and BDD100K \cite{BDD100K} are widely used for general object detection, while more datasets focus on specialized categories, such as anomalies and lane marks.

\textbf{\textit{Metrics.}} mAP averages the precision-recall curve across all classes, providing a comprehensive performance measure \cite{fastrcnn}. Meanwhile, mAP and mAR variations calculated across multiple IoU thresholds for all categories and mAP variations designed separately for small, medium, and large objects examine robustness in handling scale variation \cite{COCO}. In addition, frames per second (FPS) quantifies inference speed, balancing accuracy metrics with real-world deployment considerations for time-sensitive applications \cite{yolo}.

\subsection{Semantic Segmentation}
\label{subsec:semantic_segmentation}
\textbf{\textit{Task.}} Beyond the coarser spatial representation achieved by detection tasks, segmentation tasks provide a pixel-level understanding of images. Semantic segmentation further extends the area of interest from merely foreground objects to the entire image, where each pixel is classified according to the object category it belongs to \cite{PanopticSegmentation}. Different from detection tasks that output bounding-box coordinates, semantic segmentation delineates the boundaries of all scene entities by generating a dense classification map with the same dimensions as the input image. However, semantic segmentation does not distinguish among instances in the same semantic class, instead treating them as a single region. This limitation is particularly restrictive in traffic scenarios, where individual instances often have distinct significance.

\textbf{\textit{Methods.}} Fully Convolutional Networks (FCNs) \cite{FCN}, U-Net \cite{u-net}, and DeepLab \cite{deeplab} established the foundation of semantic segmentation by employing techniques like dilated convolutions and skip connections to preserve spatial resolution while capturing multi-scale contextual information. More recent approaches increasingly adopt transformer-based architectures, such as SEgmentation TRansformer (SETR) \cite{setr} that reformulates semantic segmentation as a sequence-to-sequence prediction problem, and SegFormer that emphasizes an efficient hierarchical transformer with a multilayer perceptron (MLP) decoder \cite{segformer}.

\textbf{\textit{Datasets.}} Cityscapes \cite{Cityscapes} is the most widely used benchmark for urban driving scenes, featuring high-resolution images captured with a stereo camera, while Mapillary Vistas \cite{Mapillary} offers a more geographically diverse collection of street-level imagery gathered across six continents using various devices. BDD100K \cite{BDD100K} is designed as a large-scale, multi-task driving benchmark built from iPhone-captured driving sequences. All three datasets support semantic, instance, and panoptic segmentation, and are therefore not discussed redundantly in Sec.~\ref{subsec:instance_segmentation} and Sec.~\ref{subsec:panoptic_segmentation}.

\textbf{\textit{Metrics.}} Mean IoU (mIoU) serves as the main metric by averaging the pixel-level IoU across all classes \cite{COCO}. The frequency-weighted IoU addresses the class imbalance by weighing the IoU of each class by its pixel frequency \cite{COCO}. Pixel accuracy (pAcc) denotes the fraction of correct pixels \cite{COCO}, while mean accuracy (mAcc) first calculates pAcc per class before averaging over all classes \cite{COCO}.

\subsection{Instance Segmentation}
\label{subsec:instance_segmentation}
\textbf{\textit{Task.}} Instance segmentation maintains pixel-level precision while refocusing attention on foreground objects rather than the entire image. It synthesizes elements of object detection and semantic segmentation by identifying individual instances with pixel-level masks \cite{BharathECCV2014}. Instance segmentation differentiates between separate instances of the same class (e.g., distinguishing between different people in a crowd) \cite{BharathECCV2014}.

\textbf{\textit{Methods.}} Mask R-CNN \cite{MaskRCNN} remains the benchmark baseline, extending Faster R-CNN \cite{fasterrcnn} with a mask prediction branch in parallel. Segmenting Objects by Locations (SOLO) \cite{SOLO} and Point-based Rendering (PointRend) \cite{PointRend} represent alternative approaches that address the instance segmentation problem by focusing on location-based instance discrimination and adaptive point-based refinement, respectively. More recent benchmark methods increasingly adopt end-to-end and transformer-based formulations, with Segmenting Objects by Learning Queries (SOLQ) \cite{solq}, Mask2Former \cite{mask2former}, and Mask DINO \cite{maskdino} serving as representative examples, while FastInst \cite{FastInst} further emphasizes real-time efficiency.

\textbf{\textit{Metrics.}} Similar to object detection, instance segmentation mainly uses mAP and its variations for evaluation, but IoU is calculated based on masks rather than bounding boxes \cite{BharathECCV2014}.

\subsection{Panoptic Segmentation}
\label{subsec:panoptic_segmentation}
\textbf{\textit{Task.}} Panoptic segmentation redirects attention to the complete image, by unifying semantic segmentation and instance segmentation into a coherent framework \cite{PanopticSegmentation}. It formally divides image pixels into two groups, including `stuff' (amorphous background regions such as sky or grass) and `thing' (countable foreground objects such as pedestrians or cars) \cite{PanopticSegmentation}. Panoptic segmentation requires the system to assign each pixel a semantic label and, for pixels belonging to the `thing' classes, to additionally assign an instance identifier (ID) \cite{PanopticSegmentation}. 

\textbf{\textit{Methods.}} Early architectures, such as Panoptic Feature Pyramid Network (FPN) \cite{PFPN}, extend Mask RCNN \cite{MaskRCNN} with a semantic segmentation branch and merge the output from both branches. 
Another taxonomy classifies panoptic segmentation approaches into bottom-up (starting with pixel-level features and grouping them into segments such as Panoptic-DeepLab \cite{panoptic-deeplab}) and top-down methods (beginning with proposals that are refined into detailed masks such as Efficient Panoptic Segmentation Network (EPSNet) \cite{EPSNet}).
More recent models, especially unified transformer-based algorithms such as Mask2Former \cite{mask2former} and oneFormer \cite{oneformer} have further advanced performance by providing a general formulation for dense image segmentation.

\textbf{\textit{Metrics.}} Panoptic quality (PQ) is the main metric, which evaluates performance considering both segmentation accuracy and recognition quality, providing a holistic assessment of scene understanding capabilities \cite{PanopticSegmentation}. In addition to the classic instance-based PQ, the region-based Parsing Covering (PC) metric \cite{Yang2019DeeperLabSI} is designed to better capture image parsing quality for `stuff' classes and larger object instances.

\subsection{Scene Graph Generation}
\textbf{\textit{Task.}} Scene graph generation (SGG) extends object detection to understand and represent the relationships between objects \cite{xu2017scenegraph}. It transforms an image into a graph-structured representation in which nodes represent localized and categorized objects, and edges represent their relationships \cite{SGG}. 

\textbf{\textit{Methods.}} Traditional methods generally follow a two-stage pipeline, classifying objects and relationships separately \cite{Zareian_2020_ECCV}. More recent studies have shifted toward end-to-end algorithms, jointly optimizing object detection and relation classification. For example, Bipartite Graph Network (BGNN) develops the bipartite graph to compute a context-aware representation of entities and predicates \cite{BGNN}. Transformers further advance the development of SGG and gradually become the dominant direction, with representatives such as Relation Transformer (RelTR) \cite{Reltr} that directly predicts (subject, predicate, object) triplets and more recent Extracting Graph from Transformer (EGTR) \cite{EGTR} towards lightweight one-stage frameworks. Meanwhile, another important line of work incorporates knowledge as additional references to alleviate data bias and enhance semantic consistency. Early examples such as Graph Bridging Network (GB-Net) \cite{Zareian_2020_ECCV} introduce commonsense graphs for pairwise matching, while more recent methods, such as Hierarchical Knowledge Enhanced Robust SGG (HiKER-SGG) \cite{hiker}, perform hierarchical reasoning over external knowledge structures to improve robustness. Beyond closed-set prediction, recent studies have also begun to explore open-vocabulary SGG (OVSGG) by leveraging knowledge transferred from large vision-language models \cite{ACC}.

\textbf{\textit{Datasets.}} Visual Genome \cite{VG} remains the dominant benchmark for SGG, while Open Images dataset \cite{OpenImages} is also widely adopted, particularly for large-scale visual relationship detection. In addition, GQA \cite{GQA} also provides scene graph annotations and can therefore serve as an auxiliary resource.

\textbf{\textit{Metrics.}}
Recall@K (R@K) measures the percentage of true relationships appearing in top K predictions \cite{lu2016visual}, while Mean Recall@K (mR@K) calculates R@K separately for each relationship class and then averages across classes to mitigate bias towards frequent relationships \cite{8954048}. SGG models are usually evaluated in three settings: 1) predicate classification (predicting predicates of the given pairs of localized objects), 2) scene graph classification (predicting predicates and categories of given localized objects), and 3) the entire SGG pipeline on the given image \cite{xu2017scenegraph}.

\subsection{3D Object Detection}
\label{subsec:3Dobjectdetection}
\textbf{\textit{Task.}} Different from 2D object detection, which computes axis-aligned bounding boxes from input images, 3D object detection localizes each instance of the target classes using 3D rotated bounding boxes \cite{comment8paper1}. It offers a more explicit representation of object geometry and spatial occupancy. Specifically, the 3D bounding box encodes the position and dimensions of the object, while its orientation is typically represented by a heading angle \cite{comment8paper2}.

\textbf{\textit{Methods.}} The two most widely adopted data modalities are images and point clouds \cite{comment8paper2}. Accordingly, existing methods are generally classified into three categories \cite{comment8paper3}. 1) \textit{Image-based methods} either project 2D detections into 3D space via template matching or geometric constraints, or lift 2D features using pseudo-LiDAR representations \cite{comment8paper2}. Recent representative methods are primarily monocular (e.g. MonoDGP \cite{monodgp}) and multi-view transformer-based models (e.g. PETRv2 \cite{petrv2}). 2) \textit{Point-cloud-based algorithms} fall into three subtypes. Point-based methods operate directly on raw point clouds and thus preserve fine-grained geometric information \cite{comment8paper2}, while voxel-based methods discretize space into regular cells to enable efficient 3D convolutional processing, at the expense of some fine-grained patterns \cite{comment8paper3}. Point-voxel-based methods are designed to leverage the strengths of both approaches. Among the benchmarks, CenterPoint \cite{comment8paper1} remains the most established baseline, while VoxelNeXt \cite{voxenext} is a newer notable representative. 3) \textit{Fusion-based strategies} address this task by integrating image and point-cloud information, sequentially or in parallel \cite{comment8paper2}. Among representative benchmarks, BEVFusion \cite{bevfusion}, which projects multi-modal features into a unified bird's-eye-view (BEV) representation, serves as a widely adopted baseline, while more recent methods such as IS-Fusion \cite{isfusion} have further advanced performance on the nuScenes dataset \cite{nuScenes}.

\textbf{\textit{Datasets.}} The KITTI Vision Benchmark Suite (KITTI) \cite{KITTI} remains widely used for evaluating monocular and stereo image-based methods, whereas nuScenes \cite{nuScenes}, Argoverse 2 \cite{Argoverse2}, and the Waymo Open Dataset \cite{Waymo} are commonly used to benchmark camera-only, LiDAR-only, and fusion-based methods.

\textbf{\textit{Metrics.}} Consistent with 2D object detection, mAP remains the primary metric \cite{comment8paper2} to evaluate performance in both the 3D space and the BEV \cite{monodgp}. Meanwhile, each benchmark dataset employs its own official evaluation metric. For example, KITTI \cite{KITTI} uses the standard interpolated AP, $AP|_{R_N}$, defined as the mean precision over a recall set $R$ composed of $N$ evenly spaced recall levels \cite{comment8paper2}. In contrast, nuScenes \cite{nuScenes} proposes a NuScenes Detection Score (NDS), a weighted average of mAP and mean average errors of the set $\varepsilon$.

\subsection{3D Occupancy Prediction}
\label{subsec:occupancyprediction}
\textbf{\textit{Task.}} The goal of 3D occupancy prediction is to determine whether each voxel in the scene is occupied by a specific semantic category \cite{ProtoOcc}. Compared with 3D bounding boxes, which simplify object shapes, 3D occupancy prediction preserves richer geometry details by discretizing scene elements into structured volumetric cells \cite{occnet}. Derived from mobile robot navigation task, occupancy prediction was originally designed for indoor scenes \cite{occsurvey}. Nevertheless, it has recently received growing attention with the emergence of a pioneering study \cite{monoscene} on outdoor scene understanding \cite{occsurvey}. This dense semantic and spatial representation enables autonomous vehicles to reason about visible structures as well as occluded regions, leading to a more complete understanding of the surrounding environment. Such fine-grained scene modeling can further benefit downstream decision-making tasks, including planning and navigation \cite{occnet}.

\textbf{\textit{Methods.}} The pioneering work, MonoScene \cite{monoscene}, explores the prediction of outdoor occupancy using only a monocular camera \cite{occsurvey}. Subsequent vision-centric methods, such as Tri-Perspective View Transformer (TPVFormer) \cite{TPVFormer}, exploit multi-view features captured by multiple cameras mounted around the vehicle. More recent benchmarks have increasingly adopted multi-modal paradigms. In the context of camera–radar fusion, HyDRa \cite{HyDRa} serves as a representative approach, projecting the sparse radar point cloud onto surround-view visual features and employing a radar-weighted backward projection \cite{OccCylindrical}. Another major fusion paradigm combines surround-view cameras with LiDAR. A representative example is BEVFusion \cite{bevfusion}, which encodes multi-modal inputs into a unified BEV representation for feature fusion. OccFusion \cite{OccFusion} further extends this trend by integrating features from surround-view cameras, LiDAR, and radar. 

\textbf{\textit{Datasets.}}
The large-scale datasets, nuScenes \cite{nuScenes} and Waymo Open \cite{Waymo}, provide multi-modal and geometry-aware data foundations for constructing high-quality occupancy annotations. Built upon these datasets, Occ3D-nuScenes and Occ3D-Waymo \cite{Occ3D} introduce visibility-aware occupancy labels, while SurroundOcc \cite{SurroundOcc} and OpenOccupancy \cite{OpenOccupancy} provide dense semantic occupancy annotations. These datasets have become widely adopted benchmarks.

\textbf{\textit{Metrics.}} Similar to 2D semantic segmentation, 3D occupancy prediction performance is typically evaluated using IoU and mIoU, which quantify class-specific prediction accuracy at the voxel level and summarize overall semantic occupancy performance across categories, respectively.

\newpage

{\scriptsize
\setlength{\tabcolsep}{2pt}
\begin{longtable}{F c c c c c c c c c c c c c c c c c c}
\caption[Summary of Vision-based datasets for road safety]{%
Summary of Vision-based Scene-understanding Datasets of Anomalies and Pertinent Entities for Road Safety}
\label{table:visual_datasets_summary}\\[-4pt]
\legendrow{\textbf{Region}: DE=\textit{Germany}, CH=\textit{Switzerland}, CN=\textit{China}, US=\textit{United States}, CA=\textit{Canada}, AU=\textit{Australia}, JP=\textit{Japan}, FR=\textit{France}, NL=\textit{Netherlands}, UK=\textit{the United Kingdom}, SG=\textit{Singapore}, SI=\textit{Slovenia}, ES=\textit{Spain}, --=\textit{Unspecified or Inapplicable}.}
\legendrow{\textbf{View} (\textit{the perspective of footage}): Eg=\textit{ego view}, Sv=\textit{surveillance view}, Lc=\textit{centered on certain area}.}
\legendrow{\textbf{Source} (\textit{data is collected by/from}):  Au=\textit{dataset authors},  Wb=\textit{web platforms}, [\textit{int}]=\textit{the citation dataset},  Ds=\textit{multiple existing datasets}.}
\legendrow{\textbf{Sensor} (\textit{data is collected using}): Cam=\textit{camera} (st=\textit{stereo camera}, dc=\textit{dashcam}, sc=\textit{surveillance camera}, tc=\textit{traffic surveillance camera}, mc=\textit{mobile phone camera}, ptz=\textit{{\scriptsize Pan-Tilt-Zoom} camera},  \scriptsize GoPro, mono=\textit{monochrome camera}, ref=\textit{reflex camera}, \checkmark=\textit{unknown or monocular camera}), and/or Oth=\textit{other} (sim=\textit{simulator}, syn=\textit{synthesizing}, eye=\textit{eye tracker},  Li=\textit{\scriptsize LiDAR}, ra=\textit{radar}, imu=\textit{inertial measurement unit}, {\scriptsize GV}=\textit{\scriptsize GoogleView}, {\scriptsize GPS}).}
\legendrow{\textbf{Level} (\textit{how critical elements are addressed}): C=\textit{image classification}, L=\textit{object localization}, D=\textit{object detection}, S=\textit{semantic segmentation}, I=\textit{instance segmentation}, and P=\textit{panoptic segmentation}.}
\legendrow{\textbf{Scene} (\textit{scenarios captured in footage}): A=\textit{traffic accident}, C=\textit{traffic congestion}, M=\textit{near-miss}, O=\textit{containing outliers}, N=\textit{normal traffic}.}
\\[-3pt]
\legendrow{\textbf{Table \thetable{} (sub-section for \textit{anomaly})}}
\toprule
\footnotesize
 & \multirow{3}{*}{Year}
 & \multicolumn{6}{c}{Data Acquisition}
 & \multicolumn{7}{c}{Labels for Critical Elements}
 & \multicolumn{3}{c}{Dataset Contents}
 & \multirow{3}{*}{Scene} \\
\cmidrule(lr){3-8} \cmidrule(lr){9-15} \cmidrule(lr){16-18}
 & & \multicolumn{2}{c}{Region} & \multirow{2}{*}{View} & \multirow{2}{*}{Source} & \multicolumn{2}{c}{Sensor}
 & \multicolumn{6}{c}{Level} & Class\footnotemark[2]
 & \multirow{2}{*}{Video\footnotemark[3]} & \multirow{2}{*}{\makecell{Image/\\Frame}} & \multirow{2}{*}{Resolution} \\
\cmidrule(lr){3-4} \cmidrule(lr){7-8} \cmidrule(lr){9-14} \cmidrule(lr){15-15}
\quad Datasets & & Country & City & & & Cam & Oth
& C & L & D & S & I & P & {\textit{p}(cat)+\textit{n}}
& & & & \\
\midrule
\endfirsthead

\multicolumn{19}{l}{\textbf{Table \thetable{} (continued sub-section for \textit{pertinent entity})}} \\
\toprule
\footnotesize
 & \multirow{3}{*}{Year}
 & \multicolumn{6}{c}{Data Acquisition}
 & \multicolumn{7}{c}{Labels for Critical Elements}
 & \multicolumn{3}{c}{Dataset Contents}
 & \multirow{3}{*}{Scene} \\
\cmidrule(lr){3-8} \cmidrule(lr){9-15} \cmidrule(lr){16-18}
 & & \multicolumn{2}{c}{Region} & \multirow{2}{*}{View} & \multirow{2}{*}{Source} & \multicolumn{2}{c}{Sensor}
 & \multicolumn{6}{c}{Level} & Class\footnotemark[2]
 & \multirow{2}{*}{Video\footnotemark[3]} & \multirow{2}{*}{\makecell{Image/\\Frame}} & \multirow{2}{*}{Resolution} \\
\cmidrule(lr){3-4} \cmidrule(lr){7-8} \cmidrule(lr){9-14} \cmidrule(lr){15-15}
\quad Datasets & & Country & City & & & Cam & Oth
& C & L & D & S & I & P & {\textit{p}(cat)+\textit{n}}
& & & & \\
\midrule
\endhead

    \multicolumn{19}{@{}l}{\textbf{Obstacle Segmentation}} \\
    \quad \href{http://wwwlehre.dhbw-stuttgart.de/~sgehrig/lostAndFoundDataset/index.html}{LostAndFound} \cite{lostAndFound} & 2016 & {\scriptsize DE} & 1 & Eg &  Au & st & & & & & & & \checkmark & 1(26)+3 & T & 2239 & 2048×1024 & O \\
    \quad \href{https://segmentmeifyoucan.com/datasets}{RoadObstacle21} \cite{chan2021segmentmeifyoucan} & 2021 & {\scriptsize DE,CH} & -- & Eg &  Au & \checkmark & & & & & \checkmark & & & 1(31)+2 & T &  412(+30) & 1920×1080 & O \\
    \multicolumn{19}{@{}l}{\textbf{Anomaly Segmentation}} \\
    \quad \href{http://wwwlehre.dhbw-stuttgart.de/~sgehrig/lostAndFoundDataset/index.html}{LostAndFound} \cite{lostAndFound} & 2016 & {\scriptsize DE} & 1 & Eg &  Au & st & & & & & & & \checkmark & 1(9)+3 & T & 2104 & 2048×1024 & O \\
    \quad \href{https://www.epfl.ch/labs/cvlab/data/road-anomaly/}{RoadAnomaly} \cite{RoadAnomalyDataset} & 2019 & -- & -- & Eg &  Wb & \checkmark & & & & \checkmark & & \checkmark & & 1(7)+1 & -- & 60 & 1280×720 & O\\
    \quad \href{https://segmentmeifyoucan.com/datasets}{RoadAnomaly21} \cite{chan2021segmentmeifyoucan} & 2021 & -- & -- & Eg &  Wb & \checkmark & & & & & \checkmark & & & 1(26)+2 & -- &  100(+10) & Varied & O\\
    \quad \href{https://fishyscapes.com/dataset}{FS Static} \cite{fishyscapes} & 2019 & DE & 3 & Eg & \cite{Cityscapes, PascalVOCDATASET} & st,\checkmark & syn & & & & \checkmark & & & 1(12)+1 & -- &  1000(+30) & 2048×1024 & O \\
    \quad \href{https://fishyscapes.com/dataset}{FS Lost\&Found} \cite{fishyscapes} & 2019 & DE & 1 & Eg & \cite{lostAndFound} & st & & & & & \checkmark & & & 1(7)+2 & -- &  275(+100) & 2048×1024 & O \\
    \quad \href{https://omnomnom.vision.rwth-aachen.de/data/stu-dataset/}{STU} \cite{nekrasov2025stu} & 2025 & AU & ${1}^{\star}$ & Eg &  Au & \checkmark &  Li & & & & & & \checkmark & 1+2 & 70 & -- & 1928×1208 & O,N \\
    \multicolumn{19}{@{}l}{\textbf{Traffic Congestion Classification}} \\
    \quad \href{http://www.openits.cn/openData4/824.jhtml}{CCTRIB} \cite{CCTRIB} & 2022 & CN & -- & Sv &  Wb & tc & & \checkmark & & & & & & 1+1(attr) & -- & 9200 & Varied & N,C \\
    \quad \href{https://www.kaggle.com/datasets/dtrnngc/ua-detrac-dataset}{UA-DETRAC} \cite{UA-DETRAC} & 2020 & CN & 2 & Sv &  Au & \checkmark & & & & \checkmark & & & & (4)(attr) & 100 &  140131 & 960×540 & N,C \\
    \multicolumn{19}{@{}l}{\textbf{Traffic Accident Classification}} \\
    \quad \href{https://ieee-dataport.org/documents/traffic-accident-detection-video-dataset-ai-driven-computer-vision-systems-smart-city}{TADD} \cite{TADD} & 2023 & -- & -- & Sv,Eg & \cite{CarCrashDataset},Wb & tc,dc,mc & & \checkmark & & & & & & 7+1 & 5691 & -- & Varied & A,N \\
    \quad \href{https://docs.google.com/document/d/12F7l4yxNzzUAISZufEd9WFhQKSefVVo_QsPdTsWxZh8/edit?tab=t.0}{CADP} \cite{CADPDataset} & 2018 & -- & -- & Sv,Eg &  Wb & tc,dc,mc & & \checkmark & & & & & & 1+1 & 1416 & -- & Varied & A,M \\
    \multicolumn{19}{@{}l}{\textbf{Traffic Accident Anticipation}} \\
    \quad \href{https://github.com/Cogito2012/CarCrashDataset?tab=readme-ov-file#download}{CCD} \cite{CarCrashDataset} & 2020 & -- & -- & Eg & \cite{BDD100K},Wb & \checkmark & & \checkmark & & & & & & 1+1(attr) & 4500 & -- & Varied & A,N \\
    \quad \href{https://github.com/smallcorgi/Anticipating-Accidents}{DAD} \cite{DAD} & 2016 & CN & 6 & Eg &  Wb & dc & & \checkmark & & \checkmark & & & & 1+1 & 1750 &  175000 & 1280×720 & A,N,O \\
    \quad \href{https://github.com/MoonBlvd/tad-IROS2019}{A3D} \cite{A3D} & 2019 & -- & -- & Eg &  Wb & dc & & \checkmark & & & & & & 1(18)+1 & 1500 &  128175 & Varied & A \\ 
    \multicolumn{19}{@{}l}{\textbf{Accident Causality Recognition}} \\
    \quad \href{https://github.com/MoonBlvd/Detection-of-Traffic-Anomaly}{DoTA} \cite{DoTA} & 2023 & -- & -- & Eg & Wb & dc & & \checkmark & & \checkmark & & & & 18+1 & 4677 &  731932 & 1280×720 & A \\
    \quad \href{https://github.com/tackgeun/CausalityInTrafficAccident/tree/master/dataset}{CTA} \cite{you2020CTA} & 2020 & -- & -- & Sv,Eg &  Wb & tc,dc & & \checkmark & & & & & & 25+1 & 1935 & -- & Varied & A \\
    \quad \href{https://huggingface.co/datasets/JeffreyChou/MM-AU}{MM-AU} \cite{MM-AU} & 2024 & -- & -- & Eg &  Ds,Wb & \checkmark & & \checkmark & \checkmark & \checkmark & & & & 58/110 &  11727 &  2195613 & Varied & A \\
    \multicolumn{19}{@{}l}{\textbf{Traffic Crime Classification}} \\
    \quad \href{https://www.kaggle.com/datasets/odins0n/ucf-crime-dataset}{UCF-Crime}\footnotemark[4] \cite{Sultani_2018_CVPR} & 2018 & -- & -- & Sv &  Wb & tc,sc & & \checkmark & & & & & & 13+1 & 1900 & -- & 320×240 & A,N \\
    \quad \href{https://msad-dataset.github.io/}{MSAD}\footnotemark[4] \cite{MSAD} & 2024 & -- & -- & Sv &  Wb & sc & & \checkmark & & & & & & 11+1 & 720 &  447236 & Varied & A,N \\
    \multicolumn{19}{@{}l}{\textbf{Collision Prediction (Ego-Involved)}} \\
    \quad CST-S3D \cite{CST-S3D} & 2022 & -- & -- & Eg & \cite{DADA-2000} & \checkmark & & \checkmark & & & & & &  3/6/15+1 & 704 & -- & 1584×660 & A \\
    \multicolumn{19}{@{}l}{\textbf{Behavior Recognition (Ego-centric)}} \\
    \quad \href{https://sites.google.com/view/crash-to-not-crash}{YouTubeCrash} \cite{CrashToNotCrash} & 2019 & -- & -- & Eg &  Wb & dc & & \checkmark & \checkmark & \checkmark & & & & 1+1 & 222 & 4440 & 710×400 & A,N \\
    \quad \href{https://sites.google.com/view/crash-to-not-crash}{GTACrash} \cite{CrashToNotCrash} & 2019 & -- & -- & Eg &  Au & & sim & \checkmark & \checkmark & \checkmark & & & & 1+1 &  11381 &  227620 & 710×400 & A \\
    \multicolumn{19}{@{}l}{\textbf{Driver Attention Prediction}} \\
    \quad \href{https://bdd-data.berkeley.edu/}{BDD-A} \cite{Xia2017PredictingDA} & 2017 & US & 3 & Eg & \cite{BDD100K} & mc & eye & & & & \checkmark & & & 1+1 & 1435 & -- & 1280×720 & N,M \\
    \quad \href{https://github.com/JWFangit/LOTVS-DADA}{DADA-2000} \cite{DADA} & 2019 & -- & -- & Eg &  Wb & \checkmark & eye & \checkmark & & & \checkmark & & & 1(54)+1 & 2000 &  658476 & 1584×660 & A,M,O \\
    \multicolumn{19}{@{}l}{\textbf{Crack Segmentation}} \\
    \quad \href{https://github.com/qinnzou/DeepCrack?tab=readme-ov-file}{CrackTree260} \cite{CrackTree} & 2012 & CN & -- & Lc &  Au & \checkmark & & & & & \checkmark & & & 1+1 & -- &  260/35100 & Varied & O \\
    \quad \href{https://github.com/cuilimeng/CrackForest-datase}{CFD} \cite{CrackForest} & 2016 & CN & 1 & Lc &  Au & mc & & & & & \checkmark & & & 1+1 & -- & 155 & Varied & O \\
    \quad \href{https://github.com/fyangneil/pavement-crack-detection?tab=readme-ov-file}{CRACK500} \cite{CRACK500} & 2020 & US & 1 & Lc &  Au & mc & & & & & \checkmark & & & 1+1 & -- &  500/3368 & Varied & O \\
    \quad \href{https://github.com/fyangneil/pavement-crack-detection}{GAPs384} \cite{CRACK500} & 2020 & DE & -- & Lc & \cite{GAPs} & mono & & & & & \checkmark & & & 1+1 & -- &  384/509 & Varied & O \\
    \quad \href{https://github.com/mqp2259/EdmCrack600}{EdmCrack600} \cite{EdmCrack600} & 2020 & CA & 1 & Lc & Au & GoPro & & & & & \checkmark & & & 1+1 & -- & 600 & 1920×1080 & O \\
    \quad \href{https://github.com/ZheningHuang/NHA12D-Crack-Detection-Dataset-and-Comparison-Study}{NHA12D} \cite{NHA12D} & 2022 & UK & -- & Lc &  Au & \checkmark & & & & & \checkmark & & & 1+1 & -- & 80 & 1920×1080 & O \\
    \multicolumn{19}{@{}l}{\textbf{Pothole Segmentation}} \\
    \quad \href{https://sites.google.com/view/pothole-600/dataset}{Pothole-600} \cite{Pothole600} & 2020 & -- & -- & Lc &  Au & st & & & & & \checkmark & & & 1+1 & -- & 67 & Varied & O \\

    \multicolumn{19}{@{}l}{\textbf{Road Damage Classification}} \\
    \quad \href{https://github.com/DearCaat/CQU-BPDD}{CQU-BPDD} \cite{CQU-BPDD} & 2020 & CN & -- & Lc &  Au & \checkmark & & \checkmark & & & & & & 1(6)+1 & -- & 60059 & 1200×900 & O,N \\
    \multicolumn{19}{@{}l}{\textbf{Road Damage Segmentation}} \\
    \quad \href{https://data.mendeley.com/datasets/kfth5g2xk3/2}{PotholeMix} \cite{SHREC} & 2022 & -- & -- & Lc &  Ds,Wb & \checkmark & & & & & \checkmark & & & 2+1 & 797 & 4340 & Varied & O \\
    \quad \href{https://huggingface.co/datasets/DarrylT/M2S-RoAD/tree/main}{M2S-RoAD} \cite{M2S-RoAD} & 2024 & AU & ${4}^{\star}$ & Eg &  Au & \checkmark &  Li & & & & \checkmark & & & 9+1 & -- & 1071 & 1928×1208 & O \\
    \multicolumn{19}{@{}l}{\textbf{Road Damage Detection}} \\
    \quad \href{https://www.tu-ilmenau.de/en/university/departments/department-of-computer-science-and-automation/profile/institutes-and-groups/institute-of-computer-and-systems-engineering/group-for-neuroinformatics-and-cognitive-robotics/data-sets-code/german-asphalt-pavement-distress-dataset-gaps}{GAPs} \cite{ExtendGAPs} & 2019 & DE & -- & Lc &  Au & mono & & & & \checkmark & & & & 5+1 & -- & 2468 & 1920×1080 & O \\
    \quad \href{https://github.com/sekilab/RoadDamageDetector?tab=readme-ov-file}{RDD2022} \cite{RDD2022} & 2022 & Glob\footnotemark[5] & -- & Eg,Lc &  Au,Wb & \checkmark,mc &  GV & & & \checkmark & & & & 4+1 & -- & 47420 & Varied & O \\
    \multicolumn{19}{@{}l}{\textbf{Traffic Weather Recognition}} \\
    \quad \href{https://acdc.vision.ee.ethz.ch/download}{ACDC} \cite{ACDC} & 2021 & CH & -- & Eg &  Au &  GoPro & & \checkmark & & \checkmark & \checkmark & & \checkmark & 4(19) & & 4006 & 1920×1080 & O \\
    \multicolumn{19}{@{}l}{\textbf{Illumination Recognition and Adaption}} \\
    \quad \href{https://www.kaggle.com/datasets/ipythonx/nighttime-driving-dataset}{NightDriving} \cite{NightDriving} & 2018 & CH & -- & Eg & Au & GoPro & & & & & \checkmark & & & 5(19) & -- & 35000 & 1920×1080 & O,N \\
    \quad \href{https://dmcv.sjtu.edu.cn/people/phd/tanxin/NightCity/index.html}{NightCity} \cite{nightcity} & 2021 & Glob\footnotemark[5] & 13+ & Eg &  Wb & dc & & & & & \checkmark & & & 1(19) & -- & 4297 & 1024×512 & O \\
    \quad \href{https://www.trace.ethz.ch/publications/2019/GCMA_UIoU/}{DarkZurich} \cite{darkzurich} & 2022 & CH & 1 & Eg &  Au &  GoPro & & & & & \checkmark & & & 3(19) & T & 8779 & 1920×1080 & O,N \\
    \multicolumn{19}{@{}l}{\textbf{Risk Identification}} \\
    \quad \href{https://hcis-lab.github.io/RiskBench/}{RiskBench} \cite{RiskBench} & 2024 & -- & -- & Eg &  Au & & sim & & & \checkmark & & \checkmark & & 3+1(attr) & 6916 & ++ & 640×256 & A,N,O \\
    \midrule
    
    \multicolumn{19}{@{}l}{\textbf{Pedestrian Intention Prediction}} \\
    \quad \href{https://data.nvision2.eecs.yorku.ca/JAAD_dataset/}{JAAD} \cite{JAAD} & 2017 & Glob\footnotemark[5] & 5 & Eg &  Au & dc,GoPro & & & & \checkmark & & & & 1+1(attr) & 346 & 82032 & Varied & N \\
    \quad \href{https://github.com/aras62/PIE}{PIE} \cite{PIE} & 2019 & CA & 1 & Eg & Au & dc & GPS & & & \checkmark & & & & 1+1(attr) & 53 & -- & 1920×1080 & N \\
    \quad STIP \cite{STIP} & 2020 & US & 8 & Eg &  Au & \checkmark & & & & \checkmark & & & & 1+1 & T &  1108176 & 1936×1216 & N \\
    \multicolumn{19}{@{}l}{\textbf{Pedestrian Pose Estimation}} \\
    \quad \href{https://deepblue.lib.umich.edu/data/concern/data_sets/6h440s98b?locale=en}{PedX} \cite{pedx} & 2019 & US & 1 & Eg &  Au & st &  Li & & & & & \checkmark & & 18+1(attr) & T & 5076 & 4112×3008 & N \\
    \quad \href{https://waymo.com/open/data/perception/}{Waymo} \cite{Waymo} & 2022 & US & 6 & Eg & Au & \checkmark &  Li & & & \checkmark & & \checkmark & & 14(2)(attr) & 1150 & 230000 & 1920×1280 & N \\
    \quad \href{https://eurocity-dataset.tudelft.nl/eval/extensions/ecpdp}{ECPDP} \cite{ECPDP} & 2021 & EU\footnotemark[6] & 31 & Eg & Au,\cite{EuroCity} & \checkmark & & & & \checkmark & & & & 17+1(attr) & -- & 46975 & 1920×1024 & N \\
    \multicolumn{19}{@{}l}{\textbf{Scene Classification}} \\
    \quad \href{https://ieee-dataport.org/documents/scenegraph-risk-assessment-dataset}{CARLA-syn} \cite{CarlaSyn} & 2022 & -- & -- & Eg & Au & & sim & \checkmark & & & & & & 1+1(attr) & Varied & Varied & 1280×720 & A,N \\
    \quad sub-NUDrive \cite{sub-NUDrive} & 2019 & JP & 1 & Eg &  Au & \checkmark & & \checkmark & & & & & & 1+1 & 860 & 43000 & 692×480 & N \\
    \multicolumn{19}{@{}l}{\textbf{Pedestrian Detection}} \\
    \quad \href{https://data.caltech.edu/records/f6rph-90m20}{Caltech}\cite{CALTECH} & 2009 & US & 3 & Eg &  Au & \checkmark & & & & \checkmark & & & & 1(4)+1 & 137 &  249884 & 640×480 & N \\
    \quad \href{https://www.nightowls-dataset.org/}{NightOwls} \cite{NightOwls} & 2018 & DE,NL,UK & 7 & Eg & Au & \checkmark & & & & \checkmark & & & & 3+2 & 40 &  279000 & 1024×640 & N \\
    \quad \href{https://github.com/CharlesShang/Detectron-PYTORCH/tree/master/data/citypersons/annotations}{CityPersons} \cite{CityPersons} & 2017 & DE,CH,FR & 27 & Eg & \cite{Cityscapes} & st & & & & \checkmark & & & & 1+6 & -- & 5000 & 2048×1024 & N \\
    \quad \href{https://eurocity-dataset.tudelft.nl/eval/user/login?_next=/eval/downloads/detection#}{ECP} \cite{EuroCity} & 2019 & EU\footnotemark[6] & 31 & Eg &  Au & \checkmark & & & & \checkmark & & & & 1+17 & -- & 47335 & 1920×1024 & N \\
    \quad \href{https://github.com/bupt-ai-cz/LLVIP/blob/main/download_dataset.md}{LLVIP} \cite{LLVIP} & 2021 & CN & 1 & Sv & Au & ptz & & & & \checkmark & & & & 1+1 & -- & 15488 & 1280×1024 & N \\
    \multicolumn{19}{@{}l}{\textbf{Cyclist Detection}} \\
    \quad \href{http://www.gavrila.net/Datasets/Daimler_Pedestrian_Benchmark_D/Tsinghua-Daimler_Cyclist_Detec/tsinghua-daimler_cyclist_detec.html}{TDC} \cite{cyclistdetection} & 2016 & CN & 1 & Eg &  Au & st & & & & \checkmark & & & & 1(4)+1 & 137 &  249884 & 2048×1024 & N \\
    \multicolumn{19}{@{}l}{\textbf{2D Lane Detection}} \\
    \quad \href{https://github.com/TuSimple/tusimple-benchmark/tree/master/doc/lane_detection}{TuSimple} \cite{TuSimpleGit} & 2017 & US & -- & Eg & Au & \checkmark & & & & \checkmark & & \checkmark & & 1+1 & 6408 &  128160 & 1280×720 & N \\
    \quad \href{https://xingangpan.github.io/projects/CULane.html}{CULane} \cite{CULane} & 2018 & CN & 1 & Eg &  Au & \checkmark & & & & \checkmark & & \checkmark & & 1+1 & -- &  133235 & 1640×590 & N \\
    \quad \href{https://datasetninja.com/bdd100k#download}{BDD100K} \cite{BDD100K} & 2020 & US & 3 & Eg &  Au & mc & & & & \checkmark & & & & 32+1 & -- &  100000 & 1280×720 & N \\
    \quad \href{https://github.com/yujun0-0/mma-net}{VIL-100} \cite{VIL100} & 2021 & CN & -- & Eg & Au,Wb & \checkmark & & & & \checkmark & & \checkmark & & 1+1 & 100 &  100000 & Varied & N \\
    \quad \href{https://github.com/dongkwonjin/RVLD}{OpenLane-V} \cite{openlane-v} & 2023 & US & 6 & Eg & \cite{openlane} & \checkmark & & & & \checkmark & & \checkmark & & 1+1 & T &  90000 & 1920×1080 & N \\
    \multicolumn{19}{@{}l}{\textbf{2D Lanemark Segmentation}} \\
    \quad ApolloScape \cite{ApolloScape} & 2020 & CN & 3 & Eg &  Au & \checkmark & & & & & \checkmark & & & 35+2 & -- &  165949 & 3384×2710 & N \\
    \multicolumn{19}{@{}l}{\textbf{3D Lane Detection}} \\
    \quad \href{https://github.com/OpenDriveLab/OpenLane/blob/main/data/README.md}{OpenLane} \cite{openlane} & 2022 & US & 6 & Eg & \cite{Waymo} & \checkmark & & & & \checkmark & & & & 14+1 & T &  200000 & 1920×1280 & N \\
    \multicolumn{19}{@{}l}{\textbf{HD Map Construction}} \\
    \quad \href{https://www.nuscenes.org/nuscenes#download}{nuScenes} \cite{nuScenes} & 2020 & US,SG & 2 & Eg & Au & \checkmark & \makecell{Li,imu,\\ra,GPS} & & & \checkmark & \checkmark\footnotemark[8] & & & 11/23 & 1000 & \makecell{40000 /\\ 1400000} & 1600×900 & N \\
    \quad \href{https://www.argoverse.org/av2.html#download-link}{Argoverse2} \cite{Argoverse2} & 2021 & US & 6 & Eg & Au & \checkmark,st & Li & & & \checkmark & \checkmark\footnotemark[8] & & & 4/30 & 1000 & 2700000 & 2048×1550 & N \\
    \multicolumn{19}{@{}l}{\textbf{Traffic Sign Detection}} \\
    \quad \href{https://www.vicos.si/resources/dfg/}{DFG} \cite{DFG} & 2020 & SI & ${6}^{\star}$ & Eg &  Au & \checkmark & & & & \checkmark & & \checkmark & & 200+1 & -- & 6957 & Varied & N \\
    \quad \href{https://cg.cs.tsinghua.edu.cn/traffic-sign/}{TT100K} \cite{Tencent2016} & 2021 & CN & ${5}^{\star}$ & Eg &  Wb & ref & & & & \checkmark & & & & 232+1 & -- &  100000 & 2048×2048 & N \\
    \quad \href{https://github.com/citlag/Traffic-Sign-Recognition}{GTSDB} \cite{GTSDB2020} & 2020 & DE & 1 & Eg &  Au & \checkmark & & & & \checkmark & \checkmark & & & 164+1 & -- & 900 & 1360×800 & N \\
    \quad \href{https://github.com/NicholasCG/GLARE_Dataset}{GLARE} \cite{Glare} & 2023 & US & 1 & Eg &  Au & dc & & \checkmark & \checkmark & & & & & 41+1 & T & 2157 & Varied & N \\
    \multicolumn{19}{@{}l}{\textbf{Traffic Sign Graph Generation}} \\
    \quad \href{https://nlpr.ia.ac.cn/databases/CASIA-Tencent%20CTSU/index.html}{CTSU} \cite{CTSU} & 2021 & CN & ${-}^{\star}$ & Lc &  Au & \checkmark & & & & \checkmark & & & & 55/2+1 & -- & 5000 & Varied & N \\
    \quad \href{https://nlpr.ia.ac.cn/pal/RS10K.html}{RS10K} \cite{RS10K} & 2023 & CN & 31 & Eg &  Au & dc & & & & & \checkmark & & & 6/5+1 & -- &  10041 & Varied & N \\
    \multicolumn{19}{@{}l}{\textbf{Traffic Light Detection}} \\
    \quad \href{https://www.kaggle.com/datasets/mbornoe/lisa-traffic-light-dataset}{LISA} \cite{LISA} & 2016 & US & 1 & Eg &  Au & st & & & & \checkmark & & & & 7+1 & T &  43007 & 1280×960 & N \\
    \quad \href{https://zenodo.org/records/12706046}{BSTLD} \cite{BOSCH} & 2017 & US & 2 & Eg & Au & \checkmark & & & & \checkmark & \checkmark & & & 13+1 & -- &  13427 & 1280×720 & N \\
    \quad \href{https://www.uni-ulm.de/en/in/institute-of-measurement-control-and-microtechnology/research/data-sets/driveu-traffic-light-dataset/}{DriveU} \cite{DriveU} & 2021 & DE & 11 & Eg & Au & st & & & & \checkmark & & & &  19200+1 & T &  40978 & 2048×1024 & N \\
    \quad \href{https://github.com/Thinklab-SJTU/S2TLD}{$\text{S}^2\text{TLD}$} \cite{s2tld} & 2023 & CN & -- & Eg &  Au & \checkmark & & & & \checkmark & & & & 5+1 & T & 5786 & Varied & N \\
    \multicolumn{19}{@{}l}{\textbf{Traffic Police Gesture Recognition}} \\
    \quad \href{https://github.com/zc402/ChineseTrafficPolicePose?tab=readme-ov-file}{TPGR} \cite{HE2020248} & 2020 & CN & 1 & Lc & Au & \checkmark & & \checkmark & & & & & & 8+1 & 21 & -- & 1080×1080 & N \\
    \multicolumn{19}{@{}l}{\textbf{Lane Change Classification and Prediction}} \\
    \quad \href{https://prevention-dataset.uah.es/}{PREVENTION} \cite{prevention} & 2019 & ES & 1 & Eg & Au & \checkmark & \makecell{Li,imu,\\ra,GPS} & & & \checkmark & & \checkmark & & 2(6)+1 & 5 & -- & 1920×600 & N \\
    \multicolumn{19}{@{}l}{\textbf{Driving Maneuver Classification}} \\
    \quad \href{https://github.com/VCCIV/BLVD/tree/master}{BLVD} \cite{BLVD} & 2019 & CN & 1 & Eg & Au & \checkmark & Li,GPS & & & \checkmark & & &  & 13(3) & 654 & 120000 & 1920×500 & N \\
    \bottomrule
\end{longtable}
}
\vspace{-0.5em}
\noindent\begin{minipage}{\textwidth}
\footnotesize
\begin{enumerate}[leftmargin=*,noitemsep,topsep=0pt]
        \scriptsize
        \item \textbf{Please note}: The checkmark symbol (\checkmark) is omitted in cells where eligibility is already indicated by other contents (e.g., citations).
        \item \textbf{\textit{p}(cat)+\textit{n}}: \textit{p} indicates the number of positive classes, while (cat) specifies how many semantic categories these positive classes belong to, and \textit{n} denotes the number of negative classes. Further, \textbf{(attr)} indicates that additional attributes (e.g., weather conditions) are also annotated.
        \item \textbf{Video}: Datasets providing temporal information via image sequences rather than continuous videos are marked with `T'. 
        \item \textbf{UCF-Crime and MSAD}: Although some footage is unrelated to traffic scenarios, the table reports the total numbers of videos and frames for both datasets.
        \item \textbf{Glob}: RDD2022 \cite{RDD2022} contains images from 6 countries; NightCity \cite{nightcity} contains footage from 9 countries; JAAD \cite{JAAD} is collected in 4 countries.
        \item \textbf{EU}: ECPDP \cite{ECPDP} and ECP \cite{EuroCity} covers footage from 12 European countries.
        \item \textbf{CARLA-syn}: It covers three synthetic datasets in \cite{CarlaSyn}: 271-syn, 1043-syn, and 306-syn, where each numerical prefix denotes the video count.
        \item \textbf{nuScenes and Argoverse2}: Although not annotated with pixel-wise masks, category-specific map elements are represented using semantic-level boundaries, such as polylines and polygons, to support HD map construction, and are therefore more closely aligned with semantic segmentation.
        \item \textbf{${\star}$}: The presence of rural data is explicitly stated in the descriptions of datasets, including {STU} \cite{nekrasov2025stu}, {M2S-RoAD} \cite{M2S-RoAD}, {DFG} \cite{DFG}, {TT100K} \cite{Tencent2016}, and {CTSU} \cite{CTSU}.
\end{enumerate}
\end{minipage}

\section{Critical Tasks and Corresponding Datasets}
\label{sec:tasks_datasets}
This section analyzes benchmark and promising tasks and datasets about critical attention-worthy elements in traffic scenarios. Each subsection corresponds to one of the aforementioned 11 categories and is organized using a clear task-to-dataset hierarchy. Within each category, the relevant tasks are first introduced, followed by the examination and visualization of the corresponding datasets under each task. The categories, tasks, and datasets are arranged according to the order in Table \ref{table:Existing Tasks Classification}.
Table~\ref{table:visual_datasets_summary} follows the same sequence and provides a comprehensive overview of the basic characteristics of all surveyed datasets, including data collection, annotation level, number of categories, data volume, and scenario type. In addition, except for datasets that have become temporarily unavailable, links are embedded into the dataset names in Table~\ref{table:visual_datasets_summary} for easy access.

\subsection{Anomaly due to Spatial Location} \label{subsec:anomaly_spatial}
 
When it comes to road safety, the most critical entities are obstacles on the road. If there is any obstacle in the path of the ego vehicle that impedes safe and normal traffic, regardless of the semantic category, it deserves the attention of the ego vehicle. Corresponding tasks have been proposed to detect obstacles in traffic scenarios. Certain surveys, such as \cite{PERUMAL2021104406}, proposed a broad notion of obstacles, encompassing both dynamic road users and static environmental elements on the front of the ego vehicle that might lead to a crash. However, in this paper, we adhere to a more focused and classic definition, based on which the entities of interest in the context of traffic obstacle analysis specifically refer to the static obstacles present in the projected path of the ego vehicle and impede safe and normal traffic. Consequently, while video sequences are available in certain datasets, such as \cite{lostAndFound}, no tracking task is proposed for this type of obstacles.

\subsubsection{Obstacle Segmentation}
\label{task:obstacle_segmentation}
Obstacle segmentation aims to identify and locate obstacles in the background. Although some literature refers to this task as `obstacle detection' \cite{lostAndFound}, we adopt the term \textit{obstacle segmentation} in our study, considering its conformity to segmentation tasks. It provides instance pixel-level labels (analogous to \hyperref[subsec:instance_segmentation]{instance segmentation}) and/or semantic pixel-level annotations (analogous to \hyperref[subsec:semantic_segmentation]{semantic segmentation}) for obstacles.

Here, we discuss two benchmark datasets, \textit{LostAndFound} \cite{lostAndFound} and \textit{RoadObstacle21} \cite{chan2021segmentmeifyoucan}. Fig.~\ref{fig:obstacle_anomaly_datasets_comparison} (a) shows representative images and corresponding annotations from both datasets. For LostAndFound~\cite{lostAndFound}, the overlay annotations were generated from JSON data, while for RoadObstacle21~\cite{chan2021segmentmeifyoucan}, the overlays represent semantic masks provided by the dataset, rendered with consistent coloring but increased transparency for better visibility.

\paragraph{\textbf{LostAndFound:}}
\label{dataset:lostandfound}
LostAndFound \cite{lostAndFound} is proposed to advance the segmentation of small traffic obstacles, a typically underrepresented but critical research area.
It consists of 112 stereo videos recorded in urban environments \cite{fishyscapes} in the Greater Stuttgart area, Germany \cite{Bogdoll_Perception_2023_IV} and includes frames at 2048×1024 resolution containing 37 obstacle types, such as cargo and traffic cones, 26 of which appear in both subsets. The respective training/test split reported in \cite{lostAndFound} and provided in the \href{http://wwwlehre.dhbw-stuttgart.de/~sgehrig/lostAndFoundDataset/index.html}{online package} is 1036/1068 and 1036/1203.
In accordance with Cityscapes \cite{Cityscapes}, each image is annotated by four types of segmentation masks and a JSON file (with semantic label, instance ID and polygon coordinates). The annotations follow a panoptic segmentation scheme, where all individual foreground obstacles are distinguished, while background pixels are categorized into three semantic groups, namely drivable, out-of-roi, and ego-vehicle areas. However, as shown in Fig.~\ref{fig:obstacle_anomaly_datasets_comparison}, the annotated free space is bounded by a quadrilateral with the bottom boundary originating from the apex of the ego vehicle logo, rather than adhering to the actual drivable area. Additionally, there is a drift between the annotated ego vehicle mask and the real region. These issues could bias the model toward learning annotation artifacts and impair boundary-sensitive learning by introducing noisy supervision.

\paragraph{\textbf{RoadObstacle21:}}
\label{dataset:RoadObstacle21}
In collaboration with \cite{10.1109/TPAMI.2023.3335152}, RoadObstacle21 (RO21) \cite{chan2021segmentmeifyoucan} also restricts the region of interest to the drivable area with obstacles ahead. Images are captured in 12 different road, lighting, and weather conditions in Germany and Switzerland, with a resolution of 1920×1080. All 442 images are \href{https://segmentmeifyoucan.com/datasets}{freely accessible}, with 327 for evaluation, 85 captured at night or in extreme weather, and 30 for validation. Ground-truth semantic masks are only provided for 30 validation images, categorizing each pixel into one of three classes, obstacles (orange), drivable area (white), and uninterested area (black). Despite the reported coverage of 31 obstacle types, they are all treated as a single semantic category, obstacle, without distinguishing individual obstacles. The temporal order among relevant images is reflected by the file name, while in some cases, the temporal relationship is incorrect (e.g., curvy-street\_umbrella\_1 and curvy-street\_umbrella\_2) or inverted (e.g., darkasphalt series). This lack of semantic and temporal precision may hinder the model capability to capture fine-grained obstacle characteristics and reliable temporal dependencies.

\input{figures_figures/1_obstacle_anomaly_datasets_comparison}

\subsection{Anomaly Due to Semantic Category} \label{subsec:anomaly_semantic}

The second type of critical scene components worth noticing is objects or regions that have never been seen before. It is commonly believed that, instead of being overconfident in approximating them to a known category, identifying these deviations as anomalies is more reliable \cite{chan2021segmentmeifyoucan}. In this context, the task named \textit{anomaly segmentation} is proposed.

\subsubsection{Anomaly Segmentation}
\label{task:anomaly_segmentation}
Traditionally, it is designed as a semantic-level task to identify anomalous regions that do not correspond to any predefined class in the training data. It has now been extended to the segmentation of anomalous instances to further identify individual unexpected objects from 2D \cite{AnomalyInstanceSegmentation} or even 3D traffic scenes \cite{nekrasov2025stu}, as a stepping stone toward safe and reliable autonomous driving. Since existing methods usually involve the evaluation of probability distribution and uncertainty or confidence scores, it is also considered a subtask of \textit{out of distribution detection} or \textit{uncertainty estimation}. While anomalies can appear anywhere in the image \cite{chan2021segmentmeifyoucan,fishyscapes}, given the driving focus nature of this task, anomalies are created or annotated within the drivable area in most images. This positioning allows these anomalies to be interpreted as road obstacles, making the corresponding images potentially suitable for the obstacle segmentation task. However, the key distinction is that anomalies are absent from the training data, whereas obstacles are present.

In this section, we analyze five 2D anomaly segmentation benchmark datasets, involving \textit{LostAndFound} \cite{lostAndFound}, \textit{RoadAnomaly} \cite{RoadAnomalyDataset}, \textit{RoadAnomaly21} \cite{chan2021segmentmeifyoucan}, \textit{Fishyscapes (including FS Static and FS LostAndFound})\cite{fishyscapes}, and a 3D anomaly segmentation dataset, \textit{Spotting the Unexpected (STU)} \cite{nekrasov2025stu}. Fig.~\ref{fig:obstacle_anomaly_datasets_comparison}(b) shows sample images and annotations from the last four datasets. Specifically, for RoadAnomaly~\cite{RoadAnomalyDataset}, the anomaly annotations were derived from instance-level masks, with each instance assigned a distinct random color. For RoadAnomaly21~\cite{chan2021segmentmeifyoucan}, the overlays represent the provided semantic masks, rendered with consistent coloring but increased transparency for better visibility. For FS Static and FS LostAndFound \cite{fishyscapes}, anomalies are rendered in red and orange with transparency for better visualization, while other regions keep the original semantic-mask colors with increased transparency.

\paragraph{\textbf{LostAndFound:}}
\label{dataset:LostAndFound}
Despite the focus on obstacles, the LostAndFound test set \cite{lostAndFound} contains 9 categories that are not present in the training or validation subset. Technically, these categories can be treated as anomalies in this dataset, though they might be common objects in traffic scenarios, such as children and bicycles \cite{chan2021segmentmeifyoucan}. In this case, individual-level distinctions are needed only for these previously unseen anomalous obstacles. The remaining 26 obstacle types that appear simultaneously in the test, training, and validation sets should be treated as negative (background) elements. 

\paragraph{\textbf{RoadAnomaly:}}
\label{dataset:RoadAnomaly}
RoadAnomaly \cite{RoadAnomalyDataset} aims to test the perception of unusual dangers on the road. It comprises 60 images curated from the website and rescaled to 1280×720, each featuring inherent anomalies. While seven main categories of anomalies are covered, such as animals and unknown types of vehicles, all anomalies are grouped into a single semantic category. All other scene elements are treated as background. Semantic and instance-level masks are \href{https://www.epfl.ch/labs/cvlab/data/road-anomaly/}{provided} to distinguish anomalous individuals from the background. Nevertheless, the limited scale, heterogeneous web-collected samples, and coarse one-class labeling of this dataset may reduce its reliability and diagnostic value of systematic evaluation, making it more suitable as a supplementary test resource.

\paragraph{\textbf{RoadAnomaly21:}}
\label{dataset:RoadAnomaly21}
As another subset of \cite{chan2021segmentmeifyoucan}, RoadAnomaly21 dataset is proposed for general anomaly segmentation in full street scenes. It contains 32 images from \cite{RoadAnomalyDataset} and 78 online images (10 for validation and 68 for test), resulting in two resolutions, 1280×720 and 2048×1024. Compared with \cite{RoadAnomalyDataset}, the diversity of anomalies increases to 26 types. Different from \textit{Fishyscapes} \cite{fishyscapes} in which anomalous objects are synthetic, anomalies in \textit{RoadAnomaly21} \cite{chan2021segmentmeifyoucan} are intrinsic parts of images, but fall outside the 19 classes in Cityscapes \cite{Cityscapes}. In addition, anomalies can appear anywhere in the image and differ in size, ranging from 0.5\% to 40\% of the image. Since authors believe that identifying all anomalous regions is more important than distinguishing each anomalous object, just semantic-level masks are created to segment the anomalous region from the background and void area. However, only annotations for validation images are \href{https://segmentmeifyoucan.com/datasets}{released}. Despite limitations similar to those of \cite{RoadAnomalyDataset}, the broad variation in anomaly category, position, scale, and appearance makes it a challenging test resource to evaluate robustness and generalizability.

\paragraph{\textbf{Fishyscapes (FS):}}
\label{dataset:fishy}
As a pioneering benchmark for uncertainty estimation in urban driving, FS focuses on semantic segmentation \cite{fishyscapes}. We discuss its two most popular subdatasets. \textbf{\textit{i)}} \textbf{\textit{FS Static}} synthesizes the anomalies by overlaying objects extracted from Pascal VOC \cite{PascalVOCDATASET} with appropriate positions and sizes on the background images employing from Cityscapes \cite{Cityscapes} validation set. The introduced anomalies, such as aircraft, cat, and sofa, do not correspond to any of the 19 predefined classes in Cityscapes \cite{Cityscapes}, nor do they typically appear in urban driving environments. It \href{https://fishyscapes.com/dataset}{publishes} 30 images with semantic mask labels for validation and provides 1000 hidden images for test. Meanwhile, \textbf{\textit{ii)}} \textbf{\textit{FS Lost \& Found}} is derived from LostAndFound \cite{lostAndFound} but only keeps images where foreground anomalies do not fit in any Cityscapes \cite{Cityscapes} classes. Images where anomalies are bikes or children are also filtered out. Consequently, it \href{https://fishyscapes.com/dataset}{releases} 100 images for validation and 275 hidden images for testing. All anomalies are treated as a single semantic category, labeled by a semantic mask to distinguish them from the \textit{background} that includes standard Cityscapes \cite{Cityscapes} classes and the \textit{void area} that covers any original background stuff absent from Cityscapes \cite{Cityscapes}. With its large-scale and complementary synthetic and real-world urban driving scenarios, FS serves as a valuable benchmark for evaluation.

\paragraph{\textbf{STU:}}
\label{dataset:stu}
As the first publicly available dataset for 3D anomaly segmentation, \href{https://omnomnom.vision.rwth-aachen.de/data/stu-dataset/}{STU} \cite{nekrasov2025stu} is collected using eight hardware-triggered
cameras and a LiDAR sensor. All sensors are mounted on the vehicle and synchronized with each other. Seventy sequences are recorded for validation (19) and test (51) and two additional sequences without anomalies are used for training. The anomalies stem from two sources. One is intentionally introduced by authors during the low-speed driving in the controlled environment and has various sizes (e.g., computer monitor), the other is discovered during the normal driving in the public roads (e.g., bucket). Three semantic categories are adopted in the annotation protocol. Objects expected on the road (e.g. pedestrians and vehicles) are annotated as inliers, and objects seen in the training set but to be ignored are labeled as unlabeled (e.g., lamp and parking meters), while anomalies are highlighted as outliers. Each anomaly in every sequence receives a unique instance label, while inliers are annotated at the semantic level, resulting in panoptic-level annotation overall. By incorporating both deliberately introduced and naturally occurring anomalies across urban and rural environments, STU achieves a better balance between coverage and realism. Its multimodal design further broadens existing evaluation settings beyond conventional 2D imagery, bringing benchmarking closer to real autonomous-driving conditions.

\subsection{Anomaly Due to Event} 
\label{subsec:anomaly_event}
Traffic anomalies also include anomalous events that occur unexpectedly and deviate from normal traffic flow, such as road construction, traffic congestion, traffic accidents, and traffic crimes. These events could rapidly change traffic conditions and significantly affect road safety. Specifically, road construction events usually involve frequent entry and exit of large vehicles and equipment and introduce additional obstacles and narrow lanes, leading to an increased level of driving complexity. Meanwhile, traffic congestion reduces visibility and magnifies driver stress. It can lead to sudden braking and aggressive driver behavior, thereby increasing the risk of rear-end collisions and other emergencies. Furthermore, traffic accidents introduce potential hazards to road users and could lead to secondary accidents due to debris or rubbernecking. In addition, crimes that occurred in traffic scenarios such as robbery and theft can distract the attention of surrounding road users and exacerbate their unpredictable decision making, increasing the probability of potential chain-reaction incidents. Hence, attentions and responses are required from drivers and Advanced Driver Assistance Systems (ADAS) towards these anomalous traffic events. A notable distinction is that according to our anomaly taxonomy, if the ego vehicle is the injured party in an accident, we conceive of the element that warrants attention in that scenario as anomalous kinematic patterns of the at-fault party. Thus, datasets only contain ego-involved accident footage are excluded from this section. Cases where only the ego vehicle is at fault (e.g., fatigue driving) are excluded from our study because of our focus on understanding the external driving environment. Only if the ego vehicle is approaching or witnessing a traffic incident scene, the element warranting attention is classified as the anomalous event. In addition, while based on our taxonomy, other emergency events, such as natural disasters and medical emergencies, are classified into categories such as abnormal weather conditions or abnormal kinematic patterns, respectively, these anomalies are excluded from this study due to the lack of relevant studies and datasets.

\subsubsection{Road Construction Detection}
\label{task:road_construction}
Vision-based road construction detection essentially applies object detection to construction scenes with task-specific labels, aiming to identify and localize road work sites, personnel \cite{PARK2023104856}, or equipment \cite{xuehui2021dataset} such as excavators \cite{Feng2024}. Since existing datasets are collected from construction sites \cite{xuehui2021dataset, duan2022soda} or online resources \cite{Feng2024}, they exhibit a clear domain shift from our focus on urban street scenes and are therefore not further analyzed in this paper.

\subsubsection{Traffic Congestion Classification}
\label{task:traffic_congestion}
It aims to recognize abnormal traffic flow patterns, characterized by increased vehicle density, reduced speeds, and longer travel times on road networks. Traditional methods largely rely on the construction and maintenance of hardware facilities, such as induction coils and GNSS. Vision-based approaches discriminate the existence or severity of congestion or describe the congestion process involved in traffic surveillance or dashcam frames \cite{8451957,s24134272,9901471}, using traffic parameters such as vehicle count, vehicle area ratio, and lane change rate \cite{9901471}.

\input{figures_figures/2_traffic_congestion_classification}

This section discusses two datasets, \textit{Chinese City Traffic Image Database (CCTRIB)} \cite{CCTRIB} and \textit{University at Albany DEtection and TRACking (UA-DETRAC)} \cite{UA-DETRAC}. Fig.~\ref{fig:congestion_datasets_comparison} presents two sample images from each dataset with overlaid ground-truth annotations. Specifically, labels for CCTRIB are extracted from TXT file and translated from Chinese into English, while bounding box coordinates and category labels for UA-DETRAC are extracted from JSON document. 

\paragraph{\textbf{CCTRIB:}}
\label{dataset:CCTRIB}
\href{http://www.openits.cn/openData4/824.jhtml}{CCTRIB} \cite{CCTRIB} contains 8471 training images and 729 testing images extracted from traffic surveillance videos of urban roads and highways in China. Resolutions vary from 480×320 to 1920×1080. The ratio between congestion images and non-congestion images in either entire dataset or each subset are 1:1. Labels include binary temporal, ternary illumination, and binary congestion status classification of each image, which, however, may not provide sufficient granularity. For example, despite distinct disparities between the two images in Fig.~\ref{fig:congestion_datasets_comparison}, both were identically labeled as `Congestion' at `Night' with `Low Illumination'. Moreover, the black bars used for privacy protection may adversely affect feature extraction and representation learning. In addition, the labels are written in simplified Chinese, which may restrict the accessibility and utilization of this dataset in the global research community.

\paragraph{\textbf{UA-DETRAC:}}
\label{dataset:UA-DETRAC}
UA-DETRAC \cite{UA-DETRAC} is a representative dataset originally designed for vehicle detection and tracking. It \href{https://www.kaggle.com/datasets/dtrnngc/ua-detrac-dataset}{provides 140,131 frames} captured with the Canon camera in 24 locations at Beijing and Tianjin in China, covering 3 traffic scene types (highway, traffic crossings, and T-junctions), and 4 illumination conditions. UA-DETRAC is valuable for developing and evaluating traffic congestion classification algorithms, specially for addressing the low resolution of traffic surveillance videos. Meanwhile, compared to vehicle-mounted cameras that suffer from a limited field of view and significant blind spots when recording in congested scenes, elevated camera views provide superior coverage \cite{UA-DETRAC}. Although the raw images offer clear advantages, domain-shifted labels introduce both feasibility and challenges. As shown in Fig.~\ref{fig:congestion_datasets_comparison}, static cars (e.g., cars parked on the road side) are not annotated and thus are reasonably not a factor affecting road congestion. However, on the other hand, although the inherent annotation (in bounding boxes) created for four classes of vehicles only complies with the dominant role of vehicles in traffic congestion, the impacts of other road users (e.g. motorcycles) are overlooked. In this context, several strategies can be adopted to address the lack of dedicated labels during the cross-domain application. Firstly, proxy metrics can be derived from the existing spatial annotations and might combine classic traffic flow factors (e.g., vehicle density or total vehicle amount) with the corresponding threshold defined, to classify the binary congestion status or multilevel congestion severity. These factors can also be used to design rule-based models or to provide references for manual annotations. Furthermore, transfer learning, unsupervised learning, or self-supervised learning are also viable approaches.

\subsubsection{Traffic Accidents Classification}
\label{task:accident_classification}
Traffic accidents classification aims to categorize traffic incidents, collisions, or near-miss events from visual data sources such as surveillance and dashcam footage, into predefined binary (accident/no accident) or multiple (e.g. type or severity of crash) classes. In this section, we analyze two relevant datasets, \textit{\underline{T}raffic \underline{A}ccident \underline{D}etection \underline{D}ataset} (TADD) \cite{TADD} and \textit{\underline{C}ar \underline{A}ccident \underline{D}etection and \underline{P}rediction} (CADP) \cite{CADPDataset}, and illustrate their representative frames extracted from raw videos with the corresponding annotations in Fig.~\ref{fig:TADD_visualization} and Fig.~\ref{fig:CADP_visualization}.

\input{figures_figures/3_traffic_accidents_classification_TADD}

\paragraph{\textbf{TADD:}}
\label{dataset:tadd}
Dashcam and traffic surveillance videos at various resolutions collected from \cite{CarCrashDataset} or online platforms are covered \cite{TADD}. The \href{https://ieee-dataport.org/documents/traffic-accident-detection-video-dataset-ai-driven-computer-vision-systems-smart-city}{released package} contains 3912 (1468 non-accident) training, 1052 (466 non-accident) validation, and 726 (349 non-accident) test images, slightly different from the amount reported in \cite{TADD}. Each video is classified as normal traffic or one of seven types of traffic accidents, allowing the learning of the differentiation between normal and abnormal traffic conditions or the classification of crash types. The category of each video is inferred from its parent directory name. However, several issues exist in the current categorizations. First, there are discrepancies between the intrinsic accident type and its label. As revealed in Fig.~\ref{fig:TADD_visualization} (a), videos in other types (e.g., side impact) are frequently misclassified as the `Backend' or `Frontend' class, while various types of incidents (e.g. near-miss, side-wipe, and rear-end cases) are frequently categorized into the `General Augmented Crash' class. Secondly, as revealed in Fig.~\ref{fig:TADD_visualization} (b), while a video may encompass multiple types of crash with heterogeneous features, TADD assigns only one category label to the entire video. These issues may introduce semantic ambiguity and potential learning noise, necessitating careful video subdivision and rigorous category refinement before potential application. In addition, the lack of temporal-level labels makes the classification of each frame infeasible.

\input{figures_figures/4_traffic_accidents_classification_CADP}

\paragraph{\textbf{CADP:}}
\label{dataset:cadp}
CADP \cite{CADPDataset} aims to cover road collisions between pedestrians and vehicles \cite{shah2018accident}. It contains \href{https://docs.google.com/document/d/12F7l4yxNzzUAISZufEd9WFhQKSefVVo_QsPdTsWxZh8/edit?tab=t.0}{1416 accident video segments} selected from 230 YouTube videos, with annotations for 205 videos. However, some are irrelevant to road collisions. For example, Fig.~\ref{fig:CADP_visualization} (a) shows a typical disparate scenario, while Figs.~\ref{fig:CADP_visualization} (b) and (c) reveal non-vehicular transport and near-miss incidents. Furthermore, as presented by the four sequential sample frames in Fig.~\ref{fig:CADP_visualization} (d), extracted 
from the same segment, a significant portion of CADP inappropriately mixes multiple irrelevant video sequences. Moreover, although temporal and spatial labels are created at the start and end frames for the collision participants, as clarified in \cite{CADPReadMe} and demonstrated in Fig.~\ref{fig:CADP_visualization} (e), the bounding boxes and temporal labels are dummies, and the participant class is undefined, limiting CADP to binary video-level classification. These data quality issues necessitate additional comprehensive video preprocessing and meticulous annotation prior to potential application.

\subsubsection{Traffic Accident Anticipation}
\label{task:accident_anticipation}
Traffic accident anticipation, conceptualizing traffic accidents as a special type of event, classifies the `future' event based on the clues in the `current' observation \cite{DAD}. It heavily relies on temporal modeling to analyze behavioral patterns and precursor events in real-time video streams, enabling the prediction of accidents before occurrence \cite{DRIVE}.

Following our established taxonomy, this section analyzes three representative datasets that cover accidents witnessed by the ego vehicle and are applicable to anticipating accidents. \textit{\underline{C}ar \underline{C}rash \underline{D}ataset} (CCD) \cite{CarCrashDataset} labels the onset and duration of accidents, satisfying the basic task requirements. \textit{\underline{D}ashcam \underline{A}ccident \underline{D}ataset} (DAD) \cite{DAD}, while lacking explicit temporal labels, supports this task through detailed frame-by-frame annotations of participant involvement and spatial localization. \textit{\underline{A}n\underline{A}n \underline{A}ccident \underline{D}etection} (A3D) \cite{A3D} enables both accident classification and anticipation through its video-level accident classes and frame-level accident occurrence labels. 

\input{figures_figures/5_traffic_accidents_anticipation_CCD}

\paragraph{\textbf{CCD:}}
\label{dataset:CCD}
\href{https://github.com/Cogito2012/CarCrashDataset?tab=readme-ov-file#download}{CCD} \cite{CarCrashDataset} supports the anticipation of car accident probability by providing 1500 dashcam videos of car accidents from YouTube channels and 3000 non-accident videos from BDD100K \cite{BDD100K}. For each accident video, frame-level label annotates when accident occurs, while video-level label categorizes environment attributes (day/night and weather) and ego-vehicle involvement. Since accident causation labels and annotations for semantic category and spatial location of accident participants are not yet available, CCD is currently limited to binary classification, such as identifying whether the footage contains a traffic accident or if a crash involving the ego vehicle. Fig.~\ref{fig:CCD_visualization} presents two accident scenarios, one involving the ego vehicle and the other between other cars. For each scenario, three representative frames are sampled, i) initial, ii) pre-collision, and iii) collision frames. Specifically, for frames ii) and iii), we superimpose labels that are extracted from the TXT document and interpreted for better understandability. We introduce an additional progress bar at the bottom that indicates the current frame and the duration of the accident. 

\input{figures_figures/6_traffic_accidents_anticipation_DAD}

\paragraph{\textbf{DAD:}}
\label{dataset:DAD}
DAD \cite{DAD} \href{https://github.com/smallcorgi/Anticipating-Accidents}{contains 1750 clips} for accident detection and anticipation, each with 100 frames at a resolution of 1280×720. The training set contains 455 of 620 positive clips and 829 of 1130 negative clips. Extracted from 620 online dashcam videos recorded in six major cities in Taiwan, China, the clips reflect distinctive regional characteristics. First, the extensive adoption of high-resolution dashcams across various vehicle types, including motorcycles, provides viewpoints beyond standard car-mounted cameras. Moreover, motorcycles are involved in 68.2\% of collisions. DAD also treats several atypical yet safety-critical events as positive cases, including traffic robbery (e.g. training clip 77), near-miss events (e.g. testing clip 60), single-vehicle crashes due to loss of balance or poor collision anticipation (e.g. training clips 24 and 340), collisions with static objects such as barrier (e.g. training clip 450), and deliberate accident scam (e.g. training clip 354). Meanwhile, frame-based labels for moving objects, including the bounding box, semantic category, tracking ID, and binary crash involvement status, facilitate object tracking and support accident detection and anticipation. Fig.~\ref{fig:DAD_visualization} visualizes the labels of incident participants and introduces a progress bar for better temporal understanding. However, DAD does not explicitly provide temporal labels to specify the start and end frame of the incident, although relevant information can be roughly inferred from the frame-to-frame involvement status of incident participants. In addition, while statistics are summarized in \cite{DAD}, DAD lacks labels that describe accident types for each frame or video. The annotation does not specify how many categories of moving objects are involved. 

\paragraph{\textbf{A3D:}}
\label{dataset:A3D}
A3D \cite{A3D} contains 1500 dashcam accident videos (128,175 frames) compiled from YouTube channels, 40\% of which are recorded from a third-person perspective and thus belong to the anomalous event category. Although A3D includes non-collision frames within each clip, it does not provide videos of normal traffic scenes. Therefore, experiments requiring such scenes must rely on supplementary datasets, such as HEV-I dataset \cite{HEV-I} used in \cite{A3D}. Meanwhile, A3D provides binary labels indicating ego-vehicle involvement at the video level and the accident occurrence at the frame level. Despite the 18 claimed types of accidents \cite{A3D}, these multiclass categorical labels are absent from the annotation file. Furthermore, all \href{https://github.com/MoonBlvd/tad-IROS2019}{204 provided YouTube links} have become inaccessible, preventing any possible data retrieval, which precludes the visualization and in-depth analysis of this dataset in our study.

\subsubsection{Accident Causality Recognition}
\label{task:accident_causality}
Beyond classification, accident causality recognition is a task to further categorize causal relationships, analyzing the sequence of events, behaviors, and factors that lead to the accident. Identifying the spatial locations of accident participants is also important for enabling a more comprehensive analysis of behavioral and trajectory patterns. By contrast, while some aforementioned benchmarks support object-level detection, they are limited to basic semantic categorization of participants (e.g., car, pedestrian). Consequently, they are suitable for identifying which objects are involved in an incident, but not for explaining why those objects are involved in or lead to it.

Three datasets are classified into this category, \textit{\underline{D}etection \underline{o}f \underline{T}raffic \underline{A}nomaly} (DoTA) \cite{DoTA}, \textit{\underline{C}ausality in \underline{T}raffic \underline{A}ccidents (CTA)} \cite{you2020CTA}, and \textit{\underline{M}ulti-\underline{M}odal \underline{A}ccident video \underline{U}nderstanding (MM-AU)} \cite{MM-AU}. Figs.~\ref{fig:DoTA_visualization} and~\ref{fig:CTA_visualization} present representative images of these datasets and superimpose the corresponding labels for illustration.

\input{figures_figures/7_accident_causality_recognition_DoTA}

\paragraph{\textbf{DoTA:}}
\label{dataset:DoTA}
Developed by the same team of A3D \cite{A3D} in 2020 and published in 2023,
\href{https://github.com/MoonBlvd/Detection-of-Traffic-Anomaly}{DoTA} \cite{DoTA} contains 4677 dashcam accident videos collected from YouTube channels and 731,932 frames with 1280×720 pixels. It adopts a What-Where-When pipeline to identify: i) the dominant collision type, ii) the location of collision participants within relevant frames, and iii) the moment when the collision became inevitable and when all anomalies leave the view or become stationary. Each video is categorized into one of 18 collision types according to ego vehicle involvement and the class or status of the accident parties. Collision frames are assigned a unique accident ID and a category name, while non-collision frames are labeled as `normal' with an accident ID of 0. Each participant in a collision frame is further annotated with a bounding box, tracking ID, and semantic label, enabling spatial-temporal tracking across seven common road-user categories. Fig.~\ref{fig:DoTA_visualization} (a) presents sample frames extracted from an accident video. According to the overlaid annotations, this accident starts at the 53rd frame (top-right) and ends at the 66th frame (bottom-middle). However, the vehicle with abnormal behavior already rushes out in the 52nd frame (top-middle) and continues its motion in the 71st frame (bottom-right). Moreover, spatial annotations disappear for one accident participant in the frame preceding the officially labeled end (bottom-left) and are missing for both accident parties in the labeled conclusion frame.

\input{figures_figures/8_accident_causality_recognition_CTA}

\paragraph{\textbf{CTA:}}
\label{dataset:CTA}
CTA \cite{you2020CTA} decomposes each accident into a pair of events, cause and effect, depicting the pre-crash status (e.g., changing lanes) or risky behavior (e.g., running during red light), and the resulting outcome for the involved party (e.g., rollover), respectively. It selects 1935 accident videos from YouTube channels, predominantly at resolutions of 1280×720 and 640×360. Instead of providing pre-extracted clips, CTA \href{https://github.com/tackgeun/CausalityInTrafficAccident/tree/master/dataset}{releases} video IDs and temporal annotations in PKL files, requiring users to retrieve the videos and extract the corresponding segments themselves. The 18 cause categories and 7 effect classes are derived from real-world traffic accident statistics. As indicated in Fig.~\ref{fig:CTA_visualization}, CTA contains surveillance (12\%) and dashcam (88\%) footage, accidents with or without ego involvement, and near-miss incidents, and labels include clip ID, time stamps, and categorization. We use the progress bar in Fig.~\ref{fig:CTA_visualization} to show the current frame and the duration of the cause and effect segments. However, the lack of instance-level labels, such as the location of accident participants in each frame, limits CTA to a basic classification task, instead of analyzing the entity at fault, the injured party of the accident, and their effects on the accident.

\input{figures_figures/9_accident_causality_recognition_MMAU}

\paragraph{\textbf{MM-AU:}}
\label{dataset:MM-AU}
\href{https://huggingface.co/datasets/JeffreyChou/MM-AU}{MM-AU} \cite{MM-AU} focuses on accident perception from ego-centric videos, curating 2,195,613 frames in 11,727 videos from four datasets \cite{DoTA, CarCrashDataset, A3D, DADA-2000} and three websites. Fig.~\ref{fig:mmau_visualization} (a) shows the decoding of multifaceted categorical labels in the Excel file provided for granular accident analysis, including 1) scene attributes (4 weather conditions, day/night, 5 scenario types, and 5 road geometries), 2) binary accident occurrence (1 for all videos due to exclusive focus on accident events), 3) temporal key moments (anomalous object appearance, accident onset, and anomaly end), and 4) causality (58 accident categories, causal factors, and preventive strategies). MM-AU also provides 422489 images, among which 391,152 have object detection annotations in \href{https://huggingface.co/datasets/JeffreyChou/MM-AU/blob/main/MMAU_Det_paper/labels.tar.gz}{JSON format}, with an average of 5.71 objects per image, a maximum of 29, and 46,693 images containing only one annotated object. 
However, MM-AU exhibits some annotation inaccuracies and inconsistencies. First, as revealed in Fig.~\ref{fig:mmau_visualization} (a), the cause is labeled as a motorcycle driving too fast, although the more direct cause appears to be wrong-way travel. Second, the anomalous behavior has already become obvious in the 37th frame before the annotated start frame (44th). Similarly, comparison with the final video frame (212th) shows that the motorcycle becomes stationary until the 186th frame, rather than the annotated end frame (140th). Meanwhile, Fig.~\ref{fig:mmau_visualization} (b) reveals inconsistent object annotation practices. Riders are sometimes labeled as `person' with the vehicle omitted (left images), whereas in other cases the rider and vehicle are jointly labeled as `motorcycle' (top-right images). 
In the bottom images, the bridge is incorrectly labeled as `car', while the partial car is mislabeled as `truck'. In addition, the ego vehicle is annotated as `car' in some images, but omitted in most cases.

\subsubsection{Traffic Crime Classification}
\label{task:traffic_crime}

Beyond traffic regulation violations identified as anomalies through kinematic patterns, criminal incidents, such as robberies occurring within traffic environments, can also pose significant risks to road safety and therefore merit investigation. Although dedicated benchmarks are limited, existing datasets containing annotated criminal events, such as \textit{UCF-Crime} \cite{Sultani_2018_CVPR}, \textit{\underline{M}ulti-\underline{S}cenario \underline{A}nomaly \underline{D}etection} (MSAD) \cite{MSAD}, and DAD \cite{DAD}, may be applicable to this task.

\paragraph{\textbf{UCF-Crime:}}
\label{dataset:UCF-Crime}
It \cite{Sultani_2018_CVPR} \href{https://www.kaggle.com/datasets/odins0n/ucf-crime-dataset}{comprises 1900} (950 positive and 950 negative) untrimmed surveillance videos collected from YouTube and LiveLeak. All videos are resized to 320×240 pixels. Each video may contain one or two criminal events of the same type. In addition to \textit{road accident}, 12 more crime classes are covered, such as \textit{assault} and \textit{robbery}
\cite{Sultani_2018_CVPR}. Some crimes occur in non-traffic environments, such as apartments or shops, making the corresponding videos unsuitable for traffic crime classification. Fig.~\ref{fig:Crime_visualization} (A) displays sample frames extracted from footage containing criminal events in traffic scenarios, including road accidents, shooting, and fighting. For videos accompanied by temporal annotations in the TXT file, the video name, crime category, and event duration are overlaid onto the corresponding frames to enhance visualization clarity. Otherwise, only categorical labels are highlighted in the category list. Furthermore, as shown in Fig.~\ref{fig:Crime_visualization} (A), road accident videos in UCF-Crime are not exclusively drawn from traffic surveillance footage.
Despite covering diverse scenario types and supervision footage, the limited amount of traffic-scenario crime footage provides insufficient domain-specific supervision and its relatively low resolution further restricts access to fine-grained visual cues. Consequently, it is unsuitable as a standalone training or evaluation resource for traffic crime classification.

\input{figures_figures/10_traffic_crime}

\paragraph{\textbf{MSAD:}}
\label{dataset:MSAD}
MSAD \cite{MSAD} aims to support the classification of diverse anomalous events in various environments, containing 240 positive and 480 negative videos, \href{https://msad-dataset.github.io/}{exclusively available for non-commercial usage} based on the request. Various resolutions are involved, with the predominant one at 1280×720. As shown in Fig.~\ref{fig:Crime_visualization} (B), the labels specify the total frame count for each video, and the start and end frames of the anomalous event. Although MSAD contains 14 scenario types and 11 anomalous event classes, the limited number of positive videos within each category restricts the learning of distinctive features for robust classification. Meanwhile, the small portion of crime footage (e.g., fighting and assault) in traffic scenes (e.g., highways and roads) undermines its potential as a standalone dataset for traffic crime classification. The atypical instances in the 39 traffic accident videos, such as a vehicle collision in a store, further exacerbate the challenge of developing generalized traffic crime classification algorithms.

\paragraph{\textbf{DAD:}}
As mentioned in \ref{dataset:DAD}, DAD \cite{DAD} includes a small proportion of intentional accident fraud and traffic-related crime footage, such as robbery. Although it only supports binary classification due to the lack of specific crime category labels, its spatial annotations enable further detection of parties involved in the criminal incidents.

\subsection{Anomaly Due to Kinematic Pattern} \label{subsec:anomaly_kinematic}
Deviations from normal kinematic patterns, characterized by unexpected velocity transitions, erratic behaviors, or non-standard spatial trajectories, provide critical probabilistic indicators of potential emerging events or systemic irregularities. Hence, the identification and prediction of these anomalous movement signatures serve as a sophisticated diagnostic mechanism to enable proactive intervention strategies. Despite the predominance focus of existing methods on spatial characteristics or temporal patterns of kinematic data \cite{10.1186/s13677-024-00611-1}, in this article we focus on vision-based benchmarks. In addition, as previously clarified, this section centers on the ego-involved emergent incidents arising from anomalous kinematic patterns, rather than pre-existing accidental events.

\subsubsection{Collision Prediction}
\label{task:collision_prediction}
Maintaining safe interactions with dynamic road users, such as vehicles, riders, and pedestrians, presents significantly greater complexity compared to navigating static obstacles. The inherent unpredictability of moving entities \cite{ALOZI2024104572}, particularly when their trajectories become erratic or deviate from expected kinematic patterns, introduces substantial computational and perceptual challenges for collision avoidance systems. By assessing factors such as trajectory intersections, spatio-temporal relative distance, and behavior patterns, the prediction of collision probabilities or the recognition of potential collisions can be enabled \cite{CarlaSyn}. It is worthy to emphasize again that the term `collision' in this section refers in particular to the interaction between the ego vehicle and other dynamic road agents. The incidents in which the ego vehicle functions as a third-party observer, with minimal potential for direct involvement, are systematically categorized as anomalous events and have been comprehensively examined in the preceding sections.

Candidate datasets can be divided into three categories based on their annotation schemes. The first type offers only temporal information and binary classification labels, denoting `\textit{when a collision occurs}', such as the ego-involved collision footage in \hyperref[dataset:CCD]{CCD} \cite{CarCrashDataset} and \hyperref[dataset:cadp]{CADP} \cite{CADPDataset}. The second type supports `\textit{when-and-what}' analysis by further incorporating categorical labels. The ego-involved single-vehicle accidents in \hyperref[dataset:DoTA]{DoTA} \cite{DoTA} fall into this category, as spatial annotations are not provided for these cases. Although \hyperref[dataset:tadd]{TADD} \cite{TADD} contains relevant ego-involved collision footage for this task, it currently lacks temporal annotations for the collision start frame. \hyperref[dataset:CST-S3D]{CST-S3D} \cite{CST-S3D} to be analyzed belongs to this class. The third type provides spatial information of road users in each frame, facilitating a deeper understanding of movement patterns and interactions beyond basic visual features. The collision footage between the ego and other vehicles in \hyperref[dataset:DoTA]{DoTA} \cite{DoTA} and \hyperref[dataset:DAD]{DAD} \cite{DAD} falls into this category. 
Despite their eligibility, the datasets \cite{DoTA, DAD, CADPDataset, CarCrashDataset, TADD} analyzed in previous sections are not discussed further in this section to avoid redundancy.

\input{figures_figures/11_collision_prediction_CST-S3D}
\paragraph{\textbf{CST-S3D:}}
\label{dataset:CST-S3D}
To address inconsistencies in DADA-2000 \cite{DADA}, \cite{CST-S3D} introduces enhanced annotations for 704 ego-involved accident videos (while this count is not specified in \cite{CST-S3D}). We denote this refined version as \textit{CST-S3D} \cite{CST-S3D}, and illustrate sample frames of incidents arising from anomalous behavior and trajectory in Fig.~\ref{fig:CST_visualization} (a). \cite{CST-S3D} uses a hierarchical classification system with 4, 7, and 16 categories to classify accidents with increasing levels of detail. However, some labels are inconsistent. In addition to the example revealed in Fig.~\ref{fig:CST_visualization} (c), videos ranging from {\small `00\_00\_02\_05\_0072'} to {\small `00\_00\_02\_05\_0120'} are labeled `\textit{Hitting pedestrian}' in the 4-class scheme but as `\textit{Hitting motorbike}' in both 7-class and 16-class schemes even though an equivalent label exists in all three schemes. In addition, the temporal interval between key frames in some videos is narrow. As shown in Fig.~\ref{fig:CST_visualization} (b), the rider with anomalous trajectories appears in the 7th frame (left), while the collision begins in the 12th frame (middle), and the consequence becomes irreversible in the 17th frame (right). By the time such anomalies are successfully identified, the remaining time for implementing evasive strategies becomes limited, setting higher requirements for predictive algorithms.

\subsubsection{Behavior Recognition}
\label{task:behavior_recognition}
Behavior recognition is a critical perception task that identifies dangerous behaviors of surrounding road agents from an ego-centric perspective, such as sudden lane changes and erratic pedestrian trajectories, to anticipate and prevent potential traffic accidents. Existing models focus on visual feature extraction, temporal sequence understanding, and complex interaction modeling.
Two representative datasets, \textit{YouTubeCrash} and \textit{GTACrash}, were introduced in \cite{CrashToNotCrash} for the task of dangerous vehicle recognition. It should be noted that while the abnormal behavior of pedestrians and cyclists recorded in JAAD \cite{JAAD} and PIE \cite{PIE} (e.g., the second scenario in Fig.~\ref{fig:JAAD_PIE_visualization}) is compatible with the current category, they focus on `\textit{intentions}' and have a large proportion of proper behavior footage. Hence, we analyze them in Section~\ref{subsec:Pertinency_Location}.

\input{figures_figures/12_behavior_recognition_YouTube_GTA}

\paragraph{\textbf{YouTubeCrash:}}
\label{dataset:YouTubeCrash}
\href{https://sites.google.com/view/crash-to-not-crash}{YouTubeCrash} \cite{CrashToNotCrash} collects dashcam YouTube videos to evaluate dangerous vehicle classification algorithms. From each video, a 2-second segment preceding the accident is extracted as the “accident clip,” while a segment from the beginning is taken as the “non-accident clip.” If these segments overlap, the non-accident clip is discarded. This process yields 122 positive and 100 negative clips, with 20 frames extracted from each clip and evenly split between the training and test sets. YouTubeCrash treats the vehicle as the sole semantic class of analysis, containing various erratic vehicle behaviors, such as abrupt lane change, sudden stops, and traffic signal violations \cite{CrashToNotCrash}. Fig.~\ref{fig:YouTubeCrash_visualization} (a) shows that a small portion of the footage captures accidents in which the ego vehicle was not involved in the initial accident but was affected by subsequent events. YouTubeCrash annotates the temporal-spatial position and binary dangerous category of each vehicle. Since each accident clip deliberately ends at the start frame of the accident, the final positions of dangerous vehicles also indicate the location of the accident, providing a geometric and temporal reference point for the collision prediction task. However, Fig.~\ref{fig:YouTubeCrash_visualization} (a) reveals the impossibility of distinguishing vehicles that receive impact in the accident from irrelevant vehicles, since they are all designated as negative samples.

\paragraph{\textbf{GTACrash:}}
\label{dataset:GTACrash}
\href{https://sites.google.com/view/crash-to-not-crash}{GTACrash} complements YouTubeCrash \cite{CrashToNotCrash} through synthetic game scenes. It contains 7720 accident and 3661 non-accident videos, with 20 frames extracted from each video. GTACrash covers a wide range of scenarios, including three weather conditions, different times of day and night, and various road types such as highways and arterials. However, since the simulator cannot replicate the full spectrum of accident patterns, some typical types of real-life accidents are excluded, such as sudden stops. Despite this inherent limitation, GTACrash illustrates that larger synthetic datasets can compensate for real-world data scarcity \cite{CrashToNotCrash}. In addition, GTACrash gives various labels, including temporal-spatial location and multiple vehicle attributes for a more comprehensive analysis.

\subsubsection{Driver Attention Prediction}
\label{task:driver_attention}
The driver attention prediction aims to identify and locate regions or objects in the driving environment that are likely to attract driver attention and influence driving behavior \cite{DADA,Xia2017PredictingDA}. Although the benchmark annotations are commonly derived from driver gaze behavior and fixation patterns \cite{palazzi2018predicting,DADA,Xia2017PredictingDA}, this perception task predicts attention over external scenes rather than inferring internal driver states and thus is within the scope of our study. Such annotated visual cues may support the development of more effective early warning and intervention mechanisms for road safety. In this section, we discuss datasets depicting critical situations, namely \textit{\underline{B}erkeley \underline{D}eep\underline{D}rive \underline{A}ttention} (BDD-A) \cite{Xia2017PredictingDA}, and accident scenarios, namely \textit{DADA-2000} \cite{DADA}. Fig.~\ref{fig:attention_datasets_comparison} presents sample frames selected from both datasets, overlaid by the corresponding ground truths in the heatmap format with greater transparency for better visualization.

\input{figures_figures/13_driver_attention}

\paragraph{\textbf{BDD-A:}}
\label{dataset:bdda}
\href{https://bdd-data.berkeley.edu/}{BDD-A} \cite{Xia2017PredictingDA} explores the attention regions in braking events where vehicle decelerates by more than 1 m/s$^2$ or its speed drops below 2 m/s. As a result, 1435 videos with 720×1280 pixels are selected from BDD100K \cite{BDD100K}, with 926 for training, 203 for validation and 306 for testing. The ground-truth attention heat maps are derived from the gaze patterns of 45 participants in laboratory-based eye tracking experiments. The central region (576×1024) of each gaze-map video aligns with the full camera video frame. The dense annotations in BDD100K \cite{BDD100K} enable finer-grained identification of attended objects if the overlap between an attention region and an object region exceeds the threshold. Although in-lab tracking may beat in-car collection in some respects, it has several inherent limitations. First, some essential contextual information can be filtered out. Second, authentic driver responses that would have occurred in real traffic scenarios cannot be fully replicated in the lab environment. For example, as shown in the first and third scenarios in Fig.~\ref{fig:attention_datasets_comparison}, the derived attention is not allocated to the approaching attention-worthy oncoming vehicle. Compared to the second scenario, attention shifts away in the third scenario even though the pedestrian is still crossing.

\paragraph{\textbf{DADA-2000:}}
\label{dataset:dada2000}
Collecting from YouTube and three Chinese video platforms, \href{https://github.com/JWFangit/LOTVS-DADA}{2000 videos with 658,476 frames} at a resolution of 1584×660 \cite{DADA} are selected for \underline{d}river \underline{a}ttention prediction in \underline{d}riving \underline{a}ccident scenarios \cite{DADA-2000}. A smaller package with 1013 videos is also \href{https://opendatalab.com/OpenDataLab/DADA2000/tree/main/raw}{released}. As revealed in Fig.~\ref{fig:attention_datasets_comparison}, DADA-2000 contains footage in which the ego vehicle witnessed the accident occurrence (left), was involved in or caused an accident (middle), or nearly missed an incident (right). These videos are classified into 54 types based on the semantic categories of accident participants and the involvement of the ego vehicle. The annotations reportedly incorporate pixel-level annotations derived from eye fixation data and frame-by-frame binary accident occurrence labels. However, object-level information, such as which object attracts the attention, is not provided. Furthermore, there are some inconsistencies in the DADA-2000 annotations. As stated in \cite{CST-S3D}, some annotations are inaccurate, particularly in classification, stemming from two main sources: misidentification of ego involvement and incorrect categorization of accident participants. In addition, the recorded driver eye movement data only shows where the driver looked, not where they should have focused for safety \cite{mine}. Therefore, it may be inadequate for supervising safety-critical attention modeling. For example, in the left image of Fig.~\ref{fig:attention_datasets_comparison}, the driver fixates on the vehicle responsible for the initial accident, overlooking approaching motorcycles and pedestrians that could lead to a secondary collision. In the right image, the driver focuses on a woman beside the car rather than the child crossing the road, who has just caused a near-miss incident.

\subsection{Anomaly Due to Status or Conditions} \label{subsec:anomaly_status}
In this paper, we focus on three types of abnormal status and conditions that can be encountered while driving. 1) \textit{Abnormal road conditions}, particularly surface defects, including longitudinal and transverse cracks, alligator and block cracks, and potholes \cite{RoadDamageSurvey, RDD2022}, increase the probability of accidents \cite{RoadDamageSurvey}. 2) \textit{Abnormal weather conditions}, such as heavy rain or fog, could alter the appearance of scene entities, reducing the visibility of human drivers, and also introducing noise into vision-based algorithms \cite{10115323}. 3) \textit{Anomalous illumination conditions}, including extreme glare, sudden lighting changes, and insufficient lighting, usually contain both under- and overexposures, seriously degrading the visual appearances and structures of scene entities \cite{nightcity} and thus posing the challenges for effective identification. This section discusses benchmarks for perceiving and adapting to these environmental and infrastructure anomalies.

\subsubsection{Crack Segmentation}
\label{task:crack_segmentation}
Pavement crack is one of the most common and significant forms of surface distress \cite{NHA12D}. The identification of cracks is essential for the maintenance and monitoring of the pavement \cite{CrackTree}. Crack segmentation classifies 2D images of the pavement surface into a binary mask, where crack pixels are differentiated from the background \cite{NHA12D}, without identifying the severity or specific type of crack. This section examines six publicly available datasets.

\input{figures_figures/14_road_damage}
\paragraph{\textbf{CrackTree260:}}
\label{dataset:CrackTree260}
CrackTree260 \cite{CrackTree}, proposed in 2012, contains 260 pavement images, each captured under natural visible light conditions using an area-array camera \cite{Deepcrack}. Each image contains one or more crack types. While \cite{CrackTree} reports a unified resolution for all images at 800×600, \href{https://github.com/qinnzou/DeepCrack?tab=readme-ov-file}{the publicly available package} provided by \cite{Deepcrack} contains 206 images at 800×600 and 54 images at 960×720. Pixel-level annotations are manually drawn along the crack curves. Due to the size limitation, data augmentation might be needed when using CrackTree260. For example, \cite{Deepcrack} employs a combination of rotation, flipping, and multi-region patch extraction, resulting in a total of 35,100 training images.

\paragraph{\textbf{\underline{C}rack\underline{F}orest \underline{D}ataset (CFD):}}
\label{dataset:CFD}
CFD \cite{CrackForest}, introduced in 2016, \href{https://github.com/cuilimeng/CrackForest-datase}{contains 155 images} of the urban road surface captured by the iPhone5 in Beijing, China. Except for the last two images with a resolution of 320×480, all other images have been resized to 480×320. Manual pixel-level annotations are available only for the first 118 images in MAT format, which further limits CFD as a standalone supervision source and as a comprehensive evaluation benchmark.

\paragraph{\textbf{CRACK500:}}
\label{dataset:CRACK500}
CRACK500 \cite{CRACK500}, proposed in 2019, \href{https://github.com/fyangneil/pavement-crack-detection?tab=readme-ov-file}{contains 500 images of pavement cracks} captured by cell phones on the main campus of Temple University. It involves 7 different resolutions, ranging from 3264×2448 (36 files) to 2560×1440 (435 files). The split ratio of images among training, validation, and test sets is 5:1:4. Pixel-level annotations are provided in PNG binary maps and in TXT files for training and test sets, while validation images are accompanied solely by binary mask files. However, 50 mask files in training set lack their counterparts. To expand the dataset while optimizing computational efficiency, CRACK500 uses an augmentation strategy that splits each original image into 16 non-overlapping patches and keeps only those with more than 1000 crack pixels, yielding 1896 images for training, 348 for validation, and 1124 for testing, accompanied by their binary masks. As shown in Fig.~\ref{fig:road_damage_datasets_comparison}, some noises, such as shadows and stains on road surfaces, are also captured in the images to evaluate the robustness of the model.

\paragraph{\textbf{GAPs384:}}
\label{dataset:GAPs384}
Selecting from \href{https://www.tu-ilmenau.de/neurob/data-sets-code/german-asphalt-pavement-distress-dataset-gaps}{German Asphalt Pavement Distress (GAPs) dataset} \cite{GAPs}, GAPs384 \cite{CRACK500} \href{https://github.com/fyangneil/pavement-crack-detection}{contains 384 crack distress images} with pixel-wise annotations. A sample image of subtle cracking damage is shown in Fig.~\ref{fig:road_damage_datasets_comparison}. A cropped version with 509 images is proposed to reduce the computational cost required by the original 1920×1080 size, dividing each image into six regions and retaining only those with more than 1000 crack pixels. Although the cropped resolution is reported as 640×540, our analysis indicates that 228 images actually have a resolution of 540×440.

\paragraph{\textbf{EdmCrack600:}}
\label{dataset:EdmCrack600}
Collected in Edmonton, Canada in 2020, EdmCrack600 \cite{EdmCrack600} \href{https://github.com/mqp2259/EdmCrack600}{provides 600 images} at 1920×1080 and their binary crack masks. A rear-mount GoPro camera is used to avoid the effects of the windshield and irrelevant pixels. As shown in Fig.~\ref{fig:road_damage_datasets_comparison}, EdmCrack600 not only captures road users, roadside infrastructure, and lane markings, but also reflects changes in illumination and weather conditions. Although the same research team introduced a larger version, EdmCrack1000 \cite{EdmCrack1000}, in 2019, we have excluded it from our analysis because it is not yet publicly available.

\paragraph{\textbf{NHA12D:}}
\label{dataset:NHA12D}
\href{https://github.com/ZheningHuang/NHA12D-Crack-Detection-Dataset-and-Comparison-Study}{NHA12D} \cite{NHA12D} is collected by \underline{N}ational \underline{H}ighways digital survey vehicles on the network \underline{A12} in UK and released in 2022. It consists of 80 pavement images at 1920×1080 and their binary masks. Images are equally distributed between concrete and asphalt surfaces, with each category containing 25 vertical-view images and 15 forward-view images. Each image focuses on individual driving lane, with incidental inclusion of various road elements, including vehicles, lane markings, road studs, water stains, and miscellaneous objects \cite{NHA12D}. Although the intensity value of positive pixels in the masks is reported as 255 \cite{NHA12D}, as shown in Fig.~\ref{fig:road_damage_datasets_comparison}, pink or green is actually used to denote the crack regions.

\subsubsection{Pothole Segmentation}
\label{task:Pothole_Segmentation}
A pothole represents a distinctive category of road distress, characterized by its irregular morphology and structural degradation of the pavement surface \cite{8788687}. Despite the importance of precise delineation, pothole segmentation remains an underexplored research domain that requires specialized datasets and further investigation \cite{8788687}.

\paragraph{\textbf{Pothole-600:}}
\label{dataset:Pothole-600}

\input{figures_figures/15_pothole_segmentation_Pothole-600}

Introduced in 2019, Pothole-600 \cite{Pothole600} \href{https://sites.google.com/view/pothole-600/dataset}{contains 67 image pairs} captured by a vehicle-mounted ZED stereo camera. It has three subsets with respective resolutions of 1028×1730, 1030×1720, and 1028×1710. A ground-truth binary map and an estimated disparity map are provided for each image. Although \cite{Pothole600} presents Pothole-600 as a pothole detection dataset, we reclassify it as a segmentation dataset because its pixel-level labels enable precise delineation of pothole regions rather than simple detection. As illustrated in Fig.~\ref{fig:pothole_visualization}, pothole-600 also includes several challenging images with noise (e.g. water stains), supporting the development and evaluation of robust algorithms.

\subsubsection{Road Damage Classification}
\label{task:road_damage_classification}
Road damage classification categorizes pavement images based on the presence, type, and severity of surface deterioration. It can be considered an important preparation for the localization and segmentation of pavement damage \cite{CQU-BPDD}. However, despite the presence of multiple types of pavement distress within the existing dataset, \textit{\underline{C}hong\underline{Q}ing \underline{U}niversity \underline{B}ituminous \underline{P}avement \underline{D}isease \underline{D}etection} (CQU-BPDD) \cite{CQU-BPDD}, only binary classification labels are available to distinguish between distressed and normal pavement images.

\input{figures_figures/16_road_damage_classification}

\paragraph{\textbf{CQU-BPDD:}}
\label{dataset:CQU-BPDD}
CQU-BPDD \cite{CQU-BPDD} consists of 60,059 bituminous pavement images collected by vehicle-mounted cameras on highways in southern China. Each image with a resolution of 1200×900 corresponds to a 2×3 meters pavement patch. The folder name in each subset is given as a classification label.
Patch-based annotation provides a naive location of the damage based on the relationship between each batch and its original pavement. Despite coverage of seven types of pavement damage, including transverse crack (1039), massive crack (6207), alligator crack (851), crack pouring (4199), longitudinal crack (2382), raveling (957), and repair (1094), most images (43,330 of 60,059) are negative samples. As revealed in Fig.~\ref{fig:cqu_visualization}, some patches in different categories share similar characteristics, making them difficult to distinguish. Some labeled normal images actually contain road damage. In addition, \href{https://github.com/DearCaat/CQU-BPDD}{the exclusive download method} makes CQU-BPDD less accessible to researchers from other regions.

\subsubsection{Road Damage Segmentation}
\label{task:road_damage_segmentation}
Road damage segmentation delimits road surface damage in pavement images or even videos \cite{SHREC}. Existing studies have focused predominantly on potholes and cracks \cite{RoadDamageSurvey}. Although these datasets, as analyzed above, can support the segmentation of cracks or potholes, comprehensive datasets that cover all types of road damage remain extremely limited. Before M2S-RoAD \cite{M2S-RoAD} which contains multimodal data with various road damages, SHREC2022 \cite{SHREC} was the only comprehensive option for pixel-wise classification, though it is primarily an assemblage of existing datasets.

\input{figures_figures/17_road_damage_segmentation_SHREC}
\paragraph{\textbf{PotholeMix:}}
\label{dataset:PotholeMix}
\href{https://data.mendeley.com/datasets/kfth5g2xk3/2}{PotholeMix} \cite{SHREC} (also called {\small SHREC2022} \cite{RoadDamageSurvey}) comprises an image collection and a video collection. The image collection consists of 4,340 RGB images and mask counterparts, where masks for images from five public datasets \cite{CRACK500,GAPs,EdmCrack600,Pothole600,cpiri} were derived from their original annotations, while those for the web-sourced images were manually created by the authors. As shown in Fig.~\ref{fig:shrec_visualization}, a unified color scheme was implemented, using red and blue to represent cracks and potholes, respectively. Meanwhile, the video collection comprises 797 RGB-D video pairs for users to extract additional images, while only 359 mask videos are provided. The last three images in Fig.~\ref{fig:shrec_visualization} visualize the triplet of a sample frame.

\input{figures_figures/18_road_damage_segmentation_M2S-RoAD}
\paragraph{\textbf{M2S-RoAD:}}
\label{dataset:M2S-RoAD}
M2S-RoAD \cite{M2S-RoAD}, introduced in 2024, is distinguished by its focus on rural environments, its road defect taxonomy, and its multimodal data. It \href{https://huggingface.co/datasets/DarrylT/M2S-RoAD/tree/main}{combines 1071 RGB images} (1928×1208) collected by a vehicle-mounted front-view camera and 43911 point-cloud frames recorded by a synchronized LiDAR sensor. Data is collected in four towns in New South Wales, Australia, and under varying weather and illumination conditions to ensure diversity. As shown in Fig.~\ref{fig:m2s_visualization}, images are classified into four subsets according to the weather conditions. M2S-RoAD features polygon-based delineation of damage boundaries and encompasses nine distinct categories of pavement deterioration in its annotation schema, including pothole, potholes with water, cracks, corrugation, rutting, slippage, drain, bump, and manhole. Fig.~\ref{fig:m2s_visualization} shows the slippage annotations in the image and the point cloud.

\subsubsection{Road Damage Detection}
\label{task:road_damage_detection}
Road damage detection focuses on localizing and classifying pavement deterioration. Specialized challenges, such as \underline{C}rowd sensing-based \underline{R}oad \underline{D}amage \underline{D}etection \underline{C}hallenge (CRDDC) \cite{10021040}, have been designed to advance this research direction. Driven by progress in computer vision and machine learning, image-based benchmarks now dominate this field \cite{RoadDamageSurvey}. Representative datasets include \textit{\underline{G}erman \underline{A}sphalt \underline{P}avement Distres\underline{s}} (GAPs) \cite{GAPs} and RDD2022 \cite{RDD2022}. Nevertheless, given the sensitivity of cameras to weather and lighting conditions, some studies have also employed depth-based sensors, such as LiDAR \cite{RoadDamageSurvey}.

\begin{figure}[!t]
    \centering
    \includegraphics[width=0.95\columnwidth,height=0.15\columnwidth]{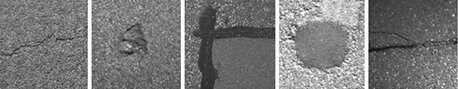}\\
    \caption{GAPs \cite{ExtendGAPs} illustration (from left to right): crack, pothole, invalid patch, applied patch, and open joints.}
    \label{fig:gaps_visualization}
\end{figure}
\paragraph{\textbf{GAPs:}}
\label{dataset:GAPs}
Originally introduced in 2017, GAPs \cite{GAPs} contains 1969 grayscale images collected from three federal roads in Germany and was extended to 2468 images (1920×1080) in 2019 with additional images from a new federal highway \cite{ExtendGAPs}. The annotations have also been optimized by enclosing the distress with a tighter bounding box to restrict the pixels of non-damage area. Although \cite{ExtendGAPs} theoretically categorizes pavement distress into six classes, 89.71\% of GAPs depict intact roads, and bleeding is excluded from the actual annotation due to the absence of acquired images. Compared to the original version that only offered patches in a size of 64×64, the extended version offers several patch sizes that show more context information, resulting in 692,377 patches of surface defects and 6,035,404 patches of the intact road. In addition, a smaller subset containing 50,000 samples is also provided to support fast experiments. \href{https://www.tu-ilmenau.de/en/university/departments/department-of-computer-science-and-automation/profile/institutes-and-groups/institute-of-computer-and-systems-engineering/group-for-neuroinformatics-and-cognitive-robotics/data-sets-code/german-asphalt-pavement-distress-dataset-gaps}{All versions are only provided to academics upon request.}

\input{figures_figures/20_road_damage_detection_RDD}
\paragraph{\textbf{RDD2022:}}
\label{dataset:rdd2022}
Released as part of CRDDC \cite{10021040}, RDD2022 \cite{RDD2022} represents the latest iteration of the RDD series and thereby we focus exclusively on this version. RDD2022 extends the scope to 47,420 road images collected from six countries to support global road damage detection. Despite only four resolutions reported in \cite{RDD2022}, \href{https://github.com/sekilab/RoadDamageDetector?tab=readme-ov-file}{released package} covers ten varieties, ranging from 512×512 to 4040×2035. As shown in Fig.~\ref{fig:rdd_visualization}, images are gathered using various devices, capturing various illumination and weather conditions \cite{RDD2022} and annotations created only for training images denote damages using enclosed bounding boxes and class labels. Although \cite{RDD2022} only claims four categories, including longitudinal cracks (\textit{D00}), transverse cracks (\textit{D10}), alligator cracks (\textit{D20}), and potholes (\textit{D40}), Fig.~\ref{fig:rdd_visualization} reveals more classes, such as \textit{Repair} and \textit{D50}. RDD2022 also includes images that do not have identifiable road damage (e.g., the last image in Fig.~\ref{fig:rdd_visualization}). In addition, some images employ gray or black masks to obscure certain areas (e.g., the first image in Fig.~\ref{fig:rdd_visualization}), requiring pre-processing before the application to avoid introducing noises to model learning.

\subsubsection{Traffic Weather Recognition}
\label{task:Traffic_Weather_Recognition}
 Although recognition of weather conditions has potential practical values in driving assistance and traffic management \cite{TWData}, it has been mainly treated as a dynamic environmental factor to evaluate the robustness and adaptation capabilities of other traffic-related tasks, such as semantic segmentation under varying weather conditions \cite{ACDC, 10115323}. Only a limited number of works specifically focus on the recognition of weather conditions in traffic scenes, and most approaches utilize conventional computer vision techniques or deep learning models adapted from general image classification tasks \cite{TWData}. 

Despite the lack of dedicated benchmarks \cite{TWData}, traffic datasets incorporating multiple weather conditions, such as WildDash \cite{WildDash} and nuScenes \cite{nuScenes}, can be leveraged by creating the corresponding labels. Datasets, such as \textit{\underline{A}dverse \underline{C}onditions \underline{D}ataset with \underline{C}orrespondences} (ACDC) \cite{ACDC}, specifically designed for perception tasks in adverse weather conditions provide a more viable solution. We only detail this dataset in this section. Meanwhile, synthesized datasets created via weather generation algorithms \cite{10115323,qian2024weatherdg} represent a promising alternative approach to address data scarcity. Several derivative datasets of Cityscapes \cite{Cityscapes}, including Foggy Zurich \cite{Foggy1}, Foggy Cityscapes \cite{Foggy2}, and RainCityscapes \cite{Rainy1}, available on the Cityscapes website, were created synthetically by adding a fog or rain layer to the original clear-weather Cityscapes images to simulate adverse weather conditions.

\input{figures_figures/21_weather_recognition_ACDC}
\paragraph{\textbf{ACDC:}}
\label{dataset:ACDC}
To facilitate semantic segmentation algorithms under adverse visual conditions, \href{https://acdc.vision.ee.ethz.ch/download}{ACDC collects} 1000 foggy, 1000 rainy, 1000 snowy and 1006 nighttime images, covering urban areas, highways and rural regions in Switzerland \cite{ACDC}, and adopts 19 semantic classes in Cityscapes \cite{Cityscapes}. The footage was recorded via GoPro and extracted at a resolution of 1920×1080. Fig.~\ref{fig:acdc_visualization} presents a representative image per adverse condition category. The even distribution across three weather categories makes ACDC adoptable for both traffic weather recognition and evaluating the robustness of perception algorithms.

\subsubsection{Illumination Perception and Adaption}
\label{task:illumination}
Accurate perception of illumination allows automatic activation of night mode without driver intervention and adaptation of headlight beam patterns based on road conditions, contributing to improved road safety while minimizing glare for other road users \cite{9449667}. Meanwhile, significant research attention has focused on robust perception algorithms against adverse illumination conditions \cite{9011028}, such as semantic segmentation in low light environments \cite{9784832, darkzurich}, nighttime pedestrian detection \cite{NightOwls}, and traffic sign recognition in intense sun glare \cite{Glare}. Under such conditions, sensor performance is often compromised by multiple degradation factors, including underexposure, increased noise, reduced color fidelity, and motion blur \cite{9784832}. 

Existing datasets focus primarily on nighttime scenarios. Since constructing an accurate and effective low-illumination dataset of real-world scenes requires costly annotation \cite{9784832}, some datasets are created using simulators, such as Synthesiscarla \cite{9784832}, or are synthesized by applying algorithmic transformation of daytime images, such as Cyclecity \cite{9784832}. However, real-world datasets remain crucial for recording authentic lighting conditions and validating model performance under nighttime conditions, such as aforementioned ACDC \cite{ACDC}. This section analyzes real-life datasets, including \textit{NightDriving} \cite{NightDriving}, \textit{DarkZurich} \cite{darkzurich}, \textit{NightCity} \cite{nightcity}, while \textit{NightOwls} \cite{NightOwls} will be examined in the pedestrian detection section due to its specific focus. Meanwhile, glare from the sun that can cause temporary blindness and impair vision sensors \cite{Glare} is also explored by several datasets, such as Glare \cite{Glare} that will be discussed in the traffic sign detection section considering its essential focus.

\input{figures_figures/22_illumination}
\paragraph{\textbf{NightDriving:}}
\label{dataset:nightdriving}
Introduced in 2018, NightDriving \cite{NightDriving} contains recordings from several areas of Switzerland, acquired with a GoPro camera at a resolution of 1920×1080. According to \cite{NightDriving}, 35000 images are extracted and partitioned into five classes based on illumination conditions, including daytime, civil twilight, nautical twilight, astronomical twilight and nighttime. However, \href{https://people.ee.ethz.ch/~daid/NightDriving/}{official link} \cite{NightDriving} has become unavailable, and only the 50 nighttime test images with annotations are \href{https://www.kaggle.com/datasets/ipythonx/nighttime-driving-dataset}{accessible online}. The pixel-level annotation, aligned with the Cityscapes schema \cite{Cityscapes}, covers 19 semantic categories. Nevertheless, the annotation may suffer from some errors. For example, as shown in Fig.~\ref{fig:illumination_visualization} (a), the pedestrian is incorrectly classified into the same category as the sidewalk.

\paragraph{\textbf{NightCity:}}
\label{dataset:NightCity}
Proposed in 2021, NightCity \cite{nightcity} \href{https://dmcv.sjtu.edu.cn/people/phd/tanxin/NightCity/index.html}{extracts 4297} (2998 training and 1299 validation) images from online ego-centric videos and unifies them at 1024×512 pixels. The footage includes more than 13 cities in 9 countries, ensuring substantial geographic and weather diversity, including challenging snow-covered streets. Manual annotation adopts the same annotation schema as Cityscapes \cite{Cityscapes} to categorize each valid pixel into one of 19 semantic classes. However, labeling poorly illuminated images remains a challenge. As shown in Fig.~\ref{fig:illumination_visualization} (a), despite the presence of multiple vehicles ahead of the ego vehicle, only the vehicle in the adjacent lane is correctly annotated, while pixels of other vehicles are classified erroneously as background.

\paragraph{\textbf{DarkZurich:}}
\label{dataset:darkzirich}
Released in 2022, DarkZurich \cite{darkzurich} captures 8779 images in Zurich using a 1080p GoPro camera mounted above the front windshield, together with the corresponding camera GPS coordinates. The training set contains 3041 daytime, 2920 twilight, and 2416 nighttime images. Validation and test sets include 50 and 151 daytime-nighttime image pairs, respectively. The annotation is enhanced by comparing each nighttime image with its daytime counterpart. Notably, instead of labeling training images, DarkZurich creates pixel-level annotations for nighttime images in validation and test sets. However, only the validation annotations are \href{https://www.trace.ethz.ch/publications/2019/GCMA_UIoU/}{publicly available}. Meanwhile, validation images predominantly depict sparsely populated street scenes, which may not effectively reflect the true performance of algorithms. Furthermore, as illustrated in Fig.~\ref{fig:illumination_visualization} (b), some objects within these images suffer from motion blur issues which can obscure object boundaries and fine details, thereby increasing the perception difficulty.

\subsection{Anomaly Due to Appearance} \label{subsec:anomaly_appearance}
Unusual or unexpected visual characteristics of road users can indicate potential safety risks, such as pedestrians with suspicious appearance, vehicles with suspected, obscured or altered license plates, and vehicles in a worse state than the minimum safety standard. Although advanced object detection algorithms provide technical foundation for perceiving these anomalies, this field remains constrained by the absence of systematic task taxonomy and comprehensive benchmark datasets. Thus, in the following subsections, we give a concise overview of the relevant tasks and datasets, without elaborating on the specific details in Table \ref{table:visual_datasets_summary}.

\subsubsection{Criminal Recognition}
\label{task:criminal_recognition}
Identifying criminals out of crowds from surveillance footage has become a crucial means in modern criminal investigations. It relies on facial recognition technology to match the captured features with a database of known offenders. Further, as revealed by Lombroso's study \cite{Zebrowitz2008SocialPF}, certain facial features and expressions can be considered reflections of the mind and thoughts and thereby used as clues to identify and predict the criminal tendency of an unknown person \cite{9293926}. However, to our knowledge, due to the special nature of the criminal recognition task, benchmark datasets that contain both criminal face databases and raw surveillance footage are absent. Consequently, researchers resort to creating their own datasets to train and evaluate algorithms, such as \cite{crimeRecognition, 9293926, 10010426}.

\subsubsection{Suspicious Vehicle Recognition}
\label{task:suspicious_veh_recognition}
Suspicious vehicle recognition is critical to target suspected vehicles in dense traffic flow and restricting suspicious and unwanted vehicles in restricted areas \cite{SuspiciousVehicleRecognition}. Considering the generic nature of vehicle exteriors, identifying suspicious vehicles merely by external visual characteristics such as color or size represents a rudimentary approach. In contrast, since license plates offer a unique identifier for each vehicle, license plate identification serves as the primary discriminative mechanism, which uses optical character recognition technologies to extract and process license plate texts. Automatic Number Plate Recognition technology is now the cornerstone of modern traffic enforcement and surveillance systems. For example, if a scanned license plate matches one in a database of vehicles that are uninsured, reported stolen, or associated with other violations, the system automatically generates an alert at the command center. Dispatch operators can then coordinate with nearby patrol units to intercept the vehicle \cite{enwiki:1255542574}. The main sensor applied in real life is infrared cameras to ensure the quality of captured images under different lighting conditions. Due to the absence of benchmark datasets, studies collect images with innocent license plates to develop object detection algorithms \cite{SuspiciousVehicleRecognition}.

\subsubsection{Damaged Vehicle Detection}
\label{task:damaged_veh_detection}
Laws and regulations in different areas impose varying requirements and restrictions on vehicles that are not operating in good condition. Identifying these unreliable vehicles is crucial to prevent potential accidents and hit-and-run incidents. Meanwhile, accurate damage identification is also vital for car insurance settlement claims. In this context, images containing multiple types of vehicles with various external damages are required to enable identification of damaged vehicles, localization of damage, and classification of damage types or severity levels. Despite the absence of benchmark datasets with surveillance or dashcam footage, several datasets with images captured by mobile phones are released for academic use, such as \cite{cardamage1, cardamage2, cardamage3}.

\subsection{Anomaly Due to Multiple Reasons} \label{subsec:anomaly_multiple}
During a journey, the ego vehicle may encounter any of the aforementioned anomalies that can jointly affect road safety. Accordingly, some studies aim to comprehensively assess anomalies in the driving environment rather than focus on a specific type \cite{RiskBench}. This category contains benchmarks if they explicitly analyze or implicitly involve more than two types of anomalies.

\subsubsection{Risk Identification}
\label{task:risk_identification}
Risk identification refers to the process of identifying potential hazards that could influence decision-making \cite{RiskBench}. Traditional evaluation approaches rely on independent datasets for different types of risk, which limits their ability to assess the comprehensive risk handling capabilities of algorithms. Recent studies tend to address this limitation by simultaneously incorporating multiple risk categories \cite{RiskBench}. However, the representative dataset, \textit{RiskBench} \cite{RiskBench}, exclusively labels one predefined type of risks per sample.

\input{figures_figures/23_multiple_reason_riskbench}


\paragraph{\textbf{RiskBench:}}
\label{dataset:riskbench}

\href{https://hcis-lab.github.io/RiskBench/}{RiskBench} \cite{RiskBench} is a scenario-based dataset with 6916 scenarios created using CARLA \cite{CARLA}. It analyzes risks arising from three types of anomalies (in our taxonomy): i) dynamic road users with \textit{abnormal kinematic patterns} that have an interaction with the ego vehicle; ii) static \textit{obstacles}, including illegal parking, street barrier, traffic cone, and traffic warning; and iii) \textit{collisions} indicating an anomalous event. It also has normal traffic scenarios, allowing for a more balanced assessment of both risk detection accuracy and false alarm rates. The design of scenarios is determined by static environmental factors (map, area, and road topology) and dynamic elements (risk type, semantic category of anomalous entity, and behavior of road users) and augmented by various illumination, weather and traffic conditions. Each scenario is extracted as sequential images at a resolution of 640×256. Due to the inherent advantages of the simulator, RiskBench provides various labels, such as distance, spatial coordinates, and instance masks. Fig.~\ref{fig:riskbench_visualization} presents three sample frames, each showing a risk category, and highlights the labeled anomaly. Annotations focus exclusively on pre-designated objects and their associated specific risks, while overlooking other potential hazards in the same or different risk categories. An example is illustrated in Fig.~\ref{fig:riskbench_visualization} (B), where neglected hazards include closer pedestrians and vehicles that could collide with the target vehicle after initial impact. This limitation may create artificial boundaries in risk assessment by focusing on preselected hazards rather than capturing the complex, interconnected nature of real-world risks.

\subsection{Pertinency Due to Spatial Location}\label{subsec:Pertinency_Location}
In driving scenarios, the pertinence of surrounding objects is inherently tied to their location. The spatial proximity and relative position of scene entities to the ego vehicle largely determine their importance for attention. For example, a vehicle in an adjacent lane has varying levels of significance depending on whether it is next to, ahead of, or behind the ego vehicle. This spatial-attention mechanism mirrors both human driving behavior and the requirements for ADAS. Identifying scene entities with a notable spatial relationship to the ego vehicle is a straightforward task to optimize attention allocation, as spatial properties such as distance and orientation are readily quantifiable from sensor data. Moreover, this spatially conditioned relevance can be formalized as a scene classification problem, where traffic scenes are categorized based on the presence and positioning of attention-worthy entities. Lane change is the most common scenario for this task, considering its critical position in real-life accidents \cite{sub-NUDrive}.

\subsubsection{Pedestrian Intention Prediction}
\label{task:pedestrian_intention}
The goal of pedestrian intention prediction is to identify attention-worthy pedestrians whose inferred intentions may result in future trajectories that intersect the path of the ego vehicle. Following \cite{PIE}, we define `\textit{intention}' as an inferred latent state rather than an observable behavior. In addition, \cite{PIE} emphasizes the role of proximity to road elements in intention reasoning. As the utility of pedestrian intention prediction is typically greatest for pedestrians occupying right-of-way or other spatially relevant regions, we categorize this task as \textit{pertinent due to spatial location}. Previous work has mainly examined two critical scenarios: predicting pedestrian crossing intentions at intersections and detecting jaywalking incidents \cite{JAAD,PIE}.

Several datasets have emerged, with three benchmarks standing out for their coverage. \textit{\underline{J}oint \underline{A}ttention in \underline{A}utonomous \underline{D}riving} (JAAD) \cite{JAAD} focuses on the behavioral cues preceding pedestrian crossing decisions and their interactions with the driver. \textit{\underline{P}edestrian \underline{I}ntention \underline{E}stimation} (PIE) \cite{PIE} largely follows JAAD while centering on all pedestrians in proximity to the road and extends the annotations to traffic devices and vehicles. \textit{\underline{S}tanford-\underline{T}RI \underline{I}ntent \underline{P}rediction} (STIP) \cite{STIP} contains urban scenarios recorded by three ego-centric cameras with intention labels and tracking ids to classify pedestrian crossing intentions.

\input{figures_figures/24_pedestrian_intention_prediction_JAAD_PIE}

\paragraph{\textbf{JAAD:}}
\label{dataset:jaad}
JAAD \cite{JAAD} collects 346 ego-centric video clips using dashboard camera or GoPro device across five locations, including North York, Kremenchuk, Lviv, Hamburg, and New York. The predominant resolution is 1280×720, while 64 clips are at 1920× 1080. 
Fig.~\ref{fig:JAAD_PIE_visualization} (a) illustrates four sample scenarios. The two scenarios on the top row represent proper crossing (left) and jaywalking (right) behavior, respectively. In the third scenario, the only person is irrelevant to the ego vehicle, while the vehicles crossing in front are the most critical. In the last scenario, the ego vehicle is at fault for failing to yield the right-of-way to the cyclist.
As illustrated, JAAD contains multidimensional labels. For each clip, contextual attributes are noted, including weather, time of day, and location. For each frame, the binary presence status of critical traffic devices is identified, including zebra crossing, pedestrian sign, stop sign, and traffic light. Within each frame, bounding boxes with an occlusion label are used to enclose instances within three categories, a) \textit{pedestrian} who is the target to attend, b) \textit{ped} who are also individual pedestrians but unlikely impact the ego vehicle (e.g., bystanders), and c) \textit{people} comprising a group of pedestrians. For each pedestrian, binary crossing intention, categorical behaviors, demographics (age group and gender), appearance characteristics, and crossing point are labeled, allowing a parallel and structured representation of the dynamics of pedestrian-vehicle interaction. Specifically, only pedestrians whose crossing trajectory intersects the front path of the approaching vehicle are classified as crossing. Behavior annotations include precrossing action (e.g. standing), attention to the approaching vehicle (e.g. looking), and response to the approaching vehicle (e.g. speed-up). Meanwhile, driver behavior tags capture the state of the approaching vehicle, such as decelerating. 
However, JAAD exhibits a notable class imbalance. The intention of most pedestrians is annotated as crossing or going to cross \cite{9423436}. And discrepancies in labels are revealed by the fourth scenario. First, the cyclist is miscategorized as a pedestrian, and the action is miscategorized as walking, which may compromise the model ability to distinguish object categories and motion patterns. Second, the standing person `ped 0\_66\_309' labeled as `ped1' in the annotation file but `ped2' in the appearance file, which may cause confusion during training.

\paragraph{\textbf{PIE:}}
\label{dataset:PIE}
In contrast to JAAD, PIE \cite{PIE} was collected under a more consistent setting, comprising \href{https://github.com/aras62/PIE}{53 video clips} recorded in Toronto using a monocular dashboard camera at a resolution of 1920×1080. GPS coordinates and vehicle states, such as heading angle and speed, are also recorded by a camera-synchronized on-board sensor. PIE largely follows the multilayered annotation framework of JAAD \cite{JAAD}, but with a different focus. Firstly, PIE provides annotations for all pedestrians in proximity to the road, regardless of their actual behavior with respect to the ego-vehicle. Each pedestrian is annotated with textual labels depicting their actions, gestural responses, attentional orientation, and crossing intentions. Unlike JAAD \cite{JAAD}, PIE does not provide labels for pedestrian appearance. In particular, PIE distinguishes between the intention of crossing and the action of crossing. However, more pedestrians are classified as non-crossing (1322 vs. 512) \cite{9423436}. For instance, in the right scenario in Fig.~\ref{fig:JAAD_PIE_visualization} (b), annotations for the two notable crossing pedestrians are absent, while all labeled pedestrians are identified as non-crossing. Meanwhile, except for crosswalk and pedestrian, PIE offers specific semantic category labels for instances in the other four classes, including vehicle (e.g., car), traffic light (e.g., regular), traffic sign (e.g., construction), and transit station (e.g., bus station). The traffic light status (e.g. red) is also specified. Bounding boxes with a unique ID and occlusion tag are created for objects in these six classes. However, Fig.~\ref{fig:JAAD_PIE_visualization} (b) reveals inaccuracies in the labeled coordinates, which may distort spatial and motion cues and introduce noise into model training.

\input{figures_figures/25_pedestrian_intention_prediction_STIP}
\paragraph{\textbf{STIP:}}
\label{dataset:STIP}
STIP \cite{STIP} contains 1,108,176 frames with 1216×1936 pixels from 8 cities in California and Michigan using three front-facing cameras mounted on the left, center, and right of the vehicle. Fig.~\ref{fig:STIP_visualization} shows the different perspectives of these cameras. STIP provides bounding boxes with binary intention labels and tracking ids for each pedestrian. Fig.~\ref{fig:STIP_visualization} overlays labels on the raw images for illustration, where red boxes indicate crossing pedestrians and green boxes for non-crossing pedestrians. However, the \href{https://stip.stanford.edu/dataset.html}{official dataset request link} has become unavailable due to technical and legal issues, restricting accessibility and limiting wider adoption.

\subsubsection{Pedestrian Pose Estimation}
\label{task:pedestrian_pose}
Pedestrian pose estimation, applying human pose estimation into traffic scenarios, aims to detect and track anatomical keypoints of pedestrians in the footage \cite{pedx,8765346}. Despite significant advances in pose estimation in an indoor or basic outdoor environment \cite{8765346, Guler2018DensePose}, the inherent challenges of dynamic driving scenarios, including varying illumination conditions, frequent occlusions, diverse pedestrian appearances, and complex background contexts, have resulted in a comparative scarcity of comprehensive studies and well-annotated datasets in this domain \cite{pedx, ECPDP}. To our knowledge, the only large-scale datasets that focus on estimating pedestrian poses are \textit{PedX} \cite{pedx}, \textit{Waymo} \cite{Waymo}, and \textit{\underline{E}uro\underline{C}ity \underline{P}ersons \underline{D}ense \underline{P}ose} (ECPDP) \cite{ECPDP}.

\begin{figure}[!t]
    \centering

    \includegraphics[width=0.92\columnwidth, height=0.4\columnwidth]{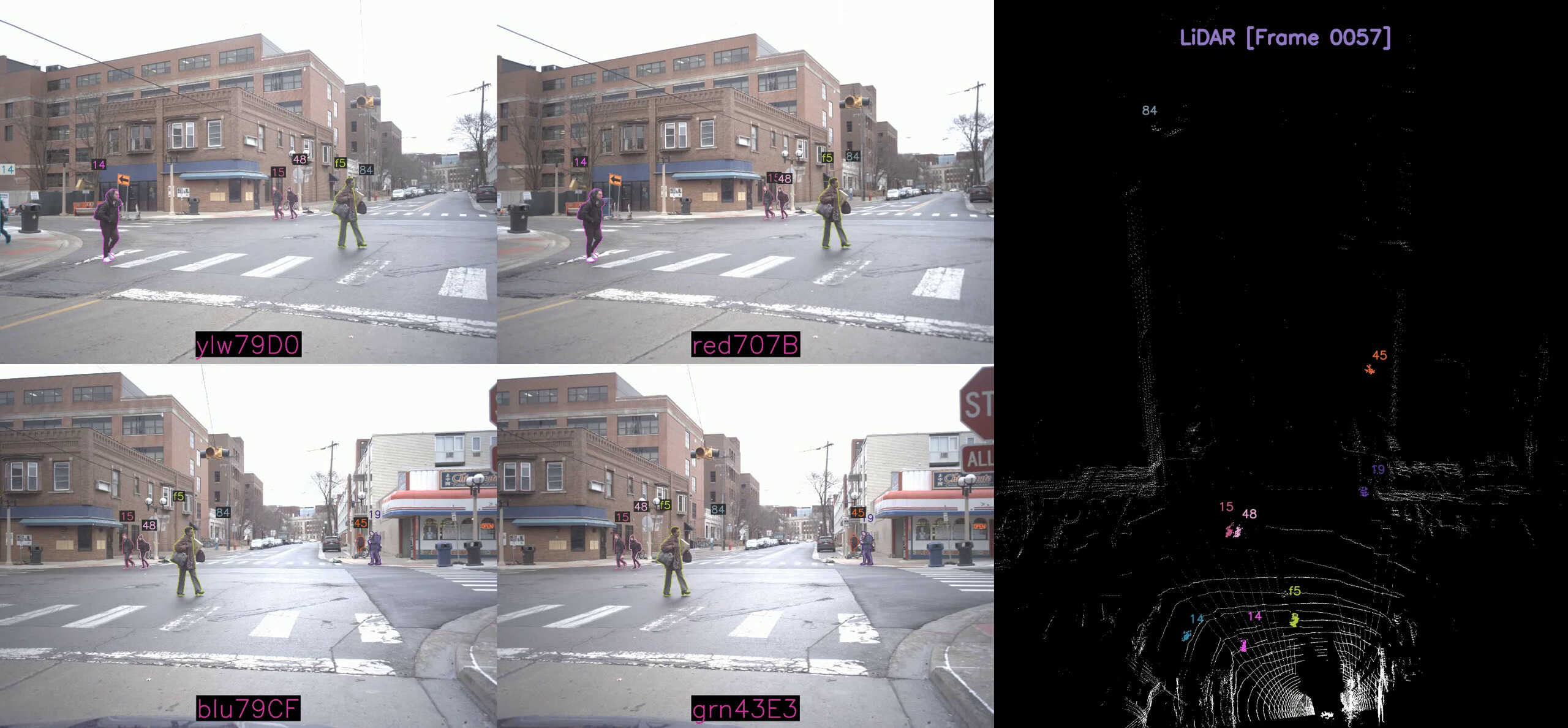}\\
    \caption{Sample annotated frame from \cite{pedx}, captured using calibrated and time-synchronized stereo cameras and LiDAR.}
    \label{fig:PedX_visualization}
\end{figure}
\paragraph{\textbf{PedX:}}
\label{dataset:PedX}
\href{https://deepblue.lib.umich.edu/data/concern/data_sets/6h440s98b?locale=en}{PedX} \cite{pedx} is a large-scale multimodal dataset designed to study pedestrian behavior and pedestrian-vehicle interactions at four-way stop intersections. Two pairs of roof-mounted stereo cameras and LiDAR are calibrated and synchronized to monitor all four crosswalks at each selected intersection in downtown Ann Arbor. PedX contains 5076 pairs of stereo images at an original resolution of 4112×3008 and a rectified resolution of 3678×2668, with 2500 frames of 3D LiDAR point clouds. Multimodal labels comprise a) polygons with tracking IDs for each pedestrian, b) coordinates for its 18 body joints, and c) 3D parameters for pose, shape, and global location. Fig.~\ref{fig:PedX_visualization} illustrates the different perspectives for a sample frame, with the main annotations overlaid. Furthermore, dense information, especially point-cloud trajectories, joint positions, and relative distance, equips PedX with the potential to predict pedestrian intention \cite{pedx} to identify notable pedestrians in intricate intersection environments. 

\paragraph{\textbf{Waymo:}}
\label{dataset:waymo_pose}
\href{https://waymo.com/open/data/perception/}{Waymo} \cite{Waymo} introduces a valuable benchmark for human pose estimation in autonomous driving scenes by extending its perception data with manually annotated keypoints for pedestrians and cyclists in both image and LiDAR modalities \cite{waymo_pose_info}. Specifically, Waymo defines 14 human body keypoints and supplies a visibility/occlusion attribute for each point. Although only 13 keypoints are explicitly listed in the official website and documentation \cite{waymo_pose}, spanning the nose, shoulders, elbows, wrists, hips, knees, and ankles, the 14th keypoint mentioned in the website corresponds to head. Initially collected in three cities, Waymo was later expanded to cover six U.S. cities. Across the 1150 scenes, Waymo derived 172.6K object annotations with 2D keypoints from camera images and 10K object annotations with 3D keypoints from LiDAR point clouds. However, due to the sparse and isolated nature of its pose annotation \cite{waymo_pose_info,waymo_pose_info_2}, Waymo is better suited for semi-supervised or weakly supervised methods. As a result, the number of papers that directly benchmark on its keypoint annotations remains relatively small compared with other Waymo tasks.

\begin{figure}[!t]
    \centering
    
    \begin{tabular}{c@{\hspace{0.01\columnwidth}}c}

        \includegraphics[width=0.45\columnwidth, trim=0 4cm 0 2cm, clip]{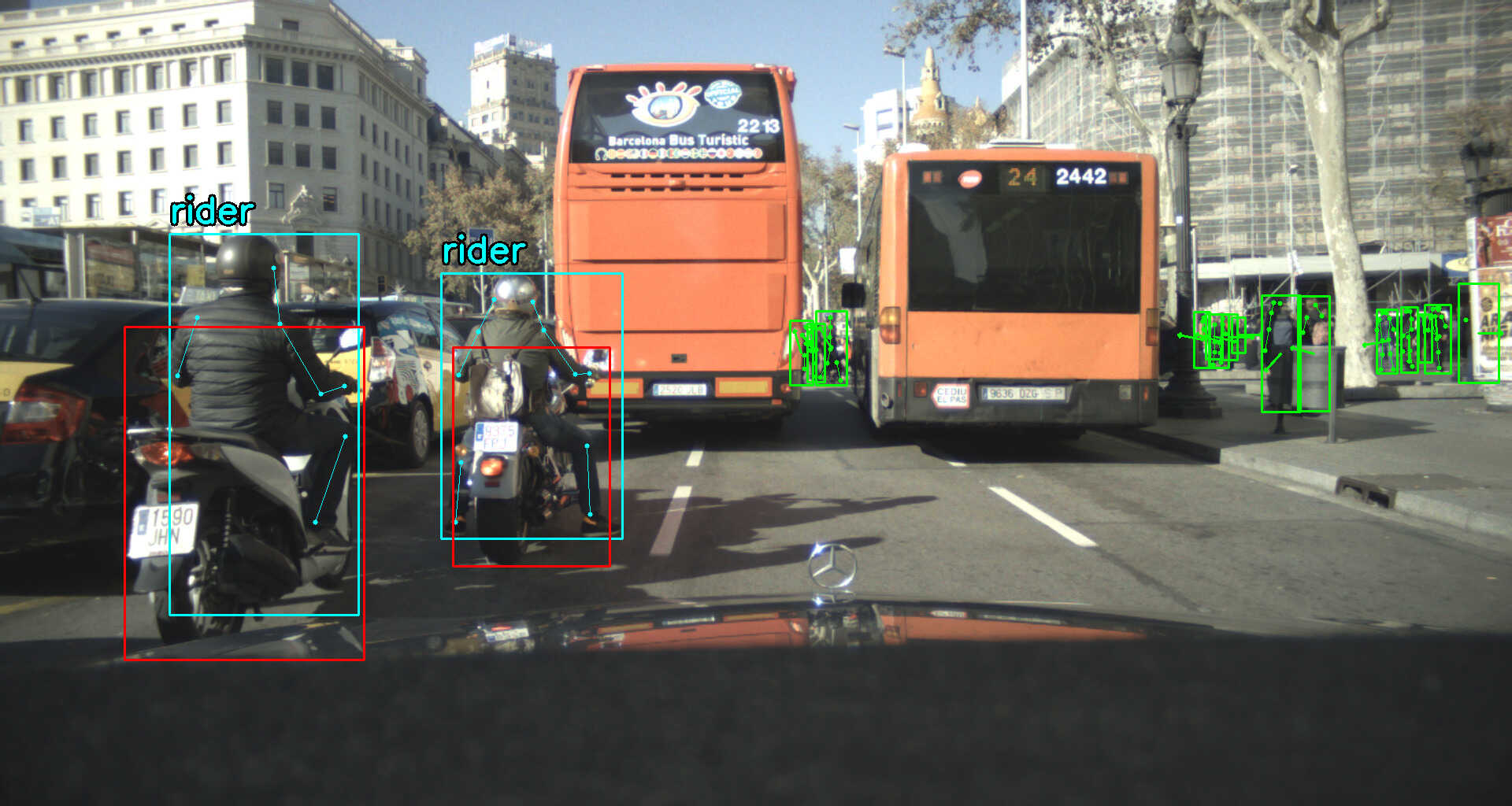} &
        \includegraphics[width=0.45\columnwidth, trim=0 4cm 0 2cm, clip]{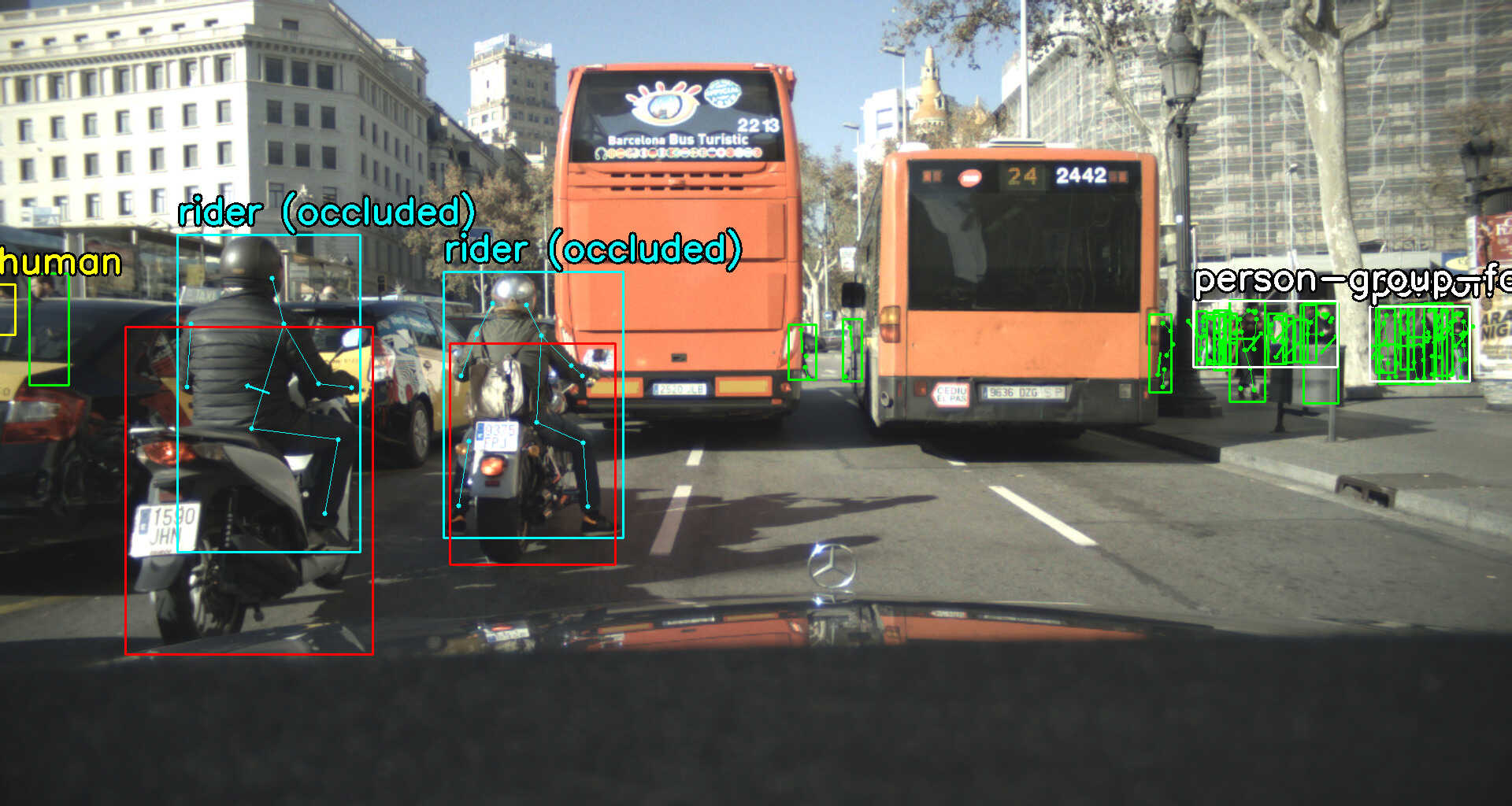} \\

        \includegraphics[width=0.45\columnwidth, trim=0 4cm 0 2cm, clip]{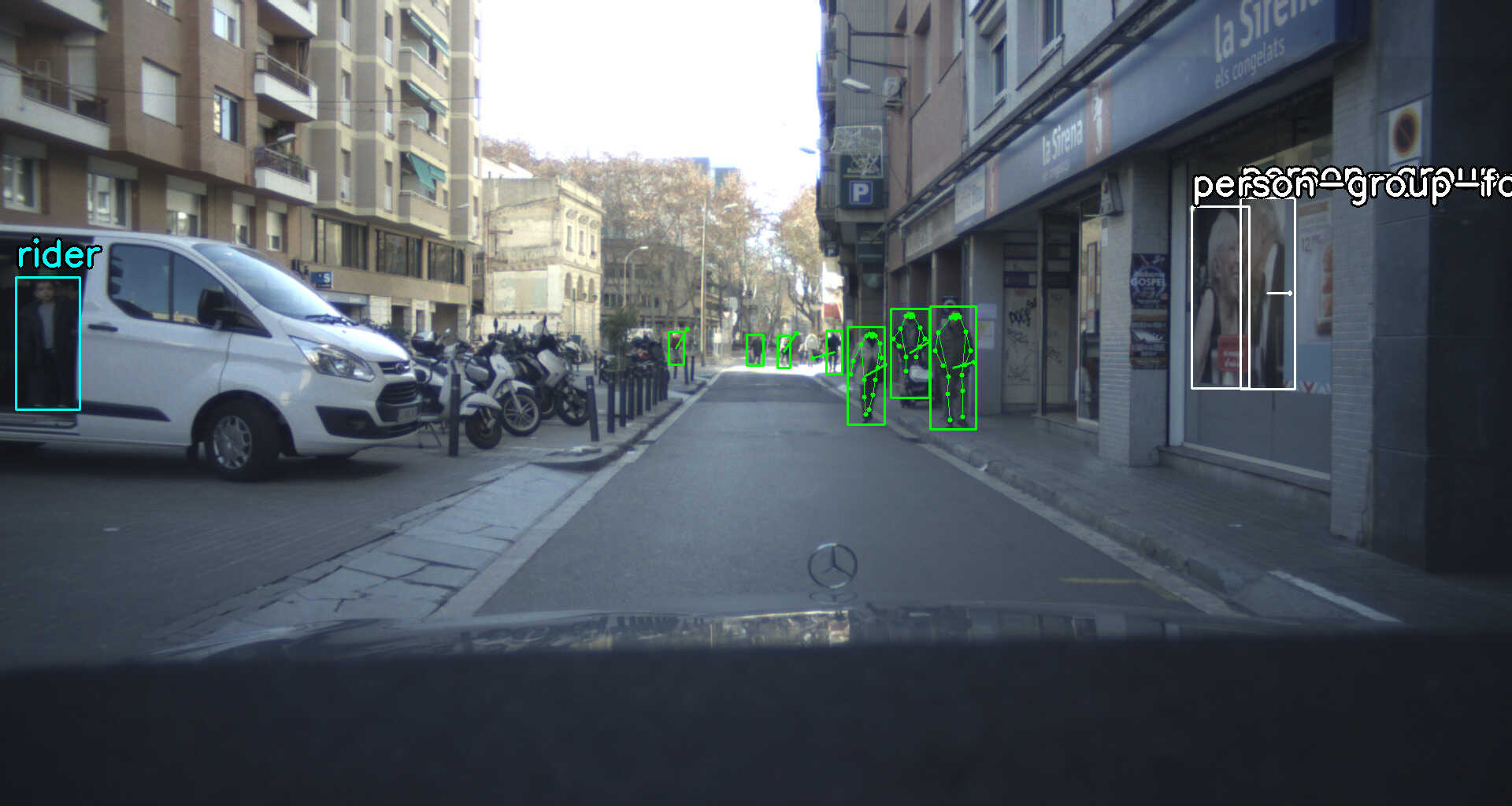} &
        \includegraphics[width=0.45\columnwidth, trim=0 4cm 0 2cm, clip]{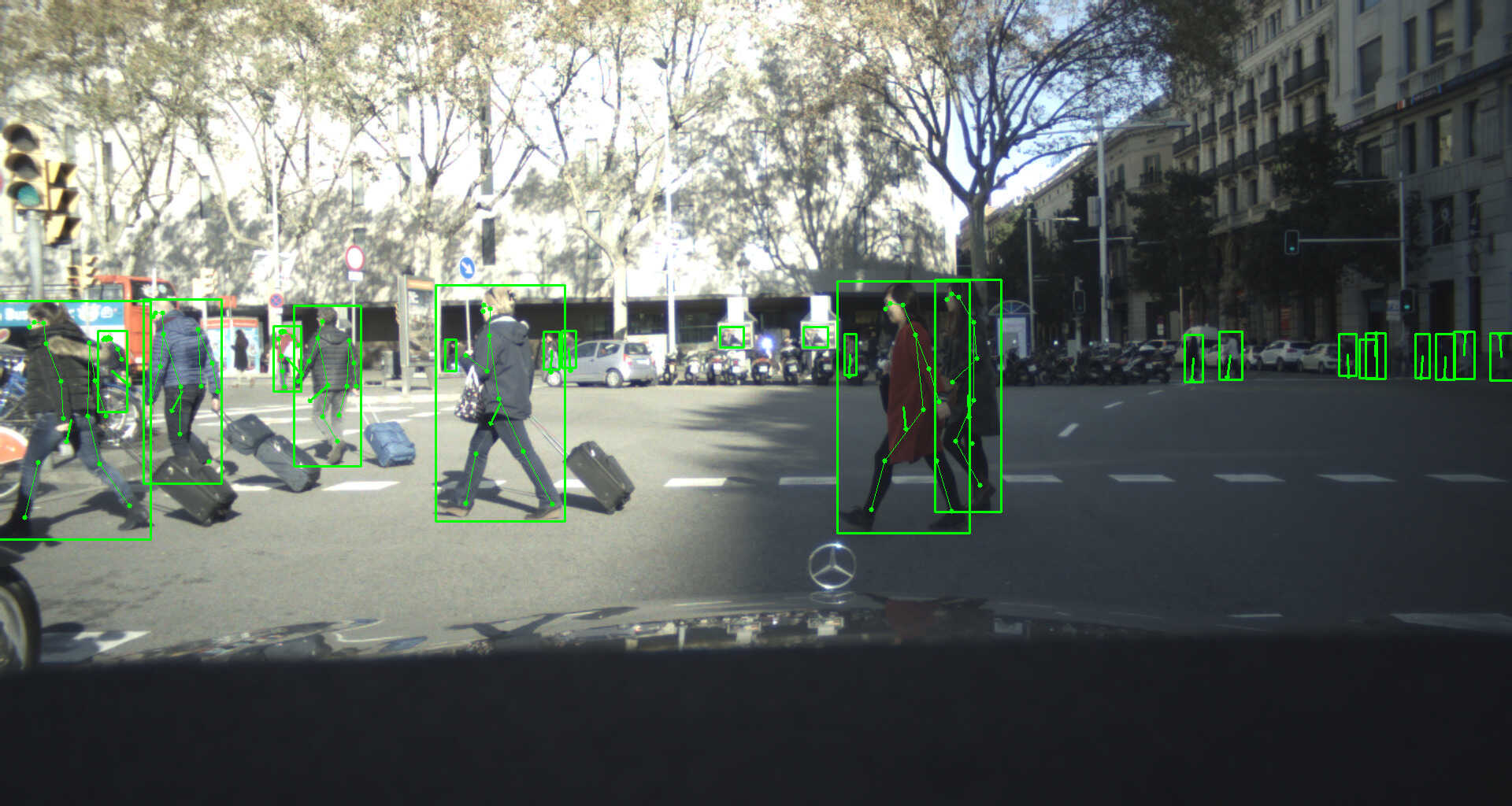} \\
        
    \end{tabular}
    
    \caption{Sample frames from ECPDP \cite{ECPDP}. The overlaid labels include bounding boxes for all instances, joint coordinates for pedestrians and riders, and orientations for eligible persons. For better illustration, category and occlusion labels for `pedestrian' (green) and `motorcycle' (red) are not displayed.}
    \label{fig:ECPDP_visualization}
\end{figure}
\paragraph{\textbf{ECPDP:}}
\label{dataset:ECPDP}
ECPDP \cite{ECPDP} extends ECP \cite{EuroCity} with a focus on crowded scenes with pedestrians overlapping. Both datasets were collected from a moving vehicle in 31 cities in 12 European countries using three synchronized cameras mounted on the front and both side mirrors. Two side cameras feature an expanded horizontal field of view of 85°. ECPDP contains 30,704 front-facing images (14,438 from ECP \cite{EuroCity}), 8263 left-view and 8008 right-view images, split into 29,570 training, 5,150 validation, and 12,255 testing images. The \href{https://eurocity-dataset.tudelft.nl/eval/extensions/ecpdp}{released images} have a resolution of 1920×1024 rather than the reported 1920×1080 \cite{ECPDP}. ECPDP defines four human-related classes, including pedestrian, rider, human, and person-group-far-away, together with 11 vehicle classes and one combined class, \textit{rider+vehicle-group-far-away}. Pedestrians and riders taller than 20 pixels are annotated with tight bounding boxes, including estimated boundaries for partially visible persons, while persons taller than 60 pixels are additionally annotated with 17 body joints. Each person receives an occlusion degree tag, and each joint is labeled with binary visibility. Annotations for testing images are kept unreleased. Fig.~\ref{fig:ECPDP_visualization} illustrates sample frames with annotations, but also reveals some issues. First, as shown in the top row, the occlusion labels for the same instance in consecutive frames are inconsistent. Second, the difference between `pedestrian' and `human' is not clarified. For example, the person in the distance who is separated by the vehicle in the top-right image is labeled as `human', while the persons on the sidewalk or motorcycles across the street in the bottom-right image are classified as `pedestrian'. Third, some classifications appear incorrect. For instance, in the bottom-left image, the passenger in the van is classified as `rider' and the poster characters are classified as `person-group-far-away'.

\subsubsection{Scene Classification}
\label{task:scene_classification}
The complexity of the driving environment is determined not only by the number of entities, but also by their spatial configurations, relative positions, motion patterns, and potential interaction \cite{CarlaSyn}. Consequently, traffic scene classification can be formulated as a hierarchical perception task that categorizes driving environments based on the presence of entities exhibiting prominent spatial relationships with the ego vehicle. The lane-change scenario is the most studied case in existing work \cite{CarlaSyn,sub-NUDrive}.
Related datasets can be broadly grouped into three types. The first type represented by \textit{CARLA-syn} \cite{CarlaSyn} synthesizes driving environments by simulating road user behaviors, the status of static entities, and contextual attributes. The second type represented by a subset of \textit{NUDrive} \cite{sub-NUDrive} collects non-accident real-world traffic scenarios, which, while not resulting in accidents, feature significant spatial relations between road users. The third type involves risky lane changes in real-life traffic accident scenarios, such as 620-dash and 571-honda \cite{CarlaSyn}. However, since the last two datasets have now become unavailable, we exclude the last type in this section.

\input{figures_figures/28_scene_classification_CarlaSyn}
\paragraph{\textbf{CARLA-syn:}}
\label{dataset:carla-syn}
Two synthetic lane-changing video datasets, 271-syn and 1043-syn \cite{CarlaSyn}, are generated using CARLA \cite{CARLA} at a resolution of 1280×720. Spatial relations were the main consideration when designing scenarios, characterized by relative distance, directional orientations, and is-in relation to the lane \cite{CarlaSyn-risk}. Based on annotator scores, each clip is assigned two binary labels: one for subjectively risky lane-change scenarios and the other for collision scenarios. However, the scoring criteria are implicit and may vary across annotators. The difference between a `risky scenario' and a `collision scenario' is not clarified. Although identifying a collision scenario should be a straightforward binary decision in which all annotators would unanimously reach the same conclusion when viewing the same traffic video, the method of label determination by rounding the average score \cite{CarlaSyn-risk} reveals the underlying disagreements of the annotators in some scenarios, highlighting the potential ambiguity in certain traffic scenarios. Although the simulated vehicle behaviors are configurable, including collision-avoidance probability and compliance with traffic lights, both datasets exhibit a strong class imbalance toward negative samples. In 271-syn, the non-collision-to-collision ratio is 6.12:1 and the safe-to-risky lane-change ratio is 2.3:1, while in 1043-syn, the corresponding values are 7.91:1 and unspecified. In this context, 306-syn is created by subsampling 1043-syn to achieve balanced distributions across both categories \cite{CarlaSyn, CarlaSyn-risk}.
Only 50699 frames extracted from 1043-syn are \href{https://ieee-dataport.org/documents/scenegraph-risk-assessment-dataset}{publicly available}. Fig.~\ref{fig:1043_visualization} presents sample frames in 1043-syn, which are claimed (A) to be risky and (B) safe for lane change, respectively \cite{CarlaSyn-risk}. However, the labels provided are neither binary for classification nor raw scores within the reported range ([-2, 2]) \cite{CarlaSyn-risk}. Despite the importance of spatial positioning, the annotation provides only a single coordinate per object (two dimensions with a zero for the third) and its lane association, preventing accurate localization of the complete object. Furthermore, similar to other synthetic datasets, the simulated scenarios are much less complex than the real-world environments. 

\begin{figure}[!t]

    \begin{tabular}{c@{\hspace{0.01\columnwidth}}c}
        \includegraphics[width=0.44\columnwidth, height=0.2\columnwidth]{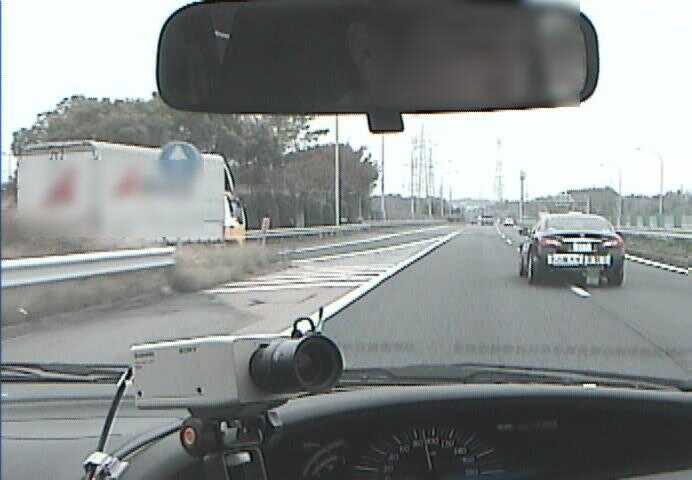} &
        \includegraphics[width=0.44\columnwidth, height=0.2\columnwidth]{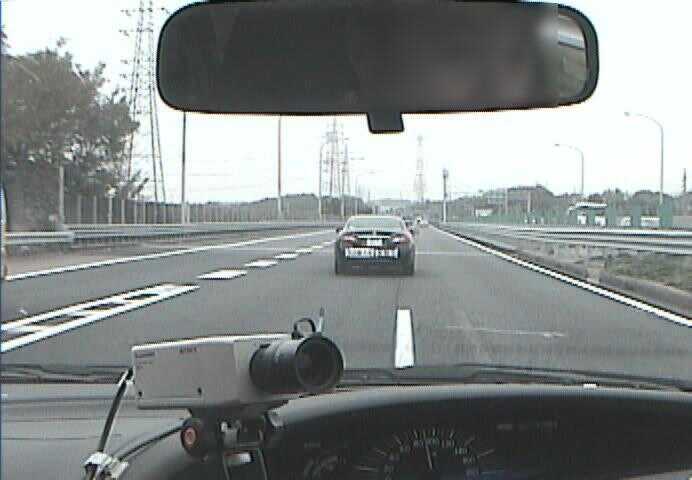} \\
        
    \end{tabular}
    
    \caption{Sample frames in sub-NUDrive \cite{CarlaSyn} reveal that multiple lane-change maneuvers contained in the single clip.}
    \label{fig:sub-NUDrive_visualization}
\end{figure}
\paragraph{\textbf{sub-NUDrive:}}
\label{dataset:sub-NUDrive}
A subset of NUDrive \cite{sub-NUDrive} includes 860 video clips of lane change captured in Nagoya, Japan, with a front-facing camera at 692 × 480. After normalizing the risk scores (ranging from one to five) given by ten annotators, the most risky 5\% lane change clips are considered risky, while others are annotated as safe. As a result, there is an unbalanced distribution of 43 risky clips to 817 safe clips. Despite claims of public availability for research purposes, \href{https://github.com/Ekim-Yurtsever/DeepTL-Lane-Change-Classification}{Github repository} provides only raw frames from two sample video clips, without accompanying ground-truth labels. According to the investigation of released frames, as shown in Fig.~\ref{fig:sub-NUDrive_visualization}, while a single clip may contain multiple lane change actions, a single binary label is assigned to the entire clip.

\subsection{Pertinency due to Semantic Category}\label{subsec:pertinency_semantic}
Different types of scene entities require different treatment in perception and decision-making, making semantic categorization of key entities essential. For example, pedestrians and cyclists, considered among the most critical components of traffic scenarios \cite{CALTECH}, represent vulnerable road users that require increased safety considerations. Traffic devices must be accurately identified and attended, as they transmit regulatory information that directly governs the behaviors of road users. Road boundaries and lane markings require precise detection to ensure proper positioning and navigation of vehicles. In this context, specialized benchmarks have been developed to identify scene entities in specific categories and analyze their significance in traffic scenes.

\subsubsection{Pedestrian Detection}
\label{task:pedestrian_detection}
Pedestrian detection task is a subfield of object detection, which involves localizing and classifying pedestrians in a traffic scene via bounding boxes.
Although the Cambridge Dictionary \cite{pedestrian} defines a \textit{pedestrian} as a person who is walking, especially in an area where vehicles go, some pedestrian detection benchmarks have expanded this definition to include broader human factors within driving environments, such as riders and stationary individuals \cite{pedx, CityPersons}. This expanded scope better reflects the diverse range of human factors that ADAS must recognize and respond to in real-world scenarios.

Several benchmark datasets have emerged to develop and evaluate pedestrian perception algorithms. \textit{Caltech} \cite{CALTECH} is a pioneering large-scale dataset for pedestrian detection in urban driving videos. \textit{NightOwls} \cite{NightOwls} is designed for nighttime pedestrian detection, targeting the challenges of low resolution and inhomogeneous illumination. LLVIP \cite{LLVIP} also focuses on nighttime scenarios by providing \textit{\underline{L}ow-\underline{L}ight \underline{V}isible-\underline{I}nfrared \underline{P}aired} surveillance images from elevated surveillance viewpoints. \textit{CityPersons} adopts finely annotated urban scenes of Cityscapes \cite{Cityscapes}, covering 27 cities in Germany, France, and Switzerland. \textit{\underline{E}uro\underline{C}ity \underline{P}ersons} (ECP) \cite{EuroCity} incorporates day and night scenarios from 31 cities in 12 of European countries. 

\input{figures_figures/30_pedestrian_detection_Caltech}
\paragraph{\textbf{Caltech:}}
\label{dataset:Caltech}
\href{https://data.caltech.edu/records/f6rph-90m20}{Caltech} \cite{CALTECH} collects 137 ego-centric videos in regular traffic in three urban cities in the Greater Los Angeles metropolitan area, divided into 11 sessions, with the first six used for training and the remainder for testing. Since \cite{CALTECH} only reports approximate statistics, we analyzed annotation files, which reveal that Caltech contains 249,884 frames at a resolution of 640×480 and 346,621 bounding boxes in 132,082 frames. As shown in Fig.~\ref{fig:caltech_visualization}, each bounding box is associated with a class label, classifying \textit{pedestrian} into `person', `people', `person?' and `person-fa', and a unique instance ID for object tracking. Statistical analysis reveals significant variations in object distributions across sessions. For example, `person' appear in 91.72\% of frames in the second session, but in only 28.58\% of frames in the third session. Meanwhile, several annotation issues in Caltech may affect its practical utility. First, as shown in Fig.~\ref{fig:caltech_visualization}, Caltech adopts a broader definition of \textit{pedestrian} that includes instances such as a `person' walking in an enclosed sky corridor (second-row images) and a sitting `person' (top-right image), which weakens alignment with the conventional traffic-oriented notion of pedestrians and may limit suitability for road-scene pedestrian detection. Second, the criterion used to differentiate between `person' and `people' is not always convincing. For instance, in the top-left scenario, some visually distinguishable individuals are grouped as `people', while others in similar contexts are labeled as `person'. Classifying the standing individual in the top-right scenario as `people' is also questionable. Moreover, the boundary between `person' and `person-fa' is unclear. For example, in the second row, the individual walking on the skywalk is classified as `person', but the closer pedestrian waiting on the sidewalk is classified as `person-fa'. These ambiguities may introduce noise into model training and weaken the reliability of category-specific evaluation. Third, omitted instances, such as the person standing in the traffic flow in the left scenario, may bias training and benchmarking by treating valid targets as background. Fourth, labeling the almost completely occluded instance, such as the person (id:0) in the bottom-left frame compared to the same instance in the right frame, may introduce unnecessary learning difficulty. Finally, the criterion for assigning the ambiguous label `person?' is also unclear. For instance, in the top-left image, the instance occluded by a pillar, which is even hard for humans to identify, is labeled as `person', while the standing person and sitting person in the third row, which are more distinguishable, are also classified as `person?'. Such inconsistent treatment of ambiguity may reduce annotation credibility and complicate the interpretation of model errors.

\paragraph{\textbf{NightOwls:}}
\label{dataset:nightowls}
Proposed in 2018,
\href{https://www.nightowls-dataset.org/}{NightOwls} \cite{NightOwls} collects 279,000 nighttime images using a vehicle-mounted camera in seven cities in Germany, the Netherlands, and the United Kingdom. These 1024×640 images are divided into training, validation, and test sets with similar distributions of pedestrian poses and heights. As shown in Fig.~\ref{fig:NightOwls_cityperson_visualization}, bounding boxes are manually created for each pedestrian, cyclist, and motorcyclist with a height exceeding 50 pixels, while printed images of people, statues, and dense crowds where individual delineation is infeasible are classified as `\textit{ignore}'. Beyond standard labels such as category and instance ids, each bounding box is associated with attributes, including occlusion status, pose configuration, difficulty level, area, and tracking ID.

\input{figures_figures/31_pedestrian_detection_NightOwls_CityPersons}
\paragraph{\textbf{CityPersons:}}
\label{dataset:CityPersons}
CityPersons \cite{CityPersons} augments Cityscapes \cite{Cityscapes} with person bounding box annotations to ensure compatibility with established pedestrian detection protocols. As shown in Fig.~\ref{fig:NightOwls_cityperson_visualization}, solid and dashed boxes denote full-body and visible-body extents, respectively. According to \cite{CityPersons}, CityPersons classifies humans into four distinct groups, including \textit{pedestrians} who are walking, running, or standing individuals, \textit{riders} who are operating bicycles or motorcycles, \textit{sitting persons}, and \textit{other persons} with unusual postures (e.g. stretching). Those fake humans, such as people on posters and reflections in the windows, are classified as `\textit{ignore}'. However, as revealed by the yellow text in Fig.~\ref{fig:NightOwls_cityperson_visualization}, annotation files contain a sixth class, `\textit{person group}', labeling the crowds where individuals cannot be distinguished. All other scene components are grouped into an out-of-interest class. In addition, while annotations are created for all 5000 images, \href{https://github.com/CharlesShang/Detectron-PYTORCH/tree/master/data/citypersons/annotations}{public access} is limited to those for 3,475 images in training and validation sets.

\input{figures_figures/32_pedestrian_detection_ECP}
\paragraph{\textbf{ECP:}}
\label{dataset:ECP}
ECP \cite{EuroCity} encompasses 47,335 images with 1920×1024 pixels captured in 31 cities in 12 European countries, featuring more than 238,000 manually labeled person instances. Each instance is annotated using a bounding box and a body orientation label, while partially occluded instances are assigned an estimated full extent together with an occlusion level. As shown in Fig.~\ref{fig:ecp_visualization} by overlaying bounding boxes for instances, analysis of annotation files reveals 17 distinct categories, including pedestrians (red), riders (green), co-riders, six rider-vehicle types (white), six vehicle-group categories (pink), `person-group-far-away' (yellow) and `rider+vehicle-group-far-away' (cyan). Fig.~\ref{fig:ecp_visualization} also visualizes body orientation labels for pedestrians and transport with riders. ECP considers other scene areas out of interest. As revealed in Fig.~\ref{fig:ecp_visualization}, various fake person instances are misclassified as `person-group-far-away', such as reflections on car surfaces (top-left), passengers visible through bus windows (middle-left), blurred regions, and human figures on posters (middle-right). Such false-positive annotations may introduce false supervisory signals during training and thereby compromise evaluation reliability by rewarding detections of irrelevant or nonexistent targets. In the bottom-left scenario, some individuals walking on the zebra crossing are labeled as `person-group-far-away', despite being closer to and more relevant to the ego vehicle than distant pedestrians on the sidewalk who are labeled as `pedestrian'. This inconsistency blurs the boundary between categories and may compromise category-specific learning and fair benchmarking. Furthermore, ECP also contains class-label errors, such as a `motorcycle' annotated as a `scooter' (top-left image) and a scooter `rider' annotated as a `pedestrian' with the scooter omitted (top-right image).
In addition, as shown in the bottom-right image, some images are substantially occluded by the ego-vehicle windshield wiper, which may introduce additional ambiguity due to the similar visual appearance as jaywalking pedestrians or collision victims in accidents involving the ego vehicle. Similarly to \hyperref[dataset:PedX]{PedX} \cite{pedx}, ECP also has an extension, \hyperref[dataset:ECPDP]{ECPDP} \cite{ECPDP}, with detailed pose annotations for 17 joint points for each instance. In particular, \href{https://eurocity-dataset.tudelft.nl/eval/user/login?_next=/eval/downloads/detection#}{access to ECP} is restricted to students and staff from academic and non-profit research institutions, subject to application.

\input{figures_figures/33_pedestrian_detection_LLVIP}
\paragraph{\textbf{LLVIP:}}
\label{dataset:llvip}
Since being introduced in 2021, LLVIP \cite{LLVIP} has been a standout for several special values. First, LLVIP employs a binocular camera system to capture visible-infrared images from elevated surveillance viewpoints, enabling broader scene coverage and reduced occlusion compared to ego-centric dashboard cameras. The \href{https://github.com/bupt-ai-cz/LLVIP/blob/main/download_dataset.md}{publicly available} release contains 15,488 image pairs at 1280×1024 pixels. Second, LLVIP records pedestrian behavior in dark environments during evening hours (6-10 PM) \cite{LLVIP}, making it valuable for developing and evaluating pedestrian detection algorithms in challenging low-light conditions. The paired visible-infrared cameras allow LLVIP to annotate pedestrians more precisely on infrared images and then transfer the annotations to the aligned visible images. However, restricted by the distance and angle of shooting, pedestrians in LLVIP are medium sized \cite{LLVIP}, making it less suitable for the detection of small distant targets or large nearby targets. Changes in lighting conditions and viewpoint pose significant challenges to existing algorithms, as revealed by experimental results \cite{LLVIP}. Meanwhile, as illustrated in Fig.~\ref{fig:LLVIP_visualization}, LLVIP regards all persons (including riders, passengers, and people squatting on sidewalks) as pedestrians and classifies all instances into the same class `person'. There are also some annotation issues, such as labeling the rider and passenger as a single instance (middle-right image). The categorization and annotation issues might partially explain the performance gap observed in the experimental results and suggest directions for future improvement. 

\subsubsection{Cyclist Detection}
\label{task:Cyclist_detection}
Cyclists, as another category of vulnerable road users, also warrant dedicated efforts to be protected from traffic accidents \cite{cyclistdetection}. Compared with the boarder scope of \textit{pedestrian detection}, \textit{cyclist detection} focuses specifically on the classification and localization of cyclists in traffic scenarios. Although many benchmark datasets, such as Cityscapes \cite{Cityscapes}, BDD100K \cite{BDD100K}, and the aforementioned pedestrian datasets (including Citypersons \cite{CityPersons} and ECP \cite{EuroCity}), offer labels for cyclists, cyclists often represent a relatively small portion of their total labeled instances. And some datasets such as LLVIP \cite{LLVIP} even fail to establish cyclists as a distinct semantic category, instead subsuming them within the pedestrian class. In contrast, the \textit{\underline{T}singhua-\underline{D}aimler \underline{C}yclist} dataset (TDC) \cite{cyclistdetection} is specialized in the detection of cyclists in various urban environments by providing dedicated annotations.

\input{figures_figures/34_cyclist_detection_TDC}
\paragraph{\textbf{TDC:}}
\label{dataset:tdc}
Proposed in 2016, \href{http://www.gavrila.net/Datasets/Daimler_Pedestrian_Benchmark_D/Tsinghua-Daimler_Cyclist_Detec/tsinghua-daimler_cyclist_detec.html}{TDC} \cite{cyclistdetection} serves as a benchmark for vision-based cyclist detection, specifically designed to capture the diverse variations in cyclist appearances, poses, and scales. Its 14,674 images, captured in Beijing by vehicle-mounted stereo cameras at a resolution of 2048 × 1024, are divided for training (9741), validation (1019), testing (2914), and an additional subset (1000) without vulnerable road users. However, despite the presence of a dedicated negative subset \cite{cyclistdetection}, as shown in the second scenario in Fig.~\ref{fig:TDC_visualization} (b), these empty scenes also appear in the test set. Quantitative analysis of annotation files reveals that 16.13\% testing images (470 of 2914) do not contain pedestrians or riders. This may skew the overall evaluation metrics, as performance in these scenes may not reflect the actual capability of the models. Meanwhile, annotations in training set are restricted to cyclist instances larger than 60 pixels with less than 10\% occlusion or truncation and no motion blur \cite{cyclistdetection}. As shown in the top-middle image in Fig.~\ref{fig:TDC_visualization} (a), even stationary cyclists are excluded. By contrast, as illustrated in the first scenario in Fig.~\ref{fig:TDC_visualization}(b), the validation and test sets extend the annotations beyond cyclists to include pedestrians, tricyclists, and motorcyclists \cite{cyclistdetection}. However, some discrepancies exist in the annotations. For instance, in Fig.~\ref{fig:TDC_visualization} (a), instance `5100034' in the top-right image received an annotation despite being barely distinguishable from the background, whereas clearly visible riders were left unannotated, such as the one adjacent to the instance `5100041' in the bottom-left image and those moving horizontally in the bottom-middle image. Furthermore, as shown in the bottom-right image in Fig.~\ref{fig:TDC_visualization} (a), the rider and her passenger are annotated as a single cyclist instance, which may confuse the model during training and impair its ability to correctly learn the different features for cyclists. Moreover, the tracking ID `1' of the cyclist in this image does not comply with the annotation numbering format, which could potentially affect training if the model architecture or data pipeline relies on ID patterns for tracking, temporal analysis, or instance associations.

\subsubsection{2D Lane Detection}
\label{task:2d_lane_detection}
Lane detection, which involves classifying and localizing lane markings on roads \cite{openlane-v}, serves as a foundation for lane departure warnings and lane keeping or trajectory planning assistance. Unlike the generic object detection task (Sec.~\ref{subsec:objectdetection}) that surrounds targets with bounding boxes \cite{MaskRCNN}, lane detection uses curves to mark the location of lane markings \cite{CULane}. Although seemingly straightforward for human drivers, this task is challenging for algorithms due to adverse weather conditions, varying illumination \cite{CULane}, diverse road surface qualities, deterioration of lane markings over time, and occlusions from other vehicles \cite{openlane-v}.

Benchmark datasets provide ego-centric front-view footage and 2D lane line labels \cite{openlanev2}. \textit{TuSimple} \cite{TuSimpleGit} is a pioneer in highway scenarios and annotates up to five lane markings per frame using cubic splines. \textit{CULane} \cite{CULane} is a large-scale dataset for urban environments and annotates four lane markings between barriers. \textit{BDD100K} \cite{BDD100K} expands the scope to three types of attributes and introduces single-line and double-line annotation methods. Although \textit{VIL-100} \cite{VIL100} introduces frame-by-frame annotations to address temporal ambiguity, its effectiveness is limited by quality issues of the annotation. \textit{OpenLane-V} \cite{openlane-v} aims to achieve temporal consistency through matrix completion techniques while utilizing footage from the OpenLane dataset \cite{openlane}.

\input{figures_figures/35_2D_lane_detection_TuSimple}
\paragraph{\textbf{TuSimple:}}
\label{dataset:tusimple}
Proposed in 2017, \href{https://github.com/TuSimple/tusimple-benchmark/tree/master/doc/lane_detection}{TuSimple} \cite{TuSimpleGit} contains 6408 video clips on US highways, with 3626 for training and 2782 for testing. Each clip is converted into 20 frames at 1280×720, but only every last frame is annotated. TuSimple \cite{TuSimpleGit} releases lane labels in JSON format together with grayscale PNG segmentation maps that distinguish lane markings by pixel intensity. Figure \ref{fig:tusimple_visualization} visualizes the two annotation formats with RGB colors for a better illustration. Instead of manually drawing lane curves, TuSimple represents each lane using points sampled at predefined vertical positions relative to the recording vehicle (h\_samples), with each point specified by its horizontal coordinate and corresponding h\_sample value \cite{TuSimpleGit}.
Although it may ease the evaluation, as shown in Fig.~\ref{fig:tusimple_visualization}, the annotated lane curves are not smooth and are even not aligned with the actual lane markings, causing models to learn inaccurate lane geometry. Furthermore, there are several inconsistencies in the annotation scheme. First, the length of the annotated lane markings varies. Some annotations extend to the far end, such as the green lane in the first image in Fig.~\ref{fig:tusimple_visualization} (a), while others terminate prematurely, such as the green lane in the second image in Fig.~\ref{fig:tusimple_visualization} (a), even if they are created for the same lane marking in different frames. Obstructed lane markings lead to the second inconsistency. In the third image in Fig.~\ref{fig:tusimple_visualization} (a), while the blue lane annotation terminates at the point of vehicle occlusion, the red lane annotation continues through the occluded area. Third, when lane markings diverge, inconsistent strategies are used in different images to determine which one should be annotated. For example, in the last three images in Fig.~\ref{fig:tusimple_visualization} (a), the cyan lane depicts a different segment of the same lane marking in the three consecutive frames. Additionally, despite the claimed maximum annotation of five lane markings per frame \cite{TuSimpleGit}, the choices of lane markings to annotate in adjacent images are inconsistent. For example, the left image in Fig.~\ref{fig:tusimple_visualization} (b) annotates four lane markings starting from the second left one, while the middle image has five lane markings annotated starting from the leftest, and the annotation in the right image includes five lane markings starting from the second left to the rightest. Moreover, while the scenarios in Fig.~\ref{fig:tusimple_visualization} (a) contain more than five visible lane markings, only four are annotated in each case, indicating an inconsistency with the annotation protocol observed in Fig.~\ref{fig:tusimple_visualization} (b). These inconsistencies and mismatches can confuse models about the stopping criteria and lane selection standards, introducing additional challenges in detecting multiple, long-range, occluded and diverging lanes.

\input{figures_figures/36_2D_lane_detection_CULane}
\paragraph{\textbf{CULane:}}
\label{dataset:CULane}
\href{https://xingangpan.github.io/projects/CULane.html}{CULane} \cite{CULane}, introduced in 2018, contains 133,235 frames at a resolution of 1640×590, including 88,880 for training, 9675 for validation, and 34,680 for test. The footage is collected by six vehicles equipped with cameras operated by different drivers in Beijing under different environmental conditions. Annotated lines are created exclusively for four traffic lane markings within the road barriers in each frame. However, what constitutes four lane markings is not explicitly defined in \cite{CULane}, for example, lane markings in four specific categories or four lane markings dividing the current lane and adjacent lanes. Despite the claim that four lanes are classified into distinct categories \cite{CULane}, the annotations actually treat each lane marking as an instance within a single category. To enable a distinguishable illustration, in Fig.~\ref{fig:culane_visualization}, we use different colors to visualize different annotated lane markings in the format of cubic splines (top) and semi-transparent masks (bottom). Fig.~\ref{fig:culane_visualization} shows that annotation is also created for occluded, invisible, and even unmarked lane markings to maintain continuity. However, analysis of annotation files also uncovers several issues. First, the ground truth for the same lane marking can vary between consecutive frames, even when the ego vehicle remains stationary, such as the second left lane marking in the second and third scenarios in Fig.~\ref{fig:culane_visualization}. This frame-to-frame variation can confuse models about the actual lane geometry and reduce temporal stability. Second, there are also some incorrect annotations. For example, in the right scenario, a single standard-width lane is erroneously divided by three separate lines, while the directional arrow marking in the center clearly indicates that it is a single lane, which may introduce false supervisory signals and lead models to learn erroneous lane structures. Furthermore, as shown in Fig.~\ref{fig:culane_visualization}, the annotations of the lane markings in some scenarios may not extend to their full length at the far end of the road. This truncation of the lane marking annotations may prevent models from learning complete lane representations, thereby reducing their ability to detect long-range lane structure accurately.

\input{figures_figures/37_2D_lane_detection_BDD100K}
\paragraph{\textbf{BDD100K:}}
\label{dataset:BDD100K}
Released in 2020, BDD100K \cite{BDD100K} is one of the largest driving video datasets. It contains 100,000 videos captured at 1280×720 pixels using vehicle-mounted phone cameras \cite{BDDGit}, together with 100,000 images extracted from the tenth second frame of each video for image-based tasks. The dataset uses a 7:1:2 split for training, validation, and testing. Due to \href{https://github.com/bdd100k/bdd100k/tree/master}{temporary inaccessibility} of the complete version, our analysis and visualization are performed based on the sample package provided by \cite{datasetninjabdd100k}. As illustrated in Fig.~\ref{fig:bdd_visualization}, BDD100K distinguishes itself from other datasets by using both single-line and double-line annotations to represent lane markings. Although the selection between single and double lines seems to correlate with the physical width of the lane markings, it is not explicitly documented. Furthermore, in addition to coordinates, each lane marking is annotated with attributes describing lane direction (parallel or vertical), lane style (full or dashed) and lane type. Although the sample package \cite{datasetninjabdd100k} contains only six types, \cite{BDD100K} reports annotations for single and double lane markings in white, yellow, and other colors, as well as crosswalk and road curb, resulting in 32 categories in total. However, inconsistencies and errors appear in the annotation schema. Fig.~\ref{fig:bdd_visualization} presents some representative cases, where we overlay annotations extracted from JSON files on the corresponding raw images and assign colors to distinguish lane markings with different attributes. 
First, some annotations deviate from the true lane boundaries (top-left image) or are erroneously applied to regions without visible markings (left red line in the top-right image and middle yellow lane in the middle-right image).
Second, annotations can incorrectly connect disparate lane markings, such as the rightmost red lanes after occluded areas in the top-right image. Misclassification is also observed. For example, the green crosswalk labeled as a parallel solid road curb in the middle-left image should be a dashed vertical crosswalk (to the vehicle path). Similarly, in the middle right image, the vertical pink line is incorrectly labeled as parallel, and the red and green lines actually depict both vertical and parallel lane markings. Moreover, BDD100K also suffers from missing annotations, evident by the unannotated single solid white lane marking adjacent to the subject vehicle in the bottom left image despite clear visibility. In addition, BDD100K also contains some scenarios without visible lane marking, such as the bottom right image. These issues can introduce supervisory noise, obscure lane geometry and category distinctions, and bias training through incomplete annotations. Additionally, they also undermine the reliability and fairness of benchmark evaluation.

\input{figures_figures/38_2D_lane_detection_VIL100}
\paragraph{\textbf{VIL-100:}}
\label{dataset:VIL-100}
As the first video-based lane dataset \cite{openlane-v} introduced in 2021, \href{https://github.com/yujun0-0/mma-net}{VIL-100} \cite{VIL100} aims to extend lane detection from single images to video sequences and use temporal information to resolve in-frame ambiguities such as occlusions. It comprises 100 videos, including 97 self-collected videos and 3 sourced from a website, from which 10,000 images were extracted at eight different resolutions ranging from 1920 × 1080 to 640 × 368. However, all images are recorded at 1920×1080 pixels in annotation files. Analysis of annotation files reveals that 80.2\% of images (8020 out of 10,000) depict highway scenes, while 19.5\% (1948 images) record city scenarios. The remaining 0.3\% lack a scene-type label. Most images contain (43. 7\%) or 5 (33.8\%) annotated lane markings, followed by 3 (15.5\%) and 2 (5.0\%), while only 20 images contain the maximum of 6 lane markings.
Fig.~\ref{fig:vil100_visualization} visualizes sample images, where we overlay annotations derived from the JSON file and assign colors based on the lane ID. Segmentation maps with black background are also presented. However, it clearly demonstrates the annotation errors in both amount and positioning, introducing supervisory noise and obscuring lane geometry. Meanwhile, there are also 175 images of 6 videos without annotations for lane markings, despite containing clearly visible lane markings, as shown in the middle column of Fig.~\ref{fig:vil100_visualization} (a), and despite the annotations of the same lane markings given in the adjacent frames, as shown in the other columns of Fig.~\ref{fig:vil100_visualization} (a). These false negatives induce training bias by systematically underrepresenting unlabeled lane markings.
Furthermore, despite the ambition of addressing occlusion ambiguities by providing temporal information, the criteria for whether to annotate a lane marking are ambiguous, leading to inconsistent annotations in even adjacent frames, as shown in Fig.~\ref{fig:vil100_visualization} (b). Such annotation inconsistencies and errors pose significant challenges in training robust lane detection models, as they may learn conflicting and incorrect patterns.

\input{figures_figures/39_2D_lane_detection_OpenlaneV}
\paragraph{\textbf{OpenLane-V:}}
\label{dataset:OpenLane-V}
Proposed in 2023, \href{https://github.com/dongkwonjin/RVLD}{OpenLane-V} \cite{openlane-v} mitigates the temporal inconsistencies in the original OpenLane \cite{openlane} annotations by implementing a semi-automatic annotation process. This enhancement reconstructs missing lane segments using matrix completion techniques and creates manual annotations for suddenly invisible lane markings to mitigate flickering artifacts. OpenLane-V consists of 90,000 images from 590 videos, with 70,000 for training and 20,000 for testing, excluding footage that contains a significant proportion of imperceptible lane segments. Despite the 2D and 3D labels contained in OpenLane \cite{openlane}, OpenLane-V \cite{openlane-v} only provides 2D coordinates. As revealed in Fig.~\ref{fig:openlane-v_visualization} (left), up to four lane markings of the ego and alternative lanes are annotated per image. Although this design emphasizes the most important lanes from the ego perspective, the resulting ego-centric bias may limit the learning of full-scene lane topology. As evidenced by Fig.~\ref{fig:openlane-v_visualization}, the stopping line and distant segments of the four lane markings are not within the scope of the OpenLane-V annotation schema. This restricted annotation scope may hinder the learning of long-range geometry and structurally important cues, while also potentially distorting benchmark evaluation.

\subsubsection{2D Lanemark Segmentation}
\label{task:lanemark}
Compared to \textit{2D lane detection} that localizes lane markings using single or double lines, \textit{2D lanemark segmentation} depicts lane markings through pixel-level masks \cite{ApolloScape}. Despite the pixel-level annotations contained in the aforementioned datasets \cite{TuSimpleGit, CULane, VIL100}, ApolloScape \cite{ApolloScape} extends its coverage to include additional markings such as directional arrows and other traffic indicators.

\begin{figure}[!t]
    \centering

    \begin{tabular}{c@{\hspace{0.01\columnwidth}}c}
         
        \includegraphics[width=0.45\columnwidth, height=0.15\columnwidth, trim=0 0 0 2.8cm, clip]{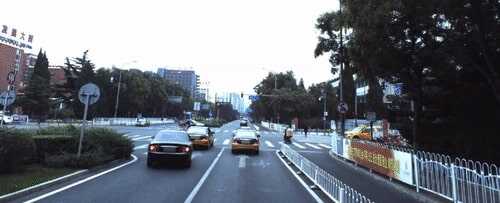} &
        \includegraphics[width=0.45\columnwidth, height=0.15\columnwidth, trim=0 0 0 2.8cm, clip]{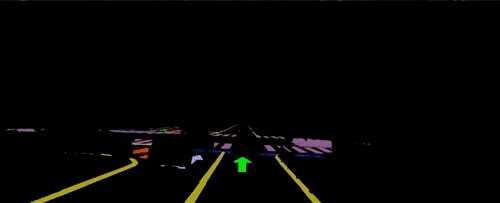} \\[-0.2em]

    \end{tabular}

    \caption{Sample frames and corresponding semantic annotations of ApolloScape \cite{ApolloScape} provided on the repository \cite{ApolloGit}.}
    \label{fig:apollo_visualization}
\end{figure}
\paragraph{\textbf{ApolloScape:}}
\label{dataset:ApolloScape}
Proposed in 2020, ApolloScape \cite{ApolloScape} is a large-scale multi-task dataset. Due to \href{https://apolloscape.auto/lane_segmentation.html}{temporary unavailability} \cite{ApolloGit}, the following analysis is based on \cite{ApolloScape} and its repository \cite{ApolloGit}. ApolloScape provides 165,949 images (3384×2710) collected from 3 Chinese sites for semantic segmentation of lanemarks, with 132,189 and 33,760 for training and testing. Although \cite{ApolloScape} defines 35 lane marking classes organized into 11 functional categories, the released package contains 27 classes \cite{ApolloGit}. Extending beyond the general lines, such as double solid yellow lines, ApolloScape \cite{ApolloScape} also covers directional arrows, speed reduction bumps, diamond and rectangle indicators, and other markings. Each class is jointly defined by the type (e.g., arrow or double solid), color, and functional attributes of the lane marks. Each pixel is categorized into one of the lanemark classes, the void class, or the ignored class. Fig.~\ref{fig:apollo_visualization} shows a sample scenario in which ApolloScape provides complete semantic segmentation by labeling the entire crosswalk area, compared to BDD100K \cite{BDD100K} which only supports the detection of crosswalks by marking the left and right boundaries.

\subsubsection{3D Lane Detection}
\label{task:3d_lane}
This task aims to generate a set of 3D curves in camera coordinates from a single front-view image to represent both lane delimiters and centerlines \cite{9008811}. It is able to overcome the inherent limitations of 2D lane detection, such as depth information loss and perspective distortion effects, due to the capability of capturing complete spatial geometry and preserving height information and elevation changes.
Several datasets have been created to support this enhanced geometric understanding. Compared with 2D datasets that contain only image-plane annotations, 3D datasets provide real-world spatial coordinates. A notable example is OpenLane \cite{openlane}.

\input{figures_figures/41_3d_lane_detection_OpenLane}
\paragraph{\textbf{OpenLane:}}
\label{dataset:openlane}
Released in 2022, \href{https://github.com/OpenDriveLab/OpenLane/blob/main/data/README.md}{OpenLane} \cite{openlane} is, to our knowledge, the first large-scale real-world 3D lane dataset, built upon the Waymo Open dataset \cite{Waymo}. It contains 200,000 frames at a resolution of 1920×1280, extracted from 1000 videos. All visible lane marking segments in each frame are annotated, including those of opposing traffic lanes where no barriers separate the directions of travel. The maximum number of annotated lane markings within a single frame reaches 24, and around 25\% frames have more than 6 annotations. Each lane marking is identified by its category among 14 predefined classes and by its relative position to the ego vehicle. However, the imbalanced distribution is notable, with approximately 90\% of the annotations belonging to three predominant classes: double yellow solid, single white solid, and single white dashed lanes \cite{openlane}. Although \cite{openlane} reported that each line is assigned a unique tracking ID throughout the segment, it is not contained in the released annotation files. OpenLane creates 2D lane annotations in the image plane and corresponding 3D coordinates in the world space. Fig.~\ref{fig:openlane_visualization} visualizes 2D annotations of lane markings that are marked as visible and assigns colors based on the classification label. Despite its scale and ambition, the annotation schema exhibits several issues. Inconsistency is most typical. First, the annotation of persistent physical lane markings in sequential frames might be intermittent, as evidenced by the left red line in the first scenario in Fig.~\ref{fig:openlane_visualization}, which is omitted in the left frame while annotated in the subsequent frame. Second, annotations for the same lane marking might be disparate in even consecutive frames. It includes spatial inconsistencies, such as varied annotation lengths (illustrated by the cyan line in the first scenario) and semantic inconsistencies, where identical lane marking is labeled as distinct classes (indicated by the different colors of the annotations for double yellow solid lines in the second scenario). These intermittent and inconsistent annotations introduce noisy and contradictory supervision, hindering the learning of temporal continuity and class discrimination. Meanwhile, annotation disparities can occur within single frames. An example is shown in the first scenario in Fig.~\ref{fig:openlane_visualization} where the single physical lane marking is redundantly annotated twice (visualized by red lines). Furthermore, as shown in Fig.~\ref{fig:openlane_visualization}, some occluded lane markings are labeled visible, indicating a defective visibility classification. In addition, the annotation schema also has some inherent limitations. First, only considering visible segments occasionally results in fragmented lane annotations, where a single continuous lane is divided into multiple discrete parts, and also leads to temporal inconsistencies \cite{openlane-v}. 
Moreover, the lane splicing process, while aimed at creating continuous lane representations, may introduce unintended artifacts. Finally, the smoothing operations implemented to reduce noise could inadvertently eliminate fine-grained but important structural details. 

\subsubsection{High-definition (HD) Map Construction}
\label{task:hd_map}
HD maps delineate the positions of road elements, especially road boundaries, lane dividers, and crosswalks \cite{comment2paper5}, providing semantically rich topological information for driving navigation \cite{comment2paper4}. In contrast to \hyperref[task:2d_lane_detection]{2D lane detection} and \hyperref[task:3d_lane]{3D lane detection}, HD maps capture not only the detailed local geometry of various road elements but also their global topological relationships, such as lane connectivity through intersections \cite{comment2paper6}. The traditional HD map was constructed offline using LiDAR-SLAM-based methods, which, however, incur high maintenance and annotation costs and are also vulnerable to localization errors \cite{comment2paper4}. More recent studies leverage multimodal data acquired from various onboard vehicle sensors to construct online HD maps with vectorized representations of the road environment \cite{comment2paper5}.
The aforementioned large-scale, multi-task datasets, nuScenes \cite{nuScenes} and Argoverse2 \cite{Argoverse2} provide multi-modal and geometry-aware data foundations and high-quality annotations for online HD map construction.

\input{figures_figures/53_hd_map_construction_nuScenes}

\paragraph{\textbf{nuScenes:}} 
\label{dataset:nuScenes}
Published in 2020, nuScenes \cite{nuScenes}
\href{https://www.nuscenes.org/nuscenes#download}{comprises 1000 20-second scenes} collected in Boston and Singapore to capture dense traffic and complex urban driving scenarios. A synchronized 360-degree sensor suite is employed for data collection, including six surround-view {\small RGB} cameras at 12 Hz, a 32-beam spinning {\small LiDAR} at 20 Hz, five radars at 13 Hz with a detection range of up to 250 meters, together with {\small GPS} and {\small IMU}. Among the 1.4 million camera images, 40,000 key frames are annotated. Compared with the original rasterized map, which contains semantic annotations only for roads and sidewalks, the vectorized semantic map expands the annotation scope to 11 classes of road elements. In addition, as a multi-task dataset, nuScenes also provides attribute-aware 3D bounding box annotations for 23 object classes, supporting 3D object detection and tracking. 
Despite the rich sensory data and comprehensive annotations, nuScenes still has several limitations that should be noted before use. First, although the dataset includes some sequences captured in adverse weather, challenging lighting conditions, and special locations, its primary focus remains urban driving scenarios under normal environmental conditions, and its geographic diversity is limited to two cities. Second, rather than adopting the commonly used 5-point threshold, nuScenes annotates objects whenever they are hit by at least one LiDAR or radar point. Since the LiDAR sensor is mounted substantially higher than the cameras, it may detect objects that are occluded from the camera viewpoint, thereby creating apparent false positives for camera-based perception such as (b) in Fig.~\ref{fig:nuscenes_examples}. Conversely, the 1-point threshold may also result in false negatives, as evidenced by the missing vehicle annotations in Fig.~\ref{fig:nuscenes_examples}(c). The same figure further shows that some objects covered by point clouds remain unannotated, introducing ambiguity for detection models. In addition, the annotated poses of traffic lights and ego vehicle are problematic in several scenes \cite{nuscenesgit}.

\begin{figure}[!t]
    \centering

    
    \includegraphics[width=1\columnwidth]{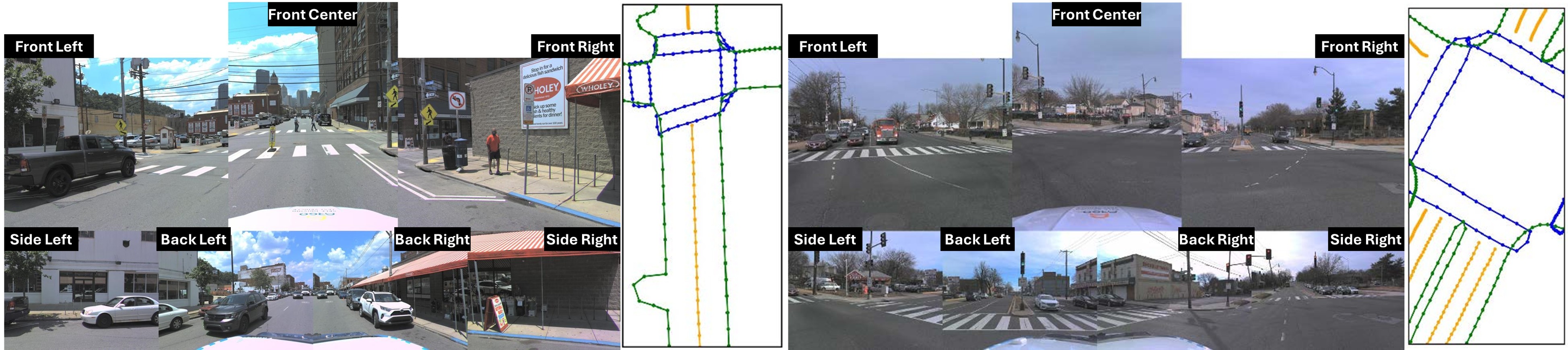}

    \caption{Surround-view images with corresponding HD map annotations (yellow: lane divider, green: road boundary, blue: crosswalk) of sample scenes in Argoverse2 \cite{Argoverse2}.}
    \label{fig:argoverse2_examples}
\end{figure}

\paragraph{\textbf{Argoverse2:}}
\label{dataset:Argoverse2}
Argoverse2 \cite{Argoverse2} is a large-scale multi-modal, multi-task benchmark collected across six U.S. cities. Its three main subsets, sensor dataset, LiDAR dataset, and motion forecasting dataset, are primarily designed for 3D object detection, point cloud forecasting, and motion forecasting, respectively \cite{Argoverse2}. In addition to these core tasks, Argoverse2 is also highly valuable for HD map construction as each scenario is paired with a structured local HD map \cite{Argoverse2}. As a result, it has been widely adopted for evaluating online map inference methods, with the sensor dataset being the most widely used subset in this context \cite{Mask2Map,arg_info}. This subset
\href{https://www.argoverse.org/av2.html#download-link}{comprises 1000 15-second scenes}, 
selected to capture diverse traffic situations, road geometries, and environmental conditions. As shown in Fig.~\ref{fig:argoverse2_examples}, each scenario is captured by seven ring cameras and two forward-facing stereo cameras at 20 fps, synchronized with dual 32-beam LiDAR sensors operating at 10 Hz \cite{Argoverse2}. The front-center camera has a resolution of 1550$\times$2048, while the remaining cameras share the resolution of 2048$\times$1550. Argoverse2 annotates lane segments and crosswalks with 3D polylines while uses polygons to depict driving areas. In particular, it provides a dense ground-surface height map in the form of a 2D raster array \cite{Argoverse2}, which offers useful geometric priors for road-surface understanding and map reconstruction. Since its scenarios are intentionally selected to emphasize challenging driving situations rather than unbiased city-scale coverage \cite{Argoverse2}, Argoverse2 is more suitable and widely adopted for benchmarking inference performance \cite{Mask2Map,arg_info}.

\subsubsection{Traffic Sign Detection}
\label{task:sign_detection}
Compared to the \textit{traffic sign recognition} task that classifies single-sign images, traffic sign detection localizes all traffic signs within an image and identifies their respective categories \cite{DFG}. It faces challenges stemming from both the inherent diversity of traffic signs (e.g. variations in shape, size, and color) and the practical deployment conditions (e.g. varying viewing angles and distances). It is crucial to compensate for human limitations in perceiving and interpreting roadway signage. Studies have revealed that drivers often overlook certain types of traffic signs, particularly pedestrian crossing signs \cite{LISA, 6335478}.

This task inherently requires large-scale datasets. However, regional variations create challenges for the development of broadly applicable datasets. Specifically, while some traffic signs follow international conventions, such as warning signs that typically feature yellow triangles with black borders \cite{Tencent2016}, traffic signs exhibit significant regional variation in languages, symbols, colors, and design standards across countries, reflecting strong geographical characteristics (e.g. the unique `Kangaroo in next 5km' sign in Australia). Consequently, existing traffic sign datasets are typically collected and organized on a country-by-country basis. This section discusses representative datasets, including \textit{DFG} \cite{DFG}, \textit{\underline{T}singhua \underline{T}encent \underline{100K}} (TT100K) \cite{Tencent2016}, \textit{\underline{G}erman \underline{T}raffic \underline{S}ign \underline{D}etection \underline{B}enchmark} (GTSDB) \cite{GTSDB2020}, and \textit{GLARE} \cite{Glare}.

\paragraph{\textbf{DFG:}}
\label{dataset:DFG}
\href{https://www.vicos.si/resources/dfg/}{DFG} \cite{DFG} is collected in rural and urban areas under favorable light and weather conditions to maintain a comprehensive inventory of Slovenian traffic signs.  
Analysis of annotation files reveals three resolutions of main images, including 6794 images (5128 training, 1666 testing) at 1920×1080 pixels, 162 images (125 training, 37 testing) at 720×576 pixels, and one training image at 864×691 pixels. Additional augmented images are also provided, leading to a package of 16,264 images in total. Although \cite{DFG} presents DFG as a traffic sign detection dataset with ground-truth bounding boxes, more than 75\% traffic sign instances (13239 of 17598) are also annotated with precise polygons, while the remaining loosely annotated instances are marked as ignore. This annotation protocol enables traffic sign segmentation. Furthermore, although traffic signs are classified into 200 categories, Fig.~\ref{fig:DFG_visualization} shows that the category naming scheme adopts the combination of Roman numerals and Arabic numerals (e.g. I-28), which lacks semantic clarity. Although the annotation schema defines a `\textit{supercategory}', all instances share the same value, `\textit{traffic\_sign}', which further obscures the actual functional distinctions among signs, such as speed limits or warning signs. Moreover, the last image in Fig.~\ref{fig:DFG_visualization} reveals missing annotations for certain images. Although these images belong to the main dataset and are listed under the `images' key in the annotation file, their corresponding entries are absent under the `annotations' key, such as training images with the ID of `vis\_0003838'. In addition, Fig.~\ref{fig:DFG_visualization} shows that some signs are captured from unusual angles. While such cases are valuable for assessing algorithmic robustness, generalization, and last-minute detection capability, their scarcity may limit effective learning and lead to unstable or less representative evaluation.
\input{figures_figures/42_traffic_sign_detection_DFG}

\begin{figure}[!t]
    \centering

    \includegraphics[width=0.95\columnwidth, height=0.12\columnwidth]{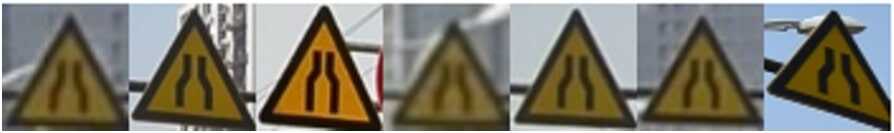} \\[0.2em]

    \fontsize{8pt}{0.5pt}\selectfont (a) Variations of Traffic Sign in Category `w8' in TDD100K \\[0.2em]

    \includegraphics[width=0.95\columnwidth, height=0.12\columnwidth]{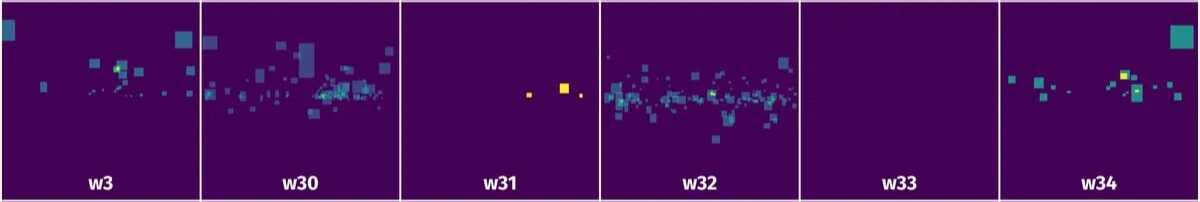} \\[0.2em]

    \fontsize{8pt}{0.5pt}\selectfont (b) Spatial Distribution of Traffic Sign Instances in Several Categories \cite{Tencent2021} \\[0.2em]

    \begin{tabular}{c@{\hspace{0.01\columnwidth}}c}

        \includegraphics[width=0.47\columnwidth, height=0.3\columnwidth, trim=0 1.3cm 0 17cm, clip]{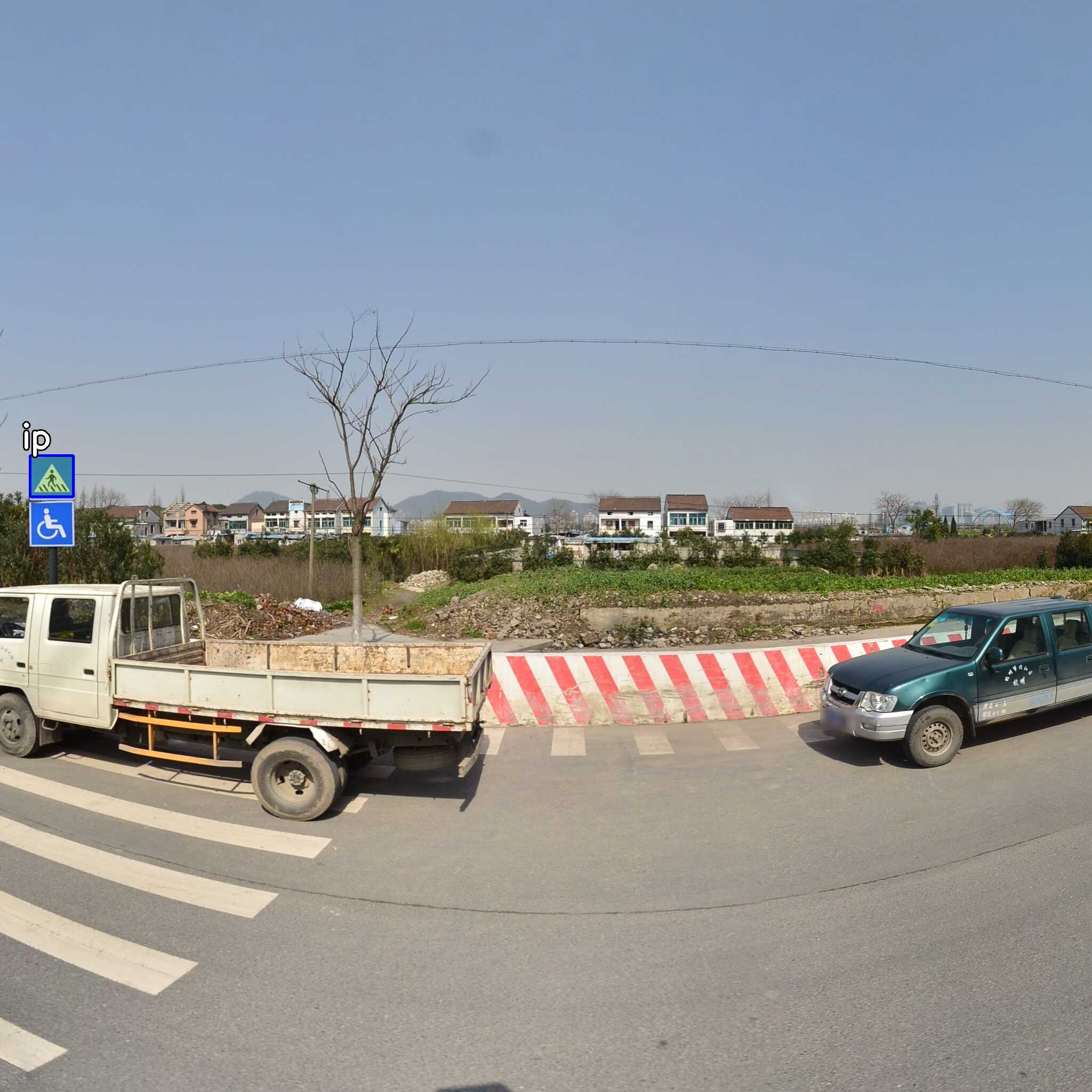} &
        \includegraphics[width=0.47\columnwidth, height=0.3\columnwidth, trim=0 1.3cm 0 17cm, clip]{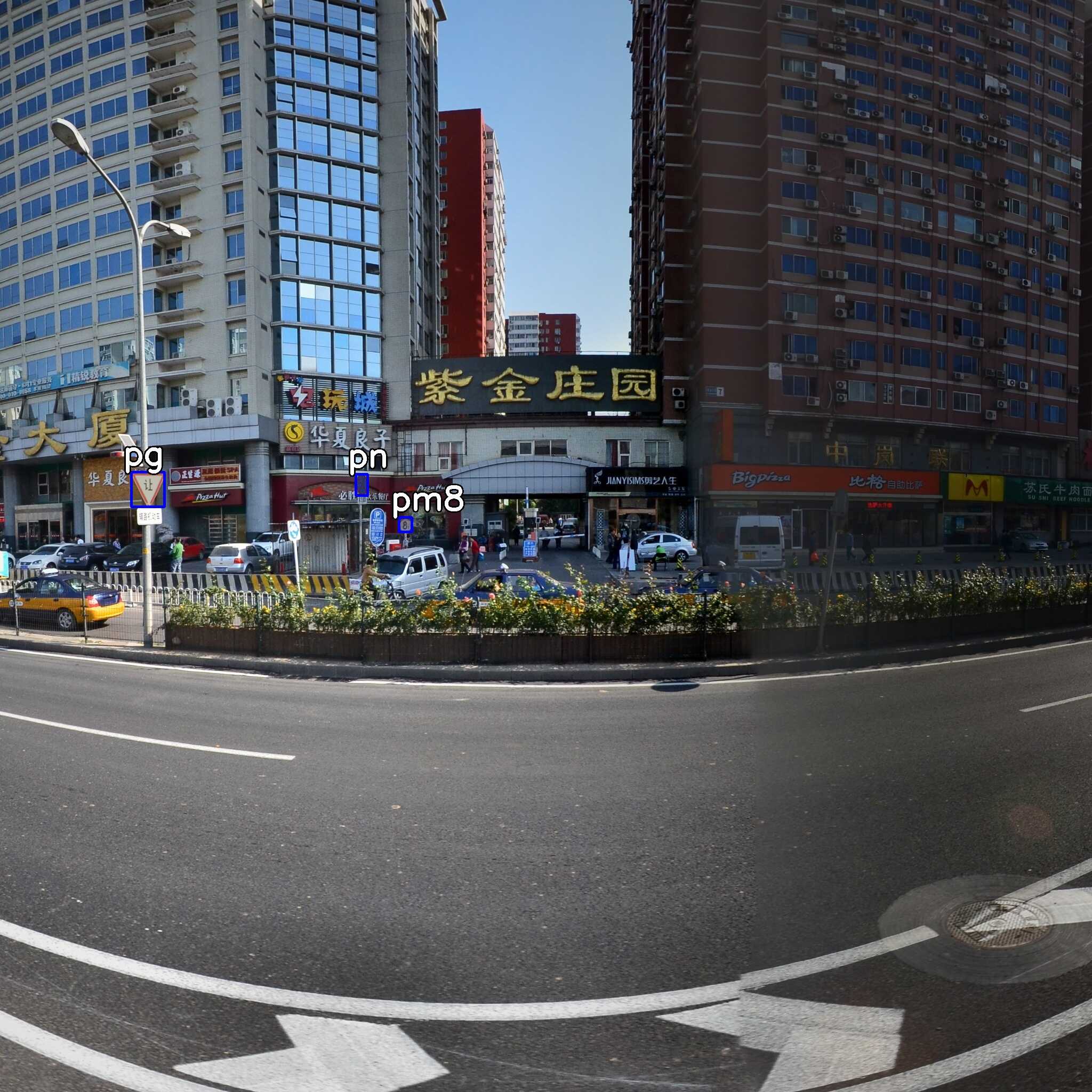} \\[0.2em]

        \includegraphics[width=0.47\columnwidth, height=0.3\columnwidth, trim=0 7cm 0 11.3cm, clip]{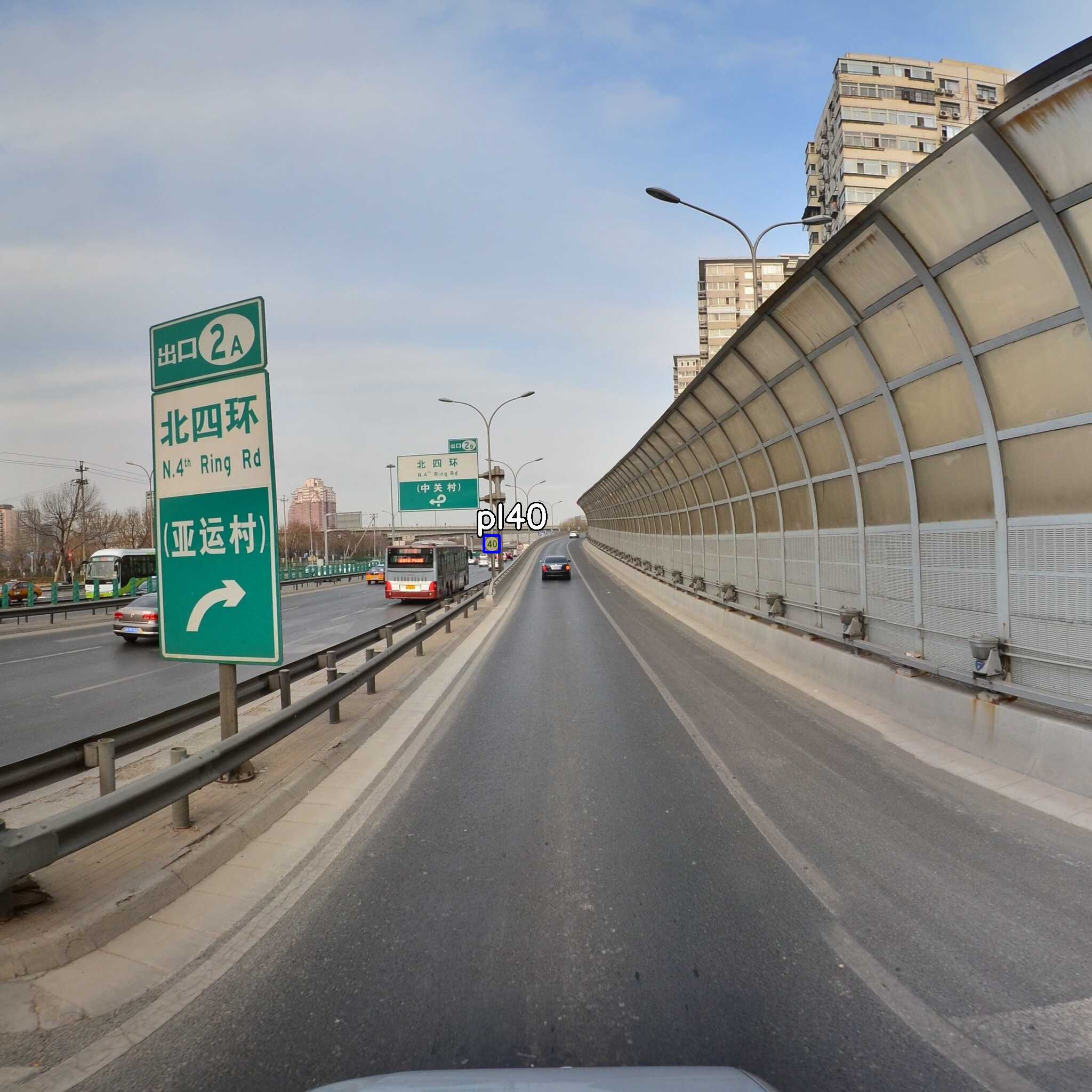} &
        \includegraphics[width=0.47\columnwidth, height=0.3\columnwidth, trim=0 9cm 0 9.3cm, clip]{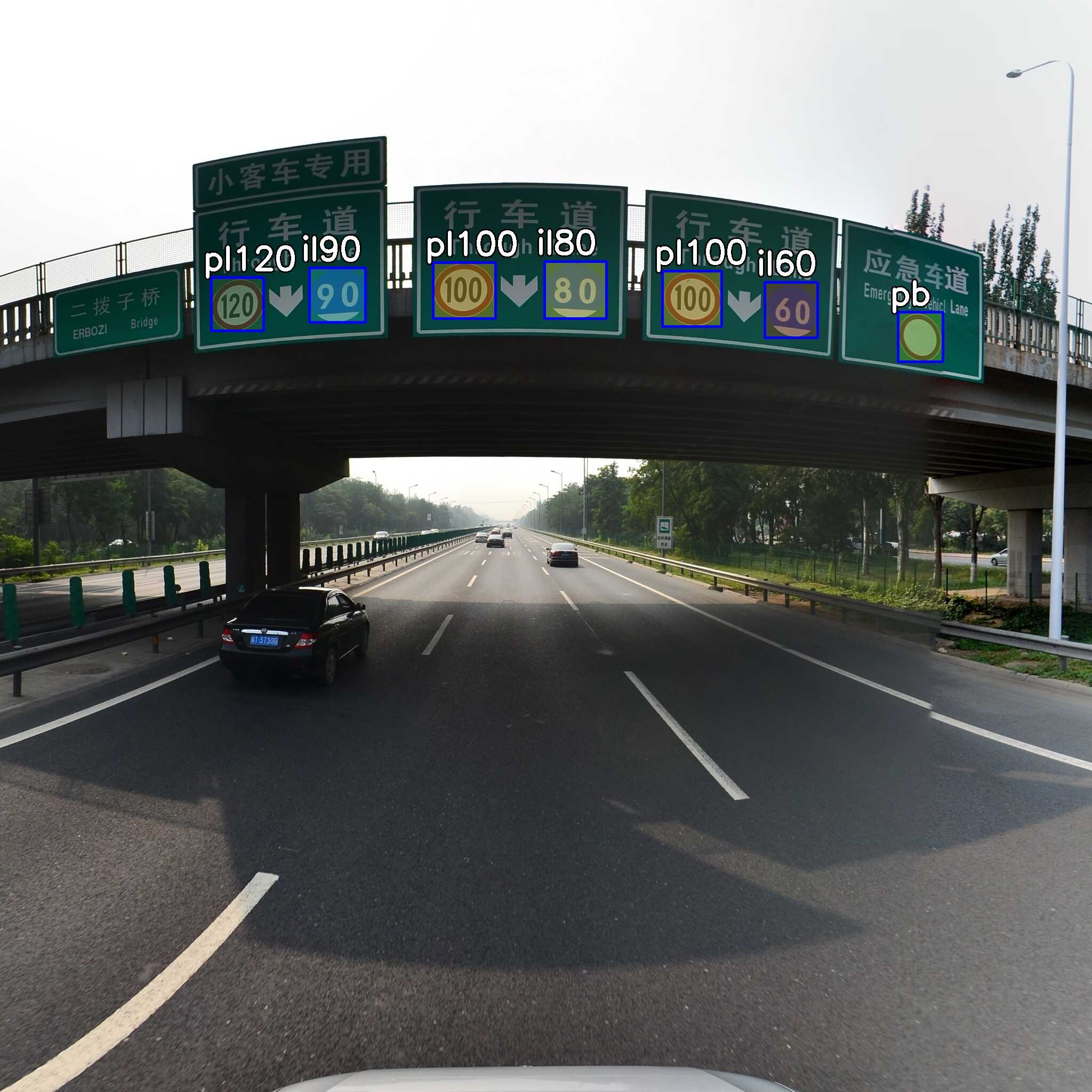} \\[0.2em]

        \multicolumn{2}{c}{\fontsize{8pt}{0.5pt}\selectfont (c) Sample Images (cropped) with Bounding Box and Category Label} \\
    \end{tabular}
    \caption{Illustrations of images and labels from TT100K \cite{Tencent2016}}
    \label{fig:TT100K_visualization}
\end{figure}
\paragraph{\textbf{TT100K:}}
\label{dataset:TT100K}
\href{https://cg.cs.tsinghua.edu.cn/traffic-sign/}{TT100K} \cite{Tencent2016} comprises image patches containing traffic signs, trimmed from panoramic streetview images in Tencent Maps. Original panoramas were captured using six single-lens reflex cameras mounted on vehicles or shoulders in about 300 Chinese cities and then stitched together \cite{Tencent2016}. The latest version proposed in 2021 \cite{Tencent2021} selects 100,000 images at 2048×2048 pixels (6105 training, 3071 testing, and 761 additional) from 5 cities and involves 232 categories of traffic signs. Each instance is annotated with a bounding box and a category label. As shown in Fig.~\ref{fig:TT100K_visualization} (a), TT100K captures traffic sign variations in each category, such as viewing angles and lighting. The spatial distribution patterns of the traffic signs (as revealed in Fig.~\ref{fig:TT100K_visualization} (b)) indicate the frequent and rare locations of the traffic signs within the images, potentially providing statistical priors for algorithm design and model training. Fig.~\ref{fig:TT100K_visualization} (c) shows that, even after cropping the top and bottom sections, the traffic signs in TT100K remain extremely small relative to the overall image dimensions due to the use of wide-angle panoramic views. This characteristic makes TT100K particularly suitable for specialized small-object detection algorithms. As shown in the top-right image of Fig.~\ref{fig:TT100K_visualization} (c), some signs are even visually integrated with the background environment, which increases detection difficulty and places higher demands on the discriminative capability and robustness of algorithms. Furthermore, as illustrated in the bottom row, the annotation protocol focuses exclusively on symbolic elements, while ignoring textual contents on traffic signs. As a result, it provides incomplete traffic sign semantics and limits the dataset value as a comprehensive benchmark. Meanwhile, the top-left image of Fig.~\ref{fig:TT100K_visualization} (c) reveals that some purely symbolic traffic signs are also omitted from the annotations, introducing inconsistent supervision and undermining evaluation reliability.

\input{figures_figures/44_traffic_sign_detection_GTSDB}
\paragraph{\textbf{GTSDB:}}
\label{dataset:GTSDB}
To evaluate the detection of symbol-based and text-based traffic signs and the classification of intraclass variability, \href{https://github.com/citlag/Traffic-Sign-Recognition}{extended GTSDB} \cite{GTSDB2020} is proposed in 2020. 
It contains 900 images (600 training and 300 testing), each cropped to 1360 × 800, captured in urban, rural, and highway areas around Bochum, Germany, during spring and autumn using a Prosilica GC1380CH camera. The annotation scope is significantly expanded to 164 classes in 9 supercategories, including prohibitory, mandatory, danger, priority, special regulation, information, direction, additional panels, and other signs. It offers raw images in PPM format, along with pixel-level masks and instance-level bounding boxes for traffic signs, though polygon coordinates require additional processing of the mask files to be extracted. Fig.~\ref{fig:GTSDB_visualization} (A) shows the raw images converted to PNG format, overlaid with bounding boxes and category labels. It reveals the significant variation in size and visibility of instances in GTSDB. Figs.~\ref{fig:GTSDB_visualization} (B) and (C) present masks provided in the `semantic' and `instances' folders of the official repository \cite{GTSDBGithub}. However, the semantic masks do not actually distinguish classes. Analysis of unique pixel values and instance contours in each provided instance mask (as overlaid in Fig.~\ref{fig:GTSDB_visualization} (C)) reveals that individual instances are not distinguished, regardless of the semantic category or instance identity. These limitations prevent GTSDB from supporting reliable class-aware semantic segmentation or true instance-level learning.

\input{figures_figures/45_traffic_sign_detection_Glare}
\paragraph{\textbf{GLARE:}}
\label{dataset:GLARE}
Released in 2023, \href{https://github.com/NicholasCG/GLARE_Dataset}{GLARE} \cite{Glare} focuses on traffic sign recognition under sun glare conditions, although our analysis indicates that the glare sources are predominantly located in the sky region of its images rather than directly on the traffic signs themselves. Specifically, GLARE classifies sun glare into four types, including isolated sun glare, sun with bright clouds, cloud-diffused glare, and camera-induced glare \cite{Glare}. Although \cite{Glare} presents GLARE as a traffic sign detection dataset, it actually only localizes one traffic sign per image and classifies it into one of 41 classes. It comprises 2157 images captured by three dashboard cameras in the Orlando area, with resolutions of 937×540 pixels (1,956 images) or 810×540 pixels (201 images). The CSV files provide additional attributes that specify whether a traffic sign is occluded and whether it appears on another road. Fig.~\ref{fig:Glare_visualization} presents sample images and overlays bounding boxes and category labels for annotated traffic signs. It indicates the diverse visibility conditions in GLARE, allowing robust algorithm development and evaluation. However, several issues have also been revealed. First, although only one traffic sign is annotated in each image \cite{Glare}, numerous images contain multiple traffic signs, without being cropped to isolate the target sign. Such unlabeled signs can introduce incomplete and potentially misleading supervision during training, and may also compromise evaluation reliability by leaving correct detections uncredited or falsely penalized. Furthermore, in scenes containing multiple traffic signs, the selection of which sign to annotate lacks consistency across adjacent frames. This is evident in Fig.~\ref{fig:Glare_visualization} (first row, last two columns), where different signs are annotated in consecutive frames of the same scene, despite all signs remaining visible. Moreover, as shown in the second row, despite the ego vehicle remaining stationary, the ground-truth bounding boxes for the same traffic sign are disparate across three consecutive frames.

\subsubsection{Traffic Sign Graph Generation}
\label{task:sign_graph}
The comprehensive understanding of traffic signs extends beyond instance-level localization and classification to include the recognition of the textual and symbolic content of each sign instance. This advanced interpretation requires a deeper analysis of the relations between traffic signs, including their spatial arrangements, hierarchical dependencies, and contextual interactions within the driving environment. In this context, several benchmarks are proposed to systematically represent traffic signs in a structured format, capturing both the individual components and their inherent relationships. We unify this emerging research direction as the task of \textit{Traffic Sign Graph Generation}. Two representative datasets proposed by the same group, \textit{\underline{C}ASIA-\underline{T}encent Chinese Traffic \underline{S}ign \underline{U}nderstanding} (CTSU) \cite{CTSU} and \textit{CASIA-Tencent \underline{R}oad \underline{S}cene} (RS10K) \cite{RS10K}, establish a foundation for this field.

\input{figures_figures/46_traffic_sign_graph_generation_CTSU}
\paragraph{\textbf{CTSU:}}
\label{dataset:CTSU}
CTSU \cite{CTSU} explores the semantic interpretation of traffic signs, \href{https://nlpr.ia.ac.cn/databases/CASIA-Tencent%20CTSU/index.html}{comprising 5000 images} with a single traffic sign as the only content in each image. These images, at varied resolutions, were captured in both urban and rural areas in China, with 4000 images for training and 1000 for testing. As indicated in Fig.~\ref{fig:ctsu_visualization}, images are divided into 13 groups according to their functionality, such as signs for interval speed measurement (top left image) and emergency lane (bottom left image). Meanwhile, \cite{CTSU} proposes a standardized format to represent sign information: `\textit{indicative information + content}'. The \textit{indicative information} (e.g., `current turn left is') is derived from the directional arrows, while the \textit{content} is determined by textual elements and specific symbols. The operator `+' represents the directional relationships between the arrow and its target element. CTSU classifies sign contents into three primary groups: arrowheads, texts, and symbols, and annotates them with bounding boxes and class labels, covering 1 text, 46 symbol, and 8 arrowhead classes. Although the formula suggests a single relationship type between arrows and their targets, CTSU actually has two relation types, including pointing relations (between arrows and target elements) and affiliation relations (among texts and symbols). Given this nature, we classify CTSU as a dataset for traffic-sign graph generation. Fig.~\ref{fig:ctsu_visualization} presents sample images overlaid with bounding boxes, instance IDs, class labels, and relation arrows. It reveals that CTSU images exhibit considerable variation in sizes and shooting angles, particularly across different functional groups, requiring algorithms to be robust to scale and viewpoint changes in real-world application. Furthermore, CTSU presents varying image quality, such as environmental noise (e.g., tree foliage) and low-resolution captures. The annotation schema can also be problematic. Labeling unrecognizable text as `\#\#\#0' introduces ambiguity into the supervision signal by forcing model to absorb a placeholder label that conflates multiple failure cases, while omitting occluded contents results in incomplete supervision and may bias the model against challenging real-world cases. In addition, annotation typos, especially for English sign texts, such as `Arrlvais' in the bottom-right image, reduce evaluation reliability, since apparent prediction errors may actually reflect annotation mistakes rather than genuine model failures.

\input{figures_figures/47_traffic_sign_graph_generation_RS10K}
\paragraph{\textbf{RS10K:}}
\label{dataset:RS10K}
RS10K \cite{RS10K}, proposed by the same group of CTSU \cite{CTSU} in 2023, supports higher-level traffic sign graph generation. Compared to CTSU \cite{CTSU}, which only focuses on component-to-component (\textit{C-C}) relations within a single traffic sign, RS10K \cite{RS10K} further covers the relations between traffic signs (\textit{S-S}) and the relations between the sign component and the corresponding traffic scene element (\textit{C-T}) (e.g. lanes), accordingly, instead of isolated traffic sign images, RS10K consists of dashcam traffic scene images captured in 31 Chinese cities. Although \cite{RS10K} reports 10,066 images, the \href{https://nlpr.ia.ac.cn/pal/RS10K.html}{publicly released package} contains 10,041 images, 51.89\% of images at 1920×1080 pixels, 46.83\% at 1280×720 pixels, and the remainder span six additional resolutions. Images also exhibit diversity in both traffic sign functions (e.g. indicating tollbooth) and road types (e.g., diversion and confluence sections). Meanwhile, RS10K \cite{RS10K} creates heterogeneous annotations, illustrated in Fig.\ref{fig:rs10k_visualization} using our custom color-coding system. First, key scene elements are delineated by polygons, including roads (yellow), lanes (green), signs (cyan), and sign contents consisting of arrows (pink), texts (red), and symbols (orange). Fig.~\ref{fig:rs10k_visualization} (A) identifies each element using `$\langle$semantic class name$\rangle$\_$\langle$instance ID$\rangle$\_$\langle$category ID$\rangle$', while each node in Fig.~\ref{fig:rs10k_visualization} (B) represents an entity or a pair of entities (gray), with $\langle$instance ID$\rangle$ as their unique identifier. Although $\langle$category ID$\rangle$ suggests the presence of subcategories within semantic classes, the meaning of these category IDs or their total count are not specified. Hence, Table \ref{table:visual_datasets_summary} only records the six main semantic categories. As shown in Fig.~\ref{fig:rs10k_visualization} (A), the annotations extend to both the opposing roadway that is separated by a median strip and the narrow lanes that require special attention for recognition. However, intersecting or parallel roadway segments, despite their physical connectivity, are labeled discrete entities, which may fragment physically connected road structure, hindering topological learning and potentially distorting evaluation. In particular, most annotations are provided in absolute coordinates, whereas sign elements are documented in relative coordinates, requiring additional analysis for proper interpretation. 
Second, entity relationships are recorded in a dedicated dictionary. To simplify annotation, RS10K \cite{RS10K} condenses the \textit{C-T} relation into the arrow-to-traffic-element (\textit{A-T}) relation, which becomes effectively sign-to-traffic-element (\textit{S-T}) relation when no arrow is present. Following CTSU \cite{CTSU}, the \textit{C-C} relations can be further categorized into association and pointing relations. As demonstrated in Fig.~\ref{fig:rs10k_visualization}, annotations also include bidirectional relations between lanes and roads. We visualize the additional relation category `all\_content\_assoc' in Fig.\ref{fig:rs10k_visualization} (B) but exclude it from the relation category count in Table~\ref{table:visual_datasets_summary}. Moreover, comparing entity and relation annotations reveals that some \textit{S-T} and \textit{C-C} association relations are omitted, which may create inconsistent supervision and ambiguity regarding whether such relations should be predicted.

\subsection{Pertinency due to Status or Condition}
\label{subsec:pertinency_status}
Beyond the inherent semantic category or spatial location, the dynamic status and temporal condition could also make traffic entities significant for traffic flow and safety decisions. For example, identifying a yellow traffic light and an emergency vehicle with active sirens beyond the simple recognition of traffic elements would be more specific to determine driving strategies. Accordingly, several specialized benchmarks have emerged to support this distinction, for example, recognizing the status of a traffic light or the gesture of a traffic police instead of solely localizing its position.

\subsubsection{Traffic Light Detection}
\label{task:light_detection}
Traffic lights, as important traffic control devices, were invented to guide, regulate, and warn drivers \cite{LISA}. 
Accurate traffic light detection, including both localization and state classification \cite{9109411}, 
is crucial for safe and compliant navigation at signalized intersections. This task remains challenging due to a) substantial variation in object scale across viewing distances, b) environmental factors that affect visibility and state interpretation, c) appearance similarity to vehicle lights, especially during precipitation, and d) regional diversity in orientation (e.g. vertical), signal types (e.g. arrow), and supplementary displays (e.g. countdown timers). 

Although early research often relied on locally-collected private datasets \cite{LISA}, several benchmarks have been developed. This section analyzes four datasets. \textit{LISA} \cite{LISA} is one of the pioneering efforts, offering images that capture day and night illumination conditions and annotations in EXCEL format. \textit{\underline{B}osch \underline{S}mall \underline{T}raffic \underline{L}ights \underline{D}ataset} (BSTLD) \cite{BOSCH} partitions its training and test sets by geographic location, introducing a notable domain gap between subsets. \textit{DriveU} \cite{DriveU} employs 6 (2018 version) or 8 (2021 version) attributes of traffic lights to represent more types of traffic lights, allowing for a more nuanced classification. Most recently, \textit{$\text{S}^2\text{TLD}$} \cite{s2tld} focuses on small and cluttered traffic lights, classifying them according to the color and state of the illumination. Representative images with annotation overlays of these four datasets are visualized in Fig.~\ref{fig:lisa_visualization} to Fig.~\ref{fig:s2tld_visualization}.

\input{figures_figures/48_traffic_light_detection_LISA}
\paragraph{\textbf{LISA:}}
\label{dataset:LISA}
Proposed in 2016, LISA \cite{LISA} contains 43,007 stereo images at 1280×960 pixels, captured in San Diego, during the day and night. It classifies traffic lights into seven categories based on type and status, including \textit{go}, \textit{go forward}, \textit{go left}, \textit{warning}, \textit{warning left}, \textit{stop}, and \textit{stop left}. As shown in Fig.~\ref{fig:lisa_visualization}, annotations, \href{https://www.kaggle.com/datasets/mbornoe/lisa-traffic-light-dataset}{provided in EXCEL files}, include the bounding box coordinates and category labels for traffic lights, together with the corresponding video frame number. However, some limitations are also revealed. First, some traffic light instances are omitted from the annotation, such as the unlabeled red-arrow traffic light in the last image, which should be labeled as \textit{stop left}. This omission may bias training and unfairly penalize correct predictions. Second, the second image demonstrates labeling errors, where a circular red signal is categorized as \textit{stop left} despite the lack of distinctive arrow-shaped characteristics, potentially introducing classification ambiguity for training and degrading fine-grained recognition. In addition, as indicated by the third image, LISA also contains images without traffic lights, which are mixed with positive images. While the inclusion of negative images can be beneficial for background learning, it may provide limited insight into fine-grained recognition and may inflate performance if a method benefits disproportionately from many easy negative examples.

\input{figures_figures/49_traffic_light_detection_BSTLD}
\paragraph{\textbf{BSTLD:}}
\label{dataset:BSTLD}
Released in 2017, \href{https://zenodo.org/records/12706046}{BSTLD} \cite{BOSCH} contains 13,427 images with 1280×720 pixels. Training images were collected in the San Francisco Bay Area, while testing images were extracted from a stereo video captured in Palo Alto. This geographical separation introduces notable variations between subsets. First, traffic-light density differs substantially. The training set contains up to 12 instances per scene with an average of 2.11, while the maximum and average in the test set are only 4 and 1.62. The second discrepancy presents in the category distribution. All 13 categories appear in the training set, while only 4 categories are present in the test set. These disparities indicate that the evaluation results on the test set may not directly reflect the learning result on the training set, while still being informative for assessing robustness and generalization under distribution shift. The low average density (overall: 1.81) indicates the frequent occurrence of unsignalized scenes (such as the right image in Fig.~\ref{fig:bstld_visualization} (b)). Annotation analysis further reveals that 38.09\% (1940 of 5093) training images and 14.24\% (1187 of 8334) testing images depict scenes without traffic lights. In addition, class imbalance is observed in both subsets. `\textit{Green}' is the dominant category, representing 48.41\% (5207 of 10756) of traffic lights in the training set and 56.12\% (7569 of 13486) in the test set. 
Meanwhile, the categorization scheme also presents several issues. First, traffic light classification appears to depend primarily on the arrow-shaped indicator. As shown in Fig.~\ref{fig:bstld_visualization} (a), the traffic light containing only an arrow (left) and the one containing a circular and an arrow-shaped indicator (right) are classified as `\textit{GreenLeft}'. Second, independent annotation of individual images results in class inconsistency of the same traffic light in adjacent frames, as illustrated in Fig.~\ref{fig:bstld_visualization} (c). In addition, some labels appear questionable. For example, although the `\textit{RedLeft}' traffic light is not present as a perfect arrow in the preceding frame (first image in Fig.~\ref{fig:bstld_visualization} (c)), classifying it as a `\textit{Red}' circle may still be inappropriate. These annotation issues can blur class boundaries and introduce noisy or ambiguous supervision.

\input{figures_figures/50_traffic_light_detection_DriveU}
\paragraph{\textbf{DriveU:}}
\label{dataset:DriveU}
Originally released in 2018 and updated in 2021, \href{https://www.uni-ulm.de/en/in/institute-of-measurement-control-and-microtechnology/research/data-sets/driveu-traffic-light-dataset/}{DriveU} \cite{DriveU} was collected in 11 German cities, with routes specifically selected to ensure traffic light coverage. To capture both near-field traffic lights around the stop line and distant ones, the dataset employs two cameras with different lenses, while annotations are provided for the left image of stereo pairs. According to the latest annotation files, DriveU contains 292,245 annotated traffic lights in 40,978 stereo image pairs at a resolution of 2048×1024, including 200,263 instances on 28,525 training images and 91,982 instances on 12,453 testing images. Instance-level annotations include bounding box coordinates, unique tracking IDs, and class labels. The original version encodes six traffic-light attributes in six-digit class codes, resulting in 344 classes. The latest version expands to eight textual attributes, including aspects, direction, occlusion, orientation, pictogram, reflection, relevance, and state, leading to 19200 classes. Fig.~\ref{fig:driveu_visualization} presents sample images converted from TIFF files, overlaying frame-level attributes (timestamp, vehicle speed and yaw rate) and core object-level attributes (unique ID, pictogram, and relevance). As indicated by the bounding box colors, DriveU comprises six state classes, including \textit{red}, \textit{yellow}, \textit{green}, \textit{off} (not involved in Fig.~\ref{fig:driveu_visualization}), \textit{unknown} (visualized in white), and \textit{red\_yellow} (visualized in orange). Fig.~\ref{fig:driveu_visualization} also reveals that the annotation schema involves multiple traffic-light types in the scene, including those for pedestrians and trams.
DriveU benefits from a high traffic-light density, averaging 7.13 annotated instances per frame, which supports learning in complex and realistic scenes. However, several issues should be noted. The high proportion of \textit{unknown} (27.2\%) and \textit{off} (7.5\%) may dilute the supervisory value of finer-grained categories, biasing algorithms toward less discriminative categories.
Fig.~\ref{fig:driveu_visualization} clearly demonstrates that while all traffic lights are assigned color attributes, many instances exhibit ambiguous characteristics, which are even challenging for human observers based solely on the images. In addition, frames with many traffic lights often contain visually poor-quality instances. As shown in the bottom-right image, which contains a maximum of 45 traffic lights, many instances are extremely small or unclear, making recognition difficult even for humans and potentially leading to less meaningful learning patterns. 

\input{figures_figures/51_traffic_light_detection_S2TLD}
\paragraph{\textbf{$\text{S}^2\text{TLD}$:}}
\label{dataset:S2TLD}
\href{https://github.com/Thinklab-SJTU/S2TLD}{$\text{S}^2\text{TLD}$} \cite{s2tld} is introduced in 2023 to support the detection of small and cluttered objects. It contains 1222 images with 1920×1080 pixels and 4564 images with 1280×720 pixels. Various environmental conditions are covered, including daylight, night, and precipitation. The 14130 traffic instances are categorized into five classes, including \textit{red}, \textit{yellow}, \textit{green}, \textit{off}, and \textit{wait on}. Each instance is annotated with a bounding box and a category label. Fig.~\ref{fig:s2tld_visualization} visualizes several sample images, with overlaid labels for annotated traffic lights, where the five categories mentioned above are distinguished by red, yellow, green, blue, and cyan, respectively. As illustrated, the categorization is based on the illumination color and the state of traffic lights, independent of numerical displays or symbolic indicators, and without differentiation between circular and arrow-shaped signals. It may limit the learning of functionally important finer-grained traffic light semantics, thereby reducing the model utility in scenarios that require lane-specific or turn-specific decision-making. The top images indicate that $\text{S}^2\text{TLD}$ \cite{s2tld} excludes pedestrian signals and bus-specific traffic lights from its scope. Consequently, annotations on 74.4\% images are homogeneous, providing limited diversity in supervisory signals and hindering the learning of less common but practically important cues. In addition, as evidenced by the bottom panel, annotation omissions are also present, which are particularly prevalent, affecting nearly all frames containing the corresponding traffic lights, introducing persistent incomplete and ambiguous supervision.

\subsubsection{Traffic Police Gesture Recognition}
\label{task:police_gesture}
In a traffic situation, the instruction given by a traffic police officer has the highest priority to follow \cite{9304675}. Hence, recognizing and interpreting hand gestures used by traffic police officers to direct vehicular movement is important for traffic management and road safety. Despite efforts and significance in recognizing gestures of traffic police \cite{9304675}, to our knowledge, the only publicly available dataset is \textit{\underline{T}raffic \underline{P}olice \underline{G}esticulous \underline{R}ecognition} (TPGR) \cite{HE2020248}.

\input{figures_figures/52_traffic_police_gesture_recognition_TPGR}
\paragraph{\textbf{TPGR:}}
\label{dataset:tpgr}

\href{https://github.com/zc402/ChineseTrafficPolicePose?tab=readme-ov-file}{TPGR} \cite{HE2020248} contains 21 imitation videos of eight typical Chinese traffic police gestures against various backgrounds (both indoor and outdoor). Twenty videos have a resolution of 1080×1080 and one training video has 540×540 pixels. Frame-to-frame annotations created for 132,831 frames are documented in a CSV file per video, which classifies the gesture present in the corresponding frame using categorical numerical labels. Fig.~\ref{fig:tpgr_visualization} presents sample frames extracted from TPGR videos. Among the texts overlaid on the images, only the numeric class ID comes from the annotation files. The video ID and frame ID are derived from the source video. The textual class name is interpreted according to \cite{HE2020248}. Fig.~\ref{fig:tpgr_visualization} shows that the recording environment, such as the indoor area and the sidewalk, is not the typical application scenario to recognize traffic police gestures. This mismatch may reduce the practical utility of this dataset by limiting real-world relevance and impairing generalization. Each video retains all frames, including those before and after gesture presentation, which introduces substantial temporal redundancy and necessitates additional preprocessing before usage. Furthermore, TPGR employs `delayed' labels to give models more time to gather classification features rather than expecting immediate recognition of the police gesture once it starts. For instance, as shown in the right scenarios in Fig.~\ref{fig:tpgr_visualization}, although the `pull over' gesture has already started at the 1109th frame and been ready at the 1115th frame, the class label is not assigned until the 1125th frame. This delayed labeling strategy introduces temporal misalignment, which may lead to ambiguous and biased supervision, incorrectly encouraging delayed prediction rather than early recognition.

\subsubsection{Traffic Salient Object Detection}
\label{task:traffic_salient}
Traffic salient object detection (SOD) highlights the saliencies in driving environments \cite{9703244}, compared to universal SOD that explores visual conspicuity in natural images \cite{mine}. Existing benchmarks are grounded in gaze behavior patterns and eye-tracking data \cite{9703244,10225448}, and therefore some studies frame the task as predicting driver attention in a specific driving environment \cite{9703244,10225448}. While earlier approaches modeled saliency primarily at the pixel level using heat maps \cite{9349146,8726396,9785373}, recent work has increasingly shifted to semantically meaningful object-level road users. \cite{7938411,8569438,9703244,10225448}. A notable application of traffic SOD is in night-time driving, where it can substantially enhance vehicle detection in challenging low-light conditions \cite{7938411}. 
However, there remains a significant gap in publicly available datasets specifically curated for the detection of salient traffic objects. As a workaround, researchers often invest efforts in creating custom annotations while leveraging benchmark traffic datasets. 

\subsection{Pertinency due to Kinematic Pattern}
\label{subsec:pertinency_kinematic}
Whether a traffic participant deserves particular attention in safety-critical driving decisions can also depend on its kinematic behavior, since its motion evolution directly affects conflict likelihood and planning urgency. Temporal motion cues, such as lateral offset, relative velocity, acceleration, heading change, and trajectory convergence, reveal behavioral relevance more explicitly than static appearance, and are therefore highly informative for maintaining driving safety. Despite the relative scarcity of dedicated studies and benchmark datasets, several tasks have been developed to characterize this form of pertinency.

\subsubsection{Lane Change Classification}
\label{task:lane_change_classification}
Cut-in and cut-out maneuvers are considered as two critical traffic scenarios \cite{lane_change_classification_2}. An automated system should be able to recognize these situations at an early stage using visual cues, similar to an attentive human driver \cite{lane_change_classification_2}. Although many existing studies of lane-change recognition rely on physical and contextual variables estimated through techniques such as sensor fusion, these intermediate representations are often noisy, difficult to measure accurately \cite{lane_change_classification_2}, and lack sufficient semantic information \cite{lane_change_classification_1}. In contrast, raw visual appearance cues offer a more robust and human-like alternative to recognize lane-change behavior \cite{lane_change_classification_2,lane_change_classification_1}. Accordingly, lane change classification is defined as a vision-based action recognition task. Given a sequence of frames in which the maneuver has already started and whose final frame is at or beyond the moment when the middle of the rear bumper crosses the lane marking, the goal is to classify the vehicle behavior as lane keeping, left lane change, or right lane change \cite{lane_change_classification_2,lane_change_classification_1}. \textit{PREVENTION} \cite{prevention} serves as the only benchmark dataset. 

\paragraph{\textbf{PREVENTION:}}
\label{dataset:PREVENTION}
Released in 2019, \href{https://prevention-dataset.uah.es/}{PREVENTION} \cite{prevention} contains urban scenarios, with a particular emphasis on critical maneuvers of surrounding vehicles in highway environments. It consists of five records with 1920$\times$600 resolution, acquired by three drivers over five different days in three areas of Madrid, totaling 356 minutes of driving and 540 km of coverage. The sensor platform comprises a forward-facing front camera, a backward-facing rear camera, one Velodyne HDL-32E LiDAR, one narrow-field-of-view long-range radar, and two broad-field-of-view radars. According to \cite{prevention}, vehicle instances are first segmented and tracked through a mixture of automatic and semi-automatic procedures, after which higher-level motion cues such as trajectories, lane geometry, and lane-change events are derived. 
However, the TXT files released on the website are not fully consistent with the data-format descriptions reported in \cite{prevention}, especially for lane change, which appears to be the core task supported by PREVENTION. While these discrepancies may stem from later revisions, the available documentation does not clearly explain them or fully specify the updated annotation schemas. Specifically, \cite{prevention} describes each lane-change maneuver using a compact five-field schema, $[id, type, frame, val1, val2]$, covering six event categories: left lane change, right lane change, cut-in, cut-out, hazardous situation, and pedestrian-related situation. However, the released \textit{lane\_changes.txt} file adopts a different seven-field structure, in which the last four fields denote the start frame, the central transition frame, the end frame, and a blinker flag. Although this updated interpretation is supported by the official C++ loader provided on the website, the meanings of the first two fields (\textit{ID} and \textit{ID\_m}) remain undefined, and the lane-change taxonomy is reduced to only two classes. Although the first field appears consistent with the vehicle identifications used in \textit{trajectories.txt}, \cite{lane_change_classification_1} treats the second field as a unique vehicle identification, and the official documentation does not offer explicit interpretations. Similarly, the released trajectory annotations follow a 10-field format, introducing two additional variables, $u$ and $v$, compared to \cite{lane_change_classification_1}, which are not defined. These inconsistencies and insufficient documentation introduce ambiguity in data parsing and may lead to inconsistent interpretations and incomparable experimental results. By contrast, the inconsistencies in \textit{lanes.txt} and \textit{labels.txt} are less consequential, since the extra fields are either trailing zeros or are explained on the website. Meanwhile, although the official materials consistently use the term `detection', PREVENTION also appears to support instance segmentation for the six road-user categories, but the absence of pixel-wise ground-truth masks in the public release may limit the direct application of this annotation layer.

\subsubsection{Lane Change Prediction}
\label{task:lane_change_prediction}
Analogous to lane change classification, lane change prediction is also defined as a vision-based three-class classification task. Given a sequence of frames in which the maneuver has not yet started and whose final frame occurs before the rear center of the vehicle reaches the lane marking, the goal is to classify whether the vehicle will remain in the current lane or change to the right or left lane. Despite being formulated as an anticipation task, this formulation may be of limited value from a real-world forewarning perspective. By the time the rear center of the vehicle reaches the lane marking, the vehicle has already partially occupied the target lane, indicating that the lane change is already nearly 50\% complete. Therefore, this setting is more consistent with late-stage maneuver recognition than with true early anticipation. Regarding the dataset, \textit{PREVENTION} \cite{prevention} is still the only benchmark adopted for this task. Hence, we do not repeat the analysis here again.




\subsubsection{Driving Maneuver Classification}
\label{task:driving_maneuvers_classification}

The classification of surrounding vehicle maneuvers broadens the problem from lane-change inference to a richer set of behaviors, such as overtaking, cutting in, and passing \cite{9304772,7789687}. A representative early vision-based study is \cite{7789687}, which classifies 14 surrounding-vehicle maneuvers in real-world highway scenes using a six-camera panoramic setup. However, this study dates back to 2016 and its evaluation was limited to ten highway sequences. In contrast, recent studies increasingly favor trajectory-derived features, event rules, and other structured motion representations \cite{10610995}. Despite the scarcity of dedicated benchmarks, \textit{BLVD} \cite{BLVD} partially fills this gap with interaction-oriented semantics.

\begin{figure}[!t]
    \centering

    
    \includegraphics[width=1\columnwidth]{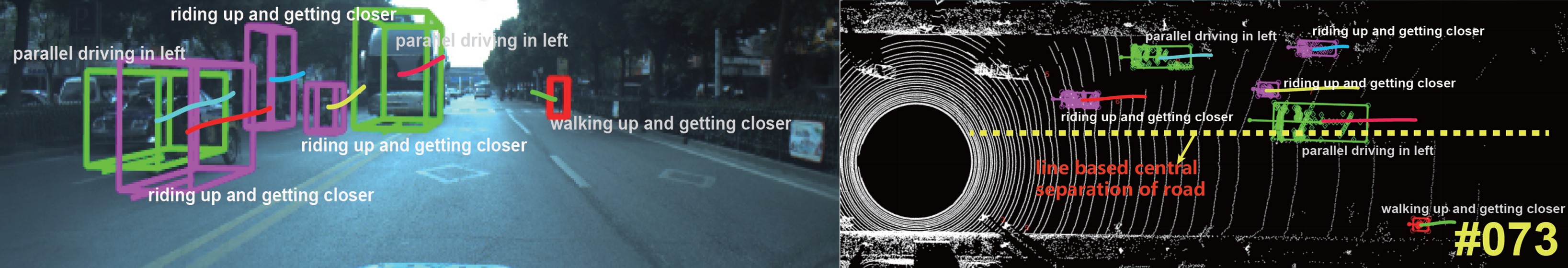}

    \caption{Sample frame from BLVD \cite{BLVD}, showing synchronized RGB imagery and 3D point clouds, with 3D bounding boxes for each traffic participant and the corresponding 5D interactive maneuver type annotated along each trajectory.}
    \label{fig:BLVD_examples}
\end{figure}

\paragraph{\textbf{BLVD:}}
\label{dataset:BLVD}
\href{https://github.com/VCCIV/BLVD/tree/master}{BLVD} \cite{BLVD} is a large-scale benchmark designed for 4D tracking, 5D interactive event recognition, and intention prediction in urban and highway driving environments. It comprises 654 calibrated clips with synchronized RGB images and 3D point clouds, collected in Changshu, China. The sensor suite consists of two cameras with a resolution of 1920$\times$500 pixels, a 64-beam LiDAR with a sensing range of 100m at 10Hz, and GPS. BLVD explicitly models the temporal evolution of surrounding traffic participants with respect to the ego vehicle. Specifically, it defines 13 ego-relative maneuver classes, such as parallel driving and overtaking from the left, and annotates them along the trajectories of 4902 instances, including vehicles, pedestrians, and riders, yielding a total of 6004 valid maneuver fragments. However, due to its limited geometric diversity and the relatively small number of samples in some maneuver classes, BLVD is less suitable as a standardized benchmark for driving maneuver classification. Nevertheless, it still provides a supportive resource for interaction-aware behavior analysis.

\subsubsection{Trajectory Prediction / Motion Forecasting}
\label{task:trajectory_prediction}
Trajectory prediction, also known as motion forecasting, aims to estimate the future paths of various dynamic traffic participants (e.g. vehicles and pedestrians) from an observed history of the scene \cite{trajectory_survey_1}. This anticipation capability enables situation awareness for safe maneuver planning and collision avoidance in complex traffic scenarios. Unlike intention prediction, which is typically considered as a discrete classification task, trajectory prediction is typically addressed as a regression task \cite{trajectory_survey_ijcai,trajectory_survey_iccv}.
Existing formulations can be broadly categorized along two orthogonal dimensions. The first is the input modality. Under the dominant \textit{structured-input forecasting} formulation, models predict future motion from pre-extracted agent trajectories or state histories, often together with contextual information such as HD maps and traffic states \cite{trajectory_survey_1}. This formulation underlies the most widely used benchmark datasets, including \textit{Argoverse2 Motion Forecasting} \cite{Argoverse2}, \textit{Waymo Open Motion} \cite{waymo_motion}, and \textit{INTERACTION} \cite{interactiondataset}, which primarily provide tracked agent histories with map-aware scene context rather than raw sensory observations. By contrast, \textit{perception-based forecasting} predicts future trajectories from raw or near-raw sensor inputs, such as LiDAR point clouds or camera streams, as exemplified by sensor-augmented benchmarks such as \textit{WOMD-LiDAR} \cite{WOMD-LiDAR} and end-to-end visual forecasting methods such as ViP3D \cite{ViP3D}. A related dataset is \textit{inD} \cite{inDdataset}, which is derived from drone videos but is typically used through released trajectory annotations rather than end-to-end visual input.
The second dimension is the forecasting target. \textit{Marginal prediction} forecasts the future of a single target agent independently, as in QCNet \cite{QCNet}, while \textit{joint prediction} forecasts multiple agents simultaneously to capture their interactions, as in QCNeXt \cite{QCNeXt}. The latter has become the dominant setting in recent autonomous-driving benchmarks.
Despite relevance to visual scene understanding, trajectory prediction datasets are not reviewed in detail here as the benchmarks generally do not use camera-based images or videos as input and are therefore outside the scope of this survey. Nevertheless, several representative datasets \cite{Argoverse2,Waymo,nuScenes} have been discussed earlier in the corresponding tasks.

\section{Discussions and Future Directions}
\label{sec:future}  

In Section \ref{sec:tasks_datasets}, we discuss 22 tasks that address traffic anomalies and 18 tasks that analyze pertinent entities. Across 34 tasks, we further examine the associated datasets, with 42 datasets on anomalies and 36 on pertinent entities. We conduct a dataset-by-dataset analysis of 78 perception datasets, examining their core properties, including sensor configurations, label design, visualization practices, and annotation quality (e.g., inconsistencies and omissions), as well as the implications for downstream learning, evaluation, generalization, and deployment. Building on these observations, we synthesize the findings through statistical summaries and a structured discussion of recurring patterns, underlying causes, practical implications, and future research directions. To avoid redundancy, we will not duplicate the elaboration of each individual dataset.

\subsection{Current Status and Research Gaps}
\subsubsection{General Topics}
\label{subsec:generaltopics}
The coverage of general topics by these datasets is summarized below, arranged by quantity in descending order.
\begin{itemize}[noitemsep]
    \item 16 involving \textbf{traffic accident scenarios} (\textit{\ref{dataset:tadd} }to\textit{ \ref{dataset:GTACrash}, \ref{dataset:dada2000}, \ref{dataset:riskbench}, }and\textit{ \ref{dataset:carla-syn}}),
    \item 12 specialized in \textbf{road damage} (\textit{\ref{dataset:CrackTree260} }to\textit{ \ref{dataset:rdd2022}}),
    \item 11 designed for \textbf{vulnerable road users} (pedestrians only: \textit{\ref{dataset:jaad}} to \textit{\ref{dataset:PedX}}, and \textit{\ref{dataset:Caltech}}; cyclists only: \textit{\ref{dataset:tdc}}; both: 
    \textit{\ref{dataset:waymo_pose}}, \textit{\ref{dataset:ECPDP}}, and \textit{\ref{dataset:nightowls}} to \textit{\ref{dataset:llvip}}),
    \item 9 centering on \textbf{lane markings} (\textit{\ref{dataset:tusimple}} to \textit{\ref{dataset:openlane}}, \textit{\ref{dataset:nuScenes}}, and \textit{\ref{dataset:Argoverse2}}), 
    \item 6 exploring \textbf{traffic signs} (\textit{\ref{dataset:DFG} }to\textit{ \ref{dataset:RS10K}}),
    \item 5 focusing on \textbf{adverse environmental conditions} (\textit{\ref{dataset:nightdriving} }to\textit{ \ref{dataset:darkzirich}, }and\textit{ \ref{dataset:nightowls}}),
    \item 5 about \textbf{unknown semantic categories} (\textit{\ref{dataset:LostAndFound}} to \textit{\ref{dataset:stu}}),
    \item 3 for \textbf{obstacles} (\textit{\ref{dataset:lostandfound}, \ref{dataset:RoadObstacle21}}, and \textit{\ref{dataset:riskbench}}),
    \item 3 involving \textbf{near-miss incidents} (\textit{\ref{dataset:cadp}, \ref{dataset:bdda} }and\textit{ \ref{dataset:dada2000}}),
    \item 2 investigating \textbf{traffic congestion} (\textit{\ref{dataset:CCTRIB} }and\textit{ \ref{dataset:UA-DETRAC}}),
    \item 2 analyzing \textbf{traffic lights} (\textit{\ref{dataset:DriveU} }and\textit{ \ref{dataset:BSTLD}}), and
    \item 1 classifying gestures of \textbf{traffic police} (\textit{\ref{dataset:tpgr}}),
    \item 0 dedicated to \textbf{traffic crimes} or \textbf{near-miss incidents}.
\end{itemize}

The above summary clearly discloses that the most underexplored domains are traffic crime, near-miss cases, traffic congestion, traffic lights, and cyclists. In particular, \textbf{\textit{traffic crime}} footage in existing datasets \cite{Sultani_2018_CVPR,MSAD} is collected from online surveillance videos, as revealed in Fig.~\ref{fig:Crime_visualization}, which have poor resolution. It further limits their utility in supporting specialized traffic crime analysis. Furthermore, due to the scarcity of publicly available \textbf{\textit{traffic congestion}} datasets and privacy and resolution limitations in the existing one \cite{CCTRIB}, elevated-view datasets \cite{UA-DETRAC} collected using professional cameras with moving vehicle annotations can also be utilized to advance traffic congestion analysis \cite{9901471}. Meanwhile, although a dataset \cite{Xia2017PredictingDA} targets critical braking events in regular traffic scenarios, existing datasets are not specifically designed for \textbf{\textit{near-miss incidents}}. Despite the challenges in capturing such events due to their transient nature, near-miss scenarios are crucial for safety analysis, as they provide insight into how severe accidents emerge and how they may be mitigated before occurring. Moreover, although the obvious visual differences between traffic lights in different states (e.g. yellow vs. red) might lead one to believe \textbf{\textit{traffic light}} identification is a straightforward field that does not require extensive studies, the reality is quite different. As discussed previously, traffic lights can vary in their physical orientation, signal types, and supplementary information display. These factors, combined with the challenges of visual similarity with vehicle lights and the effects of illumination variations, require comprehensive datasets and robust research. 
Additionally, although \textbf{\textit{riders}} are widely recognized as vulnerable road users in traffic environments \cite{cyclistdetection}, and their motion is often unpredictable and less rule-constrained \cite{BLVD}, few benchmarks have been developed specifically for this group. Existing studies typically include riders either as a semantic category for object detection or for higher-level analysis, such as pose estimation \cite{waymo_pose, ECPDP} and maneuver understanding \cite{BLVD}. However, riders are often subsumed under the broader \textit{pedestrian} category \cite{LLVIP}, treated merely as accident participants \cite{DAD,MM-AU}, or otherwise sparsely annotated, and thus receive only limited dedicated attention.
Despite the achievements in the available traffic sign datasets, due to regional variations in \textbf{\textit{traffic signs}}, the datasets typically have limited utility outside of their geographic origin. Algorithms designed for one dataset might not perform adequately when applied to different collections. In addition, as indicated in Fig.~\ref{fig:TT100K_visualization}, the spatial distribution patterns of traffic devices potentially provide statistical priors that can inform algorithm design and model optimization during the training phase.

\paragraph{\textbf{Anomaly:}}
\textbf{Comprehensiveness.}
As elaborated earlier, a comprehensive analysis of anomalies is supposed to identify all deviations from normal traffic patterns at the pixel (e.g., road damage), object (e.g., obstacles, dangerous road users, suspicious vehicles), event (e.g., accidents), and scene (e.g., environmental conditions) levels. However, although several studies recognize accidents as event-level anomalies \cite{DoTA} or cover multiple types of anomalies (collisions, obstacles, and road users with abnormal kinematic patterns \cite{RiskBench}), common approaches typically limit the `anomaly' to predesignated `unseen categories' outside of the 19 evaluation classes in Cityscapes \cite{chan2021segmentmeifyoucan,fishyscapes}. The dataset contents and annotations are also exclusively chosen and created to match the narrow definition. The lack of a recognized comprehensive taxonomy framework leads to separate territories for each task. Breaking through the barriers among the current isolated tasks to eventually developing integrated hierarchical approaches could be a promising direction. Fusing insights across all levels with attention mechanisms could contribute to a more robust and nuanced understanding of traffic anomalies. \textbf{Specificity.} The definition of deviation is actually highly context-based in traffic scenes. The complexity of the scenario results in an inherently fuzzy boundary between `anomalous' and `rare but normal'. Even human experts may also disagree on the classification of some edge cases, such as aggressive but legal maneuvers (e.g., speeding emergency vehicle). In particular, general visual and semantic clues are not adequate to distinguish between relevant anomalies (e.g., road hazards, road users with erratic behavior) and irrelevant variations (unusual but safe situations, e.g., roadside static abnormal categories). Incorporating traffic-specific knowledge, such as physical constraints and traffic rules, into anomaly models could be a promising solution, taking into account the severity and relevance of the anomaly. 

\subsubsection{Perspective}
The datasets surveyed involve three main perspectives, with 8 involving the surveillance view, 13 including local views that focus only on specific regions of interest, and 61 are based on ego views.
\paragraph{\textbf{Surveillance View:}}
Surveillance footage, typically captured from elevated positions, provides a broad and relatively unobstructed field of view, making it well suited to the analysis of large-scale traffic dynamics, such as congestion, accident evolution, and traffic violations. 
Not surprisingly, the four datasets for the classification of traffic congestion \cite{CCTRIB,UA-DETRAC} or traffic crime \cite{Sultani_2018_CVPR,MSAD} are exclusively based on surveillance videos, and the three datasets for traffic accidents analysis \cite{TADD,CADPDataset,you2020CTA} partially contain traffic surveillance footage. Such data are particularly valuable for macro-level traffic monitoring, enforcement, and emergency coordination, as they enable algorithms to capture interactions across multiple lanes and road users simultaneously.
In addition, the two datasets for traffic crime classification also include residential surveillance footage, which expands the relevance of this perspective to community safety and neighborhood monitoring. LLVIP \cite{LLVIP} further demonstrates that surveillance-view imagery can also support pedestrian detection in challenging low-light conditions.

However, surveillance-view data remain relatively scarce in public benchmarks. This is partly due to restricted access, as traffic surveillance systems are often controlled by authorized agencies and the corresponding footage is typically exclusive to authorized users. Publicly available footage, when available, usually suffers from insufficient resolution (as shown in Fig.~\ref{fig:Crime_visualization}), which constrains fine-grained recognition and precise localization. Moreover, variations in camera installation, calibration, and environmental conditions introduce substantial heterogeneity across scenes. Even when these issues are mitigated through carefully controlled acquisition, such as using professional cameras from an elevated position in UA-DETRAC \cite{UA-DETRAC} and LLVIP \cite{LLVIP}, models trained on such data may still face domain shift when transferred to real-world traffic monitoring systems with different hardware, viewpoints, and scene statistics. It suggests that models that perform well in controlled settings may struggle with the complexity and unpredictability of actual driving scenarios.
More broadly, the surveillance perspective is inherently detached from the viewpoint of drivers and onboard perception systems, which may further limit its direct applicability to ego-centric autonomous driving tasks.

\paragraph{\textbf{Local View:}}
Local-view imagery focuses tightly on specific objects or regions of interest and is therefore common in tasks requiring fine-grained visual inspection.
It is used by 11 of the 12 datasets for road damage analysis \cite{CrackTree,CrackForest,CRACK500,EdmCrack600,NHA12D,Pothole600,CQU-BPDD,SHREC,ExtendGAPs,RDD2022}, one of the two datasets for the generation of traffic signs graphs \cite{CTSU}, and the only dataset for the recognition of gestures by traffic police \cite{HE2020248}. By reducing irrelevant background clutter and centering the target region, this perspective facilitates the recognition of subtle patterns such as cracks, potholes, or detailed gestures.

However, the local view often suppresses valuable contextual information from the surrounding environment, such as surrounding traffic participants, road layout, and environmental conditions, resulting in a simplified visual setting that differs from real deployment conditions. 
This removal of real-world noise that autonomous systems must ultimately contend with can artificially lower task difficulty and may encourage models to rely on isolated appearance cues rather than context-aware reasoning. To mitigate this limitation, some datasets complement local-view images with ego-view footage, such as RDD2022 \cite{RDD2022}, while others directly adopt ego-view data for related tasks, such as M2S-RoAD \cite{M2S-RoAD} for road damage segmentation and RS10K \cite{RS10K} for traffic scene graph generation. These alternatives better preserve the visual complexity experienced by drivers and onboard perception systems, and therefore offer stronger support for realistic training and evaluation.

\definecolor{srcgreen}{RGB}{15, 110, 86}
\definecolor{perspamber}{RGB}{133, 79, 11}
\definecolor{scenblue}{RGB}{24, 95, 165}
\definecolor{sensorpurple}{RGB}{83, 74, 183}

\newcommand{\inlinebarcity}[2]{%
  \textcolor{#1}{\rule{#2}{5pt}}%
}

\newcommand{\mbar}{3.0cm}

\newcolumntype{L}{p{3.2cm}}  
\newcolumntype{B}{p{2.2cm}}  

\begin{figure}[t!]
\centering
\setlength{\tabcolsep}{5pt}
\renewcommand{\arraystretch}{0.95}
\begin{tabular}[t]{@{}p{0.47\textwidth}@{\hfill}p{0.47\textwidth}@{}}
\vspace{0pt}
    {\footnotesize
    \begin{tabular}{@{}Lr B@{}}
        \toprule
        \small\textbf{Data source} & \small$n$ & \\
        \midrule
        Author-collected    & 52 & \inlinebarcity{srcgreen}{2.00cm} \\[-1pt]
        Online resources    & 20 & \inlinebarcity{srcgreen}{0.77cm} \\[-1pt]
        Existing benchmarks & 13 & \inlinebarcity{srcgreen}{0.50cm} \\[-1pt]
        \bottomrule
    \end{tabular}
    }
    \vspace{7pt}

    {\footnotesize
    \begin{tabular}{@{}Lr B@{}}
        \toprule
        \small\textbf{Perspective} & \small$n$ & \\
        \midrule
        Ego view          & 61 & \inlinebarcity{perspamber}{2.00cm} \\[-1pt]
        Local view        & 13 & \inlinebarcity{perspamber}{0.43cm} \\[-1pt]
        Surveillance view &  8 & \inlinebarcity{perspamber}{0.26cm} \\[-1pt]
        \bottomrule
    \end{tabular}
    }
    \vspace{7pt}

    {\footnotesize
    \begin{tabular}{@{}Lr B@{}}
        \toprule
        \small\textbf{Scenario type} & \small$n$ & \\
        \midrule
        Normal traffic     & 50 & \inlinebarcity{scenblue}{2.00cm} \\[-1pt]
        Outliers           & 27 & \inlinebarcity{scenblue}{1.08cm} \\[-1pt]
        Traffic accident   & 16 & \inlinebarcity{scenblue}{0.64cm} \\[-1pt]
        Near-miss          &  3 & \inlinebarcity{scenblue}{0.12cm} \\[-1pt]
        Traffic congestion &  2 & \inlinebarcity{scenblue}{0.08cm} \\[-1pt]
        \bottomrule
    \end{tabular}
    }

&
\vspace{0pt}
    {\footnotesize
    \begin{tabular}{@{}p{4.2cm}r p{2.2cm}@{}}
        \toprule
        \small\textbf{Sensor type} & \small$n$ & \\[1pt]
        \midrule
        Monocular/Unknown camera    & 39 & \inlinebarcity{sensorpurple}{2.00cm} \\[2pt]
        Dashcam                     & 12 & \inlinebarcity{sensorpurple}{0.62cm} \\[2pt]
        Stereo camera               & 10 & \inlinebarcity{sensorpurple}{0.51cm} \\[2pt]
        LiDAR                       &  8 & \inlinebarcity{sensorpurple}{0.41cm} \\[2pt]
        Mobile phone                &  7 & \inlinebarcity{sensorpurple}{0.36cm} \\[2pt]
        Traffic surveillance camera &  5 & \inlinebarcity{sensorpurple}{0.26cm} \\[2pt]
        GoPro                       &  5 & \inlinebarcity{sensorpurple}{0.26cm} \\[2pt]
        Simulator                   &  3 & \inlinebarcity{sensorpurple}{0.15cm} \\[2pt]
        Radar                       &  2 & \inlinebarcity{sensorpurple}{0.10cm} \\[2pt]
        Monochrome                  &  2 & \inlinebarcity{sensorpurple}{0.10cm} \\[2pt]
        Surveillance camera         &  2 & \inlinebarcity{sensorpurple}{0.10cm} \\[2pt]
        PTZ camera                  &  1 & \inlinebarcity{sensorpurple}{0.05cm} \\[2pt]
        Reflex camera               &  1 & \inlinebarcity{sensorpurple}{0.05cm} \\[2pt]
        \bottomrule
    \end{tabular}
    }

\end{tabular}
\caption{Distribution of the 78 surveyed datasets by data source,
         perspective, scenario type, and sensor type.}
\label{fig:source-distribution}
\end{figure}

\subsubsection{Data Sources}
As summarized in Table~\ref{table:visual_datasets_summary} and Fig.~\ref{fig:source-distribution}, the surveyed datasets originate from three main sources. Twenty datasets contain footage selected from online resources, while 52 datasets are collected by their own authors. In addition, 13 datasets are built on existing benchmarks by synthesizing two images from each benchmark \cite{fishyscapes}, selecting images from benchmarks \cite{fishyscapes,TADD,CarCrashDataset,Xia2017PredictingDA,CRACK500,MM-AU,CST-S3D,SHREC,ECPDP,openlane-v,openlane}, refining annotations \cite{CST-S3D,SHREC,openlane-v}, and/or creating specific annotations \cite{CarCrashDataset,Xia2017PredictingDA,MM-AU,CRACK500,ECPDP,CityPersons,openlane}. 

These different sources reflect distinct trade-offs between realism, controllability, and annotation richness. Author-collected datasets, with deliberately designed sensor settings, viewpoints, and collection environments, are typically better aligned with the target task. However, they are costly to acquire and annotate and may introduce dataset-specific collection biases. Online resources provide efficient access to rare or safety-critical events, especially accidents, yet often suffer from limited metadata, inconsistent visual quality, and weak control over scene composition. Benchmark-derived datasets further reduce collection cost and may refine or extend annotation schemes, but their data quality and diversity remain inherently constrained by the original benchmarks.

\subsubsection{Sensor Configurations}
Fig.~\ref{fig:source-distribution} reveals variation in sensing configurations across the surveyed datasets, but only limited adoption of multimodal setups.
Among the 78 surveyed vision-based perception datasets, approximately 50\% rely on plain monocular imagery or do not clearly specify the camera configuration. Twelve datasets incorporate dashcam footage, owing to the widespread availability of dashboard cameras in practical driving environments. By contrast, only six datasets are based on surveillance footage, a scarcity likely attributable to the restricted accessibility of traffic surveillance data and the generally poor quality of publicly available surveillance footage. Mobile phones (n=7) represent a pragmatic and low-cost approach to data collection, but introduce variability in image quality, resolution, and mounting stability that is rarely controlled for. GoPro cameras (n=5) alleviate some limitations of mobile-phone capture, although their use remains limited. The adoption of more specialized cameras, including monochrome cameras (n=2), PTZ cameras (n=1), and reflex cameras (n=1), is marginal. In addition, sensors are mounted not only on vehicles but also on other platforms, such as shoulders in wearable setups \cite{Tencent2016}, motorcycles \cite{DAD,RDD2022}, and drones \cite{RDD2022}. Meanwhile, only 8 datasets complement camera data with LiDAR, while only 2 further incorporate radar. These observations suggest that multimodal datasets remain scarce, leaving a substantial gap relative to the sensor configurations adopted in contemporary autonomous vehicles.

\paragraph{\textbf{Real-World Data:}}
Real-world sensor data preserve the complexity of practical traffic environments, including natural occlusion, illumination variation, scene clutter, and motion dynamics. Cameras mounted near the vehicle center approximate the viewpoint encountered in actual driving and therefore provide realistic observations for downstream perception tasks.
Compared to general dashboard cameras, professional imaging devices may additionally offer higher resolution and more stable acquisition, facilitating the recognition of fine-grained details.
Beyond monocular RGB images, stereo cameras and LiDAR provide explicit depth cues, which are particularly valuable for 3D localization, scene reconstruction, and interaction reasoning. Specialized sensors, such as PTZ and monochrome cameras, are further adopted to improve perception under challenging low-light conditions. However, GoPro cameras have also been used to address adverse illumination. However, the examples in Fig.~\ref{fig:illumination_visualization} (top and bottom rows) exhibit even less clarity than the dashcam image (middle row) and appear inferior to those captured by the PTZ camera shown in  Fig.~\ref{fig:LLVIP_visualization}. 

Meanwhile, real-world collection also introduces substantial variability and noise.
Several datasets, especially those derived from online surveillance footage \cite{Sultani_2018_CVPR,MSAD,CADPDataset}, suffer from low resolution, which obscures distinctive features and weakens recognition of boundaries. 
Other datasets are affected by motion blur \cite{nightcity}, sensor noise \cite{CTSU,Pothole600}, or unconventional viewpoints \cite{DFG}. These factors are not merely incidental quality issues. They can act as confounding cues that statistically resemble anomalies, thereby increasing false positives and encouraging models to exploit acquisition artifacts rather than genuine traffic semantics. 
Consequently, these characteristics, together with the inherent complexity of traffic scenarios, directly affect the reliability of the annotation and the difficulty of the task, thus placing higher demands on the robustness of the downstream algorithms. 

\paragraph{\textbf{Simulation and synthetic data:}}
For scenarios where field data collection is expensive, dangerous, or impractical, online resources, simulation, and synthetic generation provide important alternatives. This is particularly evident for accident-related tasks. Among the 16 datasets that analyze traffic accidents, 13 are built from online footage, while 3 are generated in simulators. Compared with real-world collection, simulators offer much stronger controllability over weather, illumination, road structure, traffic participants, and event configuration. They also provide access to rich annotations, including not only spatial-temporal labels but also object attributes and physical states such as velocity and acceleration. 

However, simulation remains limited in realism. Although simulators can reproduce many traffic situations, they cannot fully capture the diversity, unpredictability, and behavioral authenticity of real-world incidents. Certain accident patterns (e.g. sudden stops \cite{CrashToNotCrash}) or natural driver reactions may therefore be underrepresented or oversimplified. Furthermore, as demonstrated in Fig.~\ref{fig:attention_datasets_comparison}, the responses of real drivers and surrounding agents in safety-critical situations are difficult to reproduce faithfully in laboratory environments. This domain gap implies that strong performance on simulated data may not translate directly to real-world robustness, especially for tasks requiring subtle behavioral understanding or anticipatory risk assessment.

\subsubsection{Collection Region}
Data collection region should be treated as a first-class characteristic rather than incidental metadata, especially for perception tasks of attention-worthy traffic elements. Such as the urban traffic density in nuScenes \cite{nuScenes} and the motorbike prevalence in DAD \cite{DAD}, such regional context shapes nearly all latent variables in a dataset, including road geometry, traffic device standards, environmental conditions, materials and maintenance quality of road infrastructure, behavior styles and density of road users. Consequently, two datasets with the same task label teach substantially different visual and behavioral patterns to a model. However, this factor remains insufficiently reported. Among the 78 datasets summarized in Table~\ref{table:visual_datasets_summary}, 19 lack clear country-level distribution information and 33 lack city-level distribution information. This reporting gap is especially evident in anomaly-related datasets, whereas pertinent-entity datasets tend to exhibit more systematic collection design and clearer geographic documentation.

\definecolor{europe}{RGB}{24, 95, 165}
\definecolor{asia}{RGB}{15, 110, 86}
\definecolor{namerica}{RGB}{133, 79, 11}
\definecolor{oceania}{RGB}{153, 60, 29}

\newcommand{\inlinebar}[3]{%
  \textcolor{#1}{\rule{#2\dimexpr#3\relax}{5pt}}%
}
\newcommand{\maxbarwidth}{3.2cm}

\newcommand{\tightrow}{\noalign{\vspace{-2pt}}}

\begin{figure}[t!]
\centering
\setlength{\tabcolsep}{4pt}

\begin{minipage}[t]{0.47\textwidth}
  \textbf{\textcolor{europe}{\small EUROPE}}\quad{\small $n = 52$}\\
  \vspace{-10pt}
  {\footnotesize
  \begin{tabularx}{\linewidth}{@{}l r X@{}}
    \toprule
    Country          & $n$ & \\
    \midrule
    Germany          & 14 & \inlinebar{europe}{1.000}{\maxbarwidth} \\[1pt]
    Switzerland      &  7 & \inlinebar{europe}{0.500}{\maxbarwidth} \\[1pt]
    United Kingdom   &  3 & \inlinebar{europe}{0.214}{\maxbarwidth} \\[1pt]
    Netherlands      &  3 & \inlinebar{europe}{0.214}{\maxbarwidth} \\[1pt]
    France           &  3 & \inlinebar{europe}{0.214}{\maxbarwidth} \\[1pt]
    Slovenia         &  3 & \inlinebar{europe}{0.214}{\maxbarwidth} \\[1pt]
    Spain            &  3 & \inlinebar{europe}{0.214}{\maxbarwidth} \\[1pt]
    Czech Republic   &  3 & \inlinebar{europe}{0.214}{\maxbarwidth} \\[1pt]
    Croatia          &  2 & \inlinebar{europe}{0.143}{\maxbarwidth} \\[1pt]
    Hungary          &  2 & \inlinebar{europe}{0.143}{\maxbarwidth} \\[1pt]
    Italy            &  2 & \inlinebar{europe}{0.143}{\maxbarwidth} \\[1pt]
    Poland           &  2 & \inlinebar{europe}{0.143}{\maxbarwidth} \\[1pt]
    Slovak Republic  &  2 & \inlinebar{europe}{0.143}{\maxbarwidth} \\[1pt]
    Norway           &  1 & \inlinebar{europe}{0.071}{\maxbarwidth} \\[1pt]
    Finland          &  1 & \inlinebar{europe}{0.071}{\maxbarwidth} \\[1pt]
    Ukraine          &  1 & \inlinebar{europe}{0.071}{\maxbarwidth} \\[1pt]
    \bottomrule
  \end{tabularx}
  }
\end{minipage}
\hfill
\begin{minipage}[t]{0.47\textwidth}

  \textbf{\textcolor{asia}{\small ASIA}}\quad{\small $n = 29$}\\
  \vspace{-15pt}
  {\footnotesize
  \begin{tabularx}{\linewidth}{@{}l r X@{}}
    \toprule
    Country       & $n$ & \\
    \midrule
    China         & 19 & \inlinebar{asia}{1.000}{\maxbarwidth} \\[-1.5pt]
    Japan         &  3 & \inlinebar{asia}{0.158}{\maxbarwidth} \\[-1.5pt]
    Singapore     &  1 & \inlinebar{asia}{0.053}{\maxbarwidth} \\[-1.5pt]
    India         &  1 & \inlinebar{asia}{0.053}{\maxbarwidth} \\[-1.5pt]
    UAE           &  1 & \inlinebar{asia}{0.053}{\maxbarwidth} \\[-1.5pt]
    South Korea   &  1 & \inlinebar{asia}{0.053}{\maxbarwidth} \\
    \bottomrule
  \end{tabularx}
  }

  \textbf{\textcolor{namerica}{\small NORTH AMERICA}}\quad{\small $n = 22$}\\
  \vspace{-15pt}
  {\footnotesize
  \begin{tabularx}{\linewidth}{@{}l r X@{}}
    \toprule
    Country        & $n$ & \\
    \midrule
    United States  & 18 & \inlinebar{namerica}{1.000}{\maxbarwidth} \\[-1.5pt]
    Canada         &  4 & \inlinebar{namerica}{0.222}{\maxbarwidth} \\
    \bottomrule
  \end{tabularx}
  }

  \textbf{\textcolor{oceania}{\small OCEANIA}}\quad{\small $n = 3$}\\
  \vspace{-15pt}
  {\footnotesize
  \begin{tabularx}{\linewidth}{@{}l r X@{}}
    \toprule
    Country    & $n$ & \\
    \midrule
    Australia  &  3 & \inlinebar{oceania}{1.000}{\maxbarwidth} \\
    \bottomrule
  \end{tabularx}
  }
\end{minipage}

\caption{Country-level data collection distribution across 59 datasets (106 country entries total), grouped by region.}
\label{fig:country-distribution}
\end{figure}
\paragraph{\textbf{Macro Geographic Distribution:}}
Fig.~\ref{fig:country-distribution} reveals that dataset collection is concentrated in a small number of geographic regions. Among the 78 surveyed datasets, 59 datasets clarify their data collection countries, yielding 106 country-level samples in total. Although several multi-country datasets inflate the apparent diversity of coverage, with the broadest examples \cite{ECPDP,EuroCity} spanning 12 European countries, the mean geographic coverage remains only 1.80 countries per dataset. And the distribution remains heavily skewed toward high-income, technologically mature nations in Europe and East Asia. Specifically, \textbf{Europe} accounts for the largest share, with 52 entries spanning 16 countries. Nevertheless, this representation is internally imbalanced, with \textit{Germany} alone accounting for 14 datasets, while several European countries appear only once or twice. \textbf{Asia} ranks second with 29 entries, but is driven by \textit{China} that appear in 19 datasets. North America contributes 22 entries and is similarly dominated by the United States, appearing in 18 datasets. \textbf{Oceania}, in contrast, is represented solely by \textit{Australia} involved in only 3 datasets. Most notably, Africa, South America, and Central Asia are entirely absent from the corpus. This geographical imbalance raises substantive questions whether the characteristics and patterns captured in these datasets and subsequently learned by algorithms can be generalized to the deployment contexts beyond the regions represented. Meanwhile, among the 29 datasets that do not report data collection countries, 3 are simulator-generated and 25 are collected from web sources. Only one dataset omits country information without an apparent methodological explanation, and this omission further propagates to one road-damage dataset that reuses it as a source dataset. Specifically, the absence of geographic metadata in the 28 web-sourced or simulated datasets reflects a methodological tradeoff to capture research subjects or scenarios that are rare, hazardous, or difficult to collect through controlled field acquisition. These datasets include 2 on road anomalies, 2 on crime activities, 1 on road damage, and 23 on traffic accidents.

\definecolor{barblue}{RGB}{83, 74, 183}
\definecolor{barpurple}{RGB}{139, 92, 246}
\definecolor{bargray}{RGB}{136, 135, 128}
\begin{figure}[t!]
\centering
\setlength{\tabcolsep}{6pt}
\renewcommand{\arraystretch}{1.3}
{\footnotesize
\begin{tabular}{@{}lrrlrl@{}}
  \toprule
  \small\textbf{Cities per dataset} & \small\textbf{Datasets} & \small\textbf{\% of 78} & & \small\textbf{\% of 45} & \\
  \midrule
  1
    & 20 & 25.6\% & \inlinebarcity{barblue}{1.82cm}
         & 44.4\% & \inlinebarcity{barpurple}{3.00cm} \\[-2pt]
  2--5
    & 11 & 14.1\% & \inlinebarcity{barblue}{1.00cm}
         & 24.4\% & \inlinebarcity{barpurple}{1.65cm} \\[-2pt]
  6--10
    &  8 & 10.3\% & \inlinebarcity{barblue}{0.73cm}
         & 17.8\% & \inlinebarcity{barpurple}{1.20cm} \\[-2pt]
  11+
    &  6 &  7.7\% & \inlinebarcity{barblue}{0.55cm}
         & 13.3\% & \inlinebarcity{barpurple}{0.90cm} \\[-2pt]
  \midrule
  Total (specified)
    & 45 & 57.7\% & & 100\% & \\[-2pt]
  \midrule
  \textit{Unspecified}
    & \textit{33} & \textit{42.3\%} & \inlinebarcity{bargray}{3.00cm}
                  & \textit{---}    & \\[-2pt]
  \bottomrule
\end{tabular}
}
\caption{Distribution of datasets by number of cities covered.
         \% of 78 is relative to all surveyed datasets
         (\textcolor{barblue}{$\blacksquare$} bars scaled to $n=33$, unspecified);
         \% of 45 is relative to datasets with a city count
         (\textcolor{barpurple}{$\blacksquare$} bars scaled to $n=20$, single-city datasets).}
\label{fig:city-distribution}
\end{figure}

\paragraph{\textbf{Intra-country Diversity:}} 
The city-level distribution of data collection further highlights the limited geographic granularity of existing datasets. Among the 45 datasets that report the number of collection cities, the distribution is heavily skewed. Twenty datasets (44\%) were collected in a single city, indicating that a large proportion of current benchmarks are derived from highly localized observations that may not generalize beyond their specific urban contexts. By contrast, only 6 datasets cover 11 or more cities, suggesting that genuinely large-scale, multi-city data collection remains underexplored. More critically, 33 datasets (42\%) do not specify any city-level information. Among them, 14 datasets are collected from real world, for which city-level provenance should in principle be available. This points to a significant transparency gap in geographic reporting.

\paragraph{\textbf{Rural Coverage:}}
Rural coverage is even more limited than country- and city-level diversity. Among the 78 surveyed datasets, only 5 explicitly state that their data collection includes rural areas. This suggests a strong urban bias in current benchmarks, while rural driving contexts remain largely overlooked. However, rural traffic scenarios differ substantially from urban roads in lane structure, pavement condition, roadside context, traffic density, illumination conditions, and frequency of interactions with vulnerable or unexpected road users, and thereby deserve dedicated attention. The lack of rural representation further limits the geographic and environmental diversity of current datasets, and may hinder the ability of trained models to generalize beyond predominantly urban deployment contexts.

\subsubsection{Annotation Design and Label Quality}
The surveyed datasets collectively span a broad range of perception annotation formats, including image-level classification, localization, object detection, semantic, instance, and panoptic segmentation, as well as relation triplets and vectorized polylines. However, this diversity is unevenly distributed across research topics, and not every task domain is supported by a well-developed spectrum of annotation formats. For example, traffic congestion is currently represented only through image-level classification, without corresponding detection- or segmentation-based benchmarks. Meanwhile, a dataset-level examination further shows that current benchmarks remain concentrated in relatively coarse forms of supervision.
There are 21 datasets that provide categorical labels, with 13 only supporting scene-level classification. Bounding-box annotations are the predominant format, comprising 1 dataset \cite{Glare} devoted exclusively to single-object localization despite the presence of multiple traffic signs, 3 datasets that contain both single-object and multi-object annotations, and 40 datasets that support multi-object detection, among which 26 are dedicated exclusively to detection tasks. Semantic masks are created in 23 datasets, 19 of which are dedicated to the semantic segmentation task. Specially, two datasets adopt semantic polylines and polygons to delineate road elements. Ten datasets contain instance masks, nine of which also support the detection task. Only 4 datasets support panoptic segmentation. In addition, relation labels are present in 3 datasets. Overall, current benchmark construction still concentrates on conventional object- or scene-level supervision, while richer annotations that support holistic scene understanding, instance-aware reasoning, and relational analysis remain limited.

\definecolor{clscol}{RGB}{133, 79, 11}
\definecolor{loccol}{RGB}{153, 60, 29}
\definecolor{detcol}{RGB}{24, 95, 165}
\definecolor{semcol}{RGB}{15, 110, 86}
\definecolor{inscol}{RGB}{83, 74, 183}
\definecolor{pancol}{RGB}{60, 52, 137}

\begin{figure}[h]
\centering
\captionsetup{justification=centering, singlelinecheck=false}
\begin{tikzpicture}[font=\small]
  \def\maxval{45}
  \def\plotW{7.5}
  \def\barH{0.30}
  \def\barGap{0.52}
  \def\labelW{2.6}

  \foreach \v in {0,5,10,15,20,25,30,35,40,45}{
    \pgfmathsetmacro{\xpos}{\v/\maxval*\plotW}
    \draw[gray!25, line width=0.4pt] (\xpos, 0.15) -- (\xpos, 5*\barGap+0.1);
    \node[below, font=\scriptsize, text=gray] at (\xpos, 0.12) {\v};
  }

  \node[below=14pt, font=\small] at (\plotW/2, 0.12) {Number of datasets};

  \foreach \task/\col/\excl/\shr/\yi in {
    {Image classification}/clscol/13/8/5,
    {Object localization}/loccol/1/3/4,
    {Object detection}/detcol/29/11/3,
    {Semantic segmentation}/semcol/19/6/2,
    {Instance segmentation}/inscol/1/9/1,
    {Panoptic segmentation}/pancol/3/1/0%
  }{
    \pgfmathsetmacro{\ypos}{\yi*\barGap+0.25}
    \pgfmathsetmacro{\exclW}{\excl/\maxval*\plotW}
    \pgfmathsetmacro{\shrW}{\shr/\maxval*\plotW}
    \pgfmathsetmacro{\totalW}{(\excl+\shr)/\maxval*\plotW}
    \pgfmathsetmacro{\total}{\excl+\shr}
    \fill[\col] (0, \ypos-\barH/2) rectangle (\exclW, \ypos+\barH/2);
    \fill[\col!35] (\exclW, \ypos-\barH/2) rectangle (\totalW, \ypos+\barH/2);
    \node[left, font=\small] at (-0.1, \ypos) {\task};
    \node[right, font=\scriptsize, text=gray] at (\totalW+0.05, \ypos)
        {\pgfmathprintnumber{\total}};
  }

  \fill[gray!70] (\plotW+0.3, 4*\barGap+0.10) rectangle (\plotW+0.65, 4*\barGap+0.37);
  \node[right, font=\scriptsize, align=left, text width=1.8cm]
      at (\plotW+0.72, 4*\barGap+0.235) {Exclusive to\\ that task};

  \fill[gray!25] (\plotW+0.3, 2*\barGap+0.10) rectangle (\plotW+0.65, 2*\barGap+0.37);
  \node[right, font=\scriptsize, align=left, text width=1.8cm]
      at (\plotW+0.72, 2*\barGap+0.235) {Also supports\\other tasks};

\end{tikzpicture}
\caption{Distribution of 78 surveyed datasets by annotation task: Solid bars for datasets exclusively supporting that task; Faded bars for datasets that also support other tasks; The combined length represents the total amount for that task.}
\label{fig:annotation-distribution}
\end{figure}

\paragraph{\textbf{Task-annotation Misclassification:}}
As analyzed in Sec.~\ref{sec:tasks_datasets}, some existing datasets provide annotations in formats that appear inconsistent with their stated tasks. For example, LostAndFound \cite{lostAndFound} is proposed as an obstacle detection dataset, while segmentation masks with semantic labels and instance IDs confirm its affiliation to the panoptic segmentation category. TADD \cite{TADD}, despite being named Traffic Accident Detection Dataset, only provides categorical labels for each video. Such mismatches blur the boundaries between benchmark tasks and can lead to confusion regarding what capabilities are actually being evaluated. Therefore, in this survey, we reassess dataset categorization based on their primary annotation formats to align each dataset with the six benchmark tasks. More broadly, those established standards should also be adhered to when proposing new tasks or determining the categorization of newly developed datasets, or a justification should be explicitly clarified about any departures from them.

\paragraph{\textbf{Inconsistency and Imbalance:}}
Inconsistent annotations have been observed in many datasets, such as the classification label (e.g. collision \cite{TADD,CST-S3D,DADA-2000} and traffic lights \cite{BOSCH}), the attribute label (e.g. occlusion \cite{ECPDP}), the decision to annotate or ignore a particular object (e.g. lane marking \cite{TuSimpleGit,VIL100}), spatial annotations \cite{CULane} and temporal information \cite{chan2021segmentmeifyoucan}. Detailed dataset-by-dataset examples and analysis have been provided in Sec.~\ref{sec:tasks_datasets}. Such inconsistencies, including ambiguous category definitions, uneven annotation criteria, and incomplete labeling, can encourage models to associate irrelevant features with the target class and give ambiguous boundaries between positive and negative. Furthermore, examination of annotation files also reveals a noticeable class imbalance in some datasets, especially in \cite{JAAD,BOSCH}. It can lead to over- or under-prediction of minority classes depending on the evaluation metric being optimized. Both the inconsistency and imbalance could reduce the reliability of the benchmark and complicate the interpretation of model performance. In this context, incorporating temporal and contextual information can help reduce the reliance on problematic static annotations. For example, temporal consistency across adjacent frames can help distinguish between genuine anomalies and transient artifacts, such as motion blur, sensor noise, or short-term occlusion, while contextual cues from neighboring agents, road layout, and scene type can prevent the model from relying solely on isolated appearance features. Meanwhile, scenario-aware thresholds may partially compensate for variations in class distribution or labeling reliability under different conditions. For example, different confidence thresholds can be used for daytime and nighttime scenes, or for urban intersections and highways. Cost-sensitive learning strategies may also alleviate bias in label distributions by assigning greater penalties to underrepresented or safety-critical classes, such as rare accident signals and  vulnerable road users.

\paragraph{\textbf{Insufficient Label Granularity:}}
First, some datasets fail to provide sufficiently granular labels for scenarios where individual differentiations are critical. As summarized in Table~\ref{table:visual_datasets_summary}, although the datasets define objects from various categories as obstacles \cite{lostAndFound,chan2021segmentmeifyoucan} or previously unseen instances \cite{lostAndFound,RoadAnomalyDataset,chan2021segmentmeifyoucan,fishyscapes}, these anomalous objects are actually treated as the same semantic category `anomaly', just to be distinguished from normal and void pixels. And only two datasets \cite{lostAndFound,RoadAnomalyDataset} support the identification of anomalies at the instance level. However, different anomalous instances present in the same scene often exhibit different properties, such as types and locations. They will require different levels of attention and responses \cite{mine}. Treating anomalies as a single class and without identifying individual instances makes it challenging to assess their relative importance, preventing the development of targeted strategies. In this context, instance masks with applicable attribute labels, such as severity of anomalies and intention of pedestrians, could enable object tracking and more effective analysis, facilitating customized decision making.

Furthermore, the lack of nuanced labels is manifested most prominently in classification datasets. For example, datasets for traffic congestion \cite{CCTRIB}, traffic crime \cite{Sultani_2018_CVPR,MSAD}, traffic accidents \cite{TADD,CADPDataset}, and road damage \cite{CQU-BPDD} only provide binary or multiclass labels for each video or image, without localizing the corresponding events or damages. In particular, despite containing high-resolution footage and diverse scenarios, some datasets also lack finer-grained annotations to fully utilize their potential. For example, CTA \cite{you2020CTA} classifies cause and effect segments of each accident video, and CST-S3D \cite{CST-S3D} applies hierarchies to categorize the collision. However, neither of them provides spatial annotations of accident participants, making it impossible to analyze accident causality through more targeted object tracking technologies. In this context, more detailed categorical labels (e.g., severity level) and more nuanced spatial labels would facilitate the development of algorithms that can not only recognize problematic scenarios but also pinpoint their exact locations for more effective intervention. However, not all existing datasets are amenable to such a label extension. The introduction of finer-grained labels also necessitates high-resolution footage, as both human annotators and machine learning models require sufficient detailed features to discern and delineate the critical elements. 

Meanwhile, higher-level labels might also be absent in datasets that support finer-grained tasks. For example, datasets delineating their research subjects with bounding boxes or masks usually do not have a scene-level category label (e.g., the type and severity of collisions) or environmental attribute labels (e.g., road type and weather condition). However, categorical descriptive labels at the image or video level might be beneficial to enable a prospective analysis of algorithm performance in different types of scenario. For example, grouping braking events based on street types in BDD-A \cite{Xia2017PredictingDA} might be conducive to analyzing attention patterns in different scenes. Classifying collision types within videos in \cite{CrashToNotCrash} could support the diagnosis of performance in recognizing dangerous behaviors in specific crash scenarios. Furthermore, despite the objective of anticipating accidents, DAD \cite{DAD} only provides spatial annotations for moving objects, suggesting that the location of accidents can only be roughly inferred based on the frame-to-frame involvement status of incident participants.

\subsection{Future Directions}
Building on the above analysis, we summarize actionable recommendations and future directions from the perspectives of task formulation, existing dataset adoption, future dataset construction, and future benchmark design, aiming to advance safety-oriented attention-worthy traffic perception.
\subsubsection{Unexplored Analytical Tasks}
From the perspective of conventional vision task categories, including classification, localization, detection, and segmentation, insufficient label granularity and footage quality have largely restricted existing benchmarks to coarse scene-level categorization or isolated object recognition, leaving several promising finer-grained and more context-sensitive tasks underexplored. Based on the examination in Section~\ref{sec:tasks_datasets} and the summaries in Table~\ref{table:Existing Tasks Classification} and Table~\ref{table:visual_datasets_summary}, we organize these gaps at the task-level as follows.
\begin{itemize}[noitemsep]
    \item Localization of abnormal events: road construction, traffic congestion, traffic accidents, and traffic crime;
    \item Detection and segmentation of unsurfaced roads;
    \item Classification of adverse environmental conditions and road conditions according to the hazard level;
    \item Detection of suspicious and damaged vehicles;
    \item Detection and segmentation of pedestrians and riders with dangerous behaviors (e.g., jaywalking);
    \item Detection and segmentation of multi-type anomalies;
    \item Detection and segmentation of users with right-of-way;
    \item Detection of specific places (e.g. parking, petrol station, speed bumps, and shared areas);
    \item Classification and detection of driving maneuvers (especially based on visual cues);
    \item Generation of traffic scene graphs with traffic-specific relations; 
    \item Detection and segmentation of multi-type pertinent entities.
\end{itemize}

\paragraph{\textbf{Recommendations:}}
While the above unexplored directions are summarized under conventional task categories, future task design should progressively move beyond these standard formulations toward interaction reasoning, temporal understanding, or fine-grained contextual interpretation for more comprehensive attention-worthy traffic perception.
\textbf{\textit{Taxonomic Alignment}:} Tasks should maintain clear alignment between task definitions and supervision formats, so that the provided labels accurately reflect the capability being evaluated. Mismatches between task naming and the supervision format can blur task boundaries and lead to misleading benchmark interpretation. 
\textbf{\textit{Analytical Granularity}:} Task formulation should move beyond coarse recognition toward more fine-grained and safety-aware analysis. In particular, tasks that are currently defined only at the image or video classification level should be progressively extended to instance-level, and severity-aware reasoning. For example, identifying congestion regions and right-of-way road users would provide more actionable information than scene-level labels alone. 
\textbf{\textit{Temporal Reasoning}:} Understanding temporal information should be emphasized in the future tasks. Many attention-worthy events or behaviors cannot be adequately characterized from static frames because their significance often depends not only on instantaneous appearance but also on how motions, interactions, and scene conditions evolve over time. For example, whether an unusual object is operationally irrelevant or requires urgent response, and whether a vehicle is behaving normally or developing into a collision risk, often cannot be determined without temporal reasoning. Accordingly, algorithms should be encouraged to reason about temporal continuity, motion trends, and interaction dynamics rather than rely solely on frame-level visual cues. Such capability is also important for reducing false alarms caused by transient artifacts, short-term occlusion, or ambiguous single-frame appearances.
\textbf{\textit{Contextual Relationality}:} Future tasks should support context-aware and relational formulations, such as traffic-specific scene graph generation, right-of-way reasoning, and ego-centric hazard analysis. In traffic scenarios, the significance of scene elements emerges from various context information rather than on its own semantic category alone.
\textbf{\textit{Operational Relevance}:} Future task design should distinguish more clearly between visual anomaly and operational relevance. Some rare elements may be visually unusual but operationally unimportant, while some ordinary elements can be highly attention-worthy because of their position, status, or interaction with the ego vehicle. Future formulations should characterize not only whether an element is anomalous, but also its significance for driving safety, for example, in terms of danger level or required response priority. 
\textbf{\textit{Practical Utility}:} Future benchmarks should align more closely with real-world deployment requirements. Beyond conventional closed-set recognition, they should place greater emphasis on open-world and long-tail perception, since traffic environments are inherently diverse and safety-critical elements can be rare. They should also better reflect the multimodal sensing configurations of real driving systems, enabling more realistic evaluation of practical perception capabilities. For example, richer dynamic information from vehicles and traffic sensing systems may improve the understanding of evolving traffic situations and support more robust reasoning.

\subsubsection{Recommendations for Future Benchmark Evaluation}
Current evaluations are mainly based on the mean performance on held-out test sets that share similar distributions with the training data and often focus on an isolated element of interest. However, these evaluations may be insufficient for attention-worthy traffic perception tasks, where models are expected to remain reliable under adverse circumstances or respond appropriately when presented in an unfamiliar scenario. Just as future task design should move beyond coarse and isolated formulations, future benchmark evaluation should also move beyond average in-domain accuracy as the dominant criterion. 
\textbf{\textit{Robustness}:} Future benchmark evaluation should place greater emphasis on robustness. For example, performance can be assessed using a dedicated auxiliary subset that contains footage captured in challenging conditions, such as adverse weather, low illumination, occlusion, and small or distant targets. More fundamentally, given the prevalence of annotation defects in existing datasets, whether learning under imperfect supervision could achieve a reliable performance is also important to evaluate, since robustness to noisy or incomplete labels is itself highly significant to practical traffic perception.
\textbf{\textit{Generalization}:} For tasks addressed within a closed-set dataset collected in specific regions and with specific sensor configurations, evaluations should assess how well the learned representations transfer across scenario types, cities, countries, sensor settings, or related analytical tasks. Particular attention should also be given to model behavior under open-world conditions. Accordingly, cross-domain and cross-sensor evaluation should become more common, which could better reveal whether reported performance gains reflect genuine generalization or merely adaptation to dataset-specific biases.
\textbf{\textit{Uncertainty}:}
Future benchmarking should also assess whether the model can accurately represent uncertainty when the input data is unclear or unfamiliar. This is especially important in traffic scenarios, as visually unclear observations, rare events, or open-world elements can lead to overconfident but unreliable predictions.
\textbf{\textit{Safety Relevance}:} Traffic-specific benchmarks should be formulated to reflect the practical importance of different prediction outcomes. In traffic scenarios, some errors are far more consequential than others. For example, underestimating imminent hazards is more fatal than miscategorizing two distant vehicles. Incorporating risk-aware metrics, priority-sensitive analysis, or scenario-based assessment of errors in safety-critical situations would therefore provide a more interpretable and practically meaningful evaluation.
\textbf{\textit{Efficiency}:} Computational efficiency of a perception system especially matters in safety-critical traffic scenario. It determines whether critical traffic elements can be identified in time for downstream decision-making, and whether strong offline performance remains usable under the latency, memory, and computational constraints of real-world deployment. Therefore, benchmark evaluation should explicitly include indicators such as runtime, inference latency, throughput, memory usage, and overall resource consumption.

\subsubsection{Strategies on Using Existing Benchmarks}
Before adoption, existing benchmarks should first be carefully examined to identify defects that can distort model learning or evaluation, ranging from low-level data quality issues to problematic annotation design. For example, for datasets with temporally adjacent frames, consistency checking can help identify contradictory labels assigned to the same element across frames. Once such issues have been identified, the datasets should be adopted with explicit awareness of their specific limitations, rather than being treated as uniformly reliable and comprehensive evaluation resources. Several strategies could be considered to mitigate identified limitations before implementing the experiments.
\textbf{\textit{Data Cleaning}:} After identifying the defects, such as missing, inconsistent, or ambiguous labels, the most straightforward and effective solution is to correct the label. However, for some large-scale datasets where relabeling is less feasible, the affected samples can be filtered or isolated as special subsets. This cleaning strategy could eliminate or even fundamentally reduce the noise introduced to training and evaluation.
\textbf{\textit{Complementary Resources}:}
The limitations of a single benchmark may be alleviated by incorporating complementary data resources, so as to exploit the value of existing datasets while partially offsetting their deficiencies. For example, when the adopted dataset suffers from weak, incomplete, or noisy supervision, a second benchmark with higher annotation quality may be incorporated as an auxiliary source for training, validation, or reference annotation. Similarly, if the original dataset has a narrow geographic coverage, another benchmark can complement to reduce bias induced by narrow data sources. 
\textbf{\textit{Customized Evaluation}:} When adopting a dataset with clearly identified limitations, the evaluation metrics should be adapted accordingly for justified interpretation and fair comparison. For example, for datasets with long-tail distribution, incomplete annotation, or restricted scenario coverage, imbalance-aware evaluation could be more informative and diagnostically valuable.
\textbf{\textit{Cross-domain Application}:}
Cross-domain application of higher quality datasets can break through the limitations of sparse labels and poor resolutions of existing resources within the target domain. More importantly, insights and patterns discovered in one domain can be applied to solve problems in another, potentially leading to novel solutions or contributing to a more integrated understanding of traffic scenes. For example, pedestrian pose may provide contextual cues for predicting traffic light states and potential collisions. Conversely, traffic light states, traffic sign contents, and lane marking types can also provide semantic information to predict pedestrian intention or driving behavior. Furthermore, annotations originally designed for a particular perception task can provide supplementary supervision for other perception tasks when underlying scene elements are shared. For example, BDD-A \cite{Xia2017PredictingDA} creates attention heatmaps based on gaze patterns on BDD100K \cite{BDD100K} images, while the inherent detection and segmentation labels enable the identification of attended regions and objects. Applying vehicle detection and tracking datasets with annotated moving vehicles, such as \cite{UA-DETRAC}, to the traffic congestion domain could equip plain classification algorithms with the ability to identify the congested area and the congestion process. In addition, large-scale datasets with real-world and simulated accident videos originally designed for dangerous vehicle recognition \cite{CrashToNotCrash} could also be generalized for a broader analysis of traffic accidents by introducing spatial and temporal annotations at the accident level.

\subsubsection{Recommendations for Future Dataset Construction}
Accordingly, new datasets should be designed to better support comprehensive and safety-critical traffic perception tasks by addressing current limitations in task alignment, geographic diversity, annotation completeness, practical usability, and documentation transparency.
Beyond standard hardware considerations, such as sensor calibration, temporal synchronization, and well-defined benchmarking protocols, our analysis identifies several key shortcomings that should be addressed in next-generation datasets:
\textbf{\textit{Conformity}:} Annotations should be explicitly aligned with the intended task and provide consistent, unambiguous supervision. For example, datasets that include only instance-level class labels and bounding box annotations should be clearly defined as supporting object detection, rather than being ambiguously positioned as segmentation datasets. Clear task definitions help avoid misuse and improve reproducibility and fair evaluation.
\textbf{\textit{Diversity}:} Increased geographic and contextual diversity is essential for improving model generalization. Datasets should cover multiple countries with different traffic regulations and driving behaviors, as well as multiple cities within each country that exhibit varied geographic and environmental characteristics. Both urban and rural settings should be represented, along with a broad spectrum of traffic scenarios, including intersections, school zones, highways, residential areas, suburban roads, and commercial districts. Achieving such diversity will likely require coordinated, large-scale, multi-institutional data collection efforts. In particular, region-specific datasets are important for addressing underrepresented areas that are currently lacking in existing benchmarks but are critical for the safe and reliable deployment of autonomous driving systems.
\textbf{\textit{Reliability}:} Minimizing annotation defects by reducing incomplete, inconsistent, and noisy supervision, while improving category clarity and consistency. Practical strategies include multi-stage annotation and verification pipelines to identify omissions and labeling errors, as well as temporal consistency constraints to ensure stable labeling of the same traffic entities across frames. In addition, category taxonomies should be carefully designed to preserve operationally meaningful distinctions, avoiding overly vague or weakly discriminative labels that can degrade learning performance.
\textbf{\textit{Completeness}:} Multi-level supervision enables a single dataset to support a broad spectrum of tasks, ranging from coarse scene understanding to fine-grained instance-level, temporal, and interaction-aware analysis. Achieving such extensibility requires both high-quality data and a well-structured annotation framework. The underlying data should exhibit sufficient visual fidelity, temporal continuity, and contextual coverage to support the incremental addition of annotation layers. Given the substantial cost of rich annotations, comprehensive multi-level labeling is more realistically achieved through community-driven efforts and collaborative extensions, rather than relying solely on the original dataset creators.
\textbf{\textit{Practicability}:} Datasets should more closely reflect real-world deployment conditions. In particular, multimodal data should be accurately calibrated and temporally synchronized to complement visual data in a manner consistent with sensing configurations in intelligent vehicles. This enables the development of perception systems that jointly leverage appearance, geometric, and motion cues. Furthermore, coverage of long-tail, yet safety-critical scenarios, such as near-miss events, should be strengthened. Despite their importance for safety evaluation and proactive risk assessment, such cases remain underrepresented in existing benchmarks.
\textbf{\textit{Transparency}:} Clear and detailed documentation of data acquisition conditions, sensor configurations, annotation protocols, and known limitations should be provided, while releasing as much data and metadata as feasible. More standardized and extensible benchmark designs would facilitate continuous updates, including the integration of new scenarios, annotation types, and sensing modalities, thereby better supporting sustained progress toward realistic and safety-oriented traffic perception.

\section{Conclusions}
\label{sec:conclusion}
In this paper, we propose an innovative taxonomy, comprehensively classifying the critical elements that warrant attention in traffic scenarios into two main groups, 13 categories. This unified analytical framework bridges inherently related but historically isolated fields. We comprehensively analyze 40 tasks and 78 datasets. The cross-domain investigation examines and illustrates the raw footage and annotations of each dataset in detail, including their characteristics, advantages, and limitations. We further discuss statistical analysis on general topics, data perspectives, data sources, sensors, annotations, and categorization, providing insight on existing limitations and promising solutions. Beyond comprehensive details, we also provide recapitulatory tables and navigation guides for efficient comparison and targeted information retrieval. We envision that our comprehensive work will equip researchers with a holistic overview of the domain, facilitate the strategic selection of appropriate resources for their studies, and illuminate critical gaps that merit further investigation. We contribute to necessary further developments in this rapidly evolving area by laying a foundation towards optimized accessible resources and more informed research directions.

\bibliographystyle{elsarticle-num} 
\bibliography{references}

\end{document}